\documentclass{article}

    \PassOptionsToPackage{numbers, compress}{natbib}

\usepackage[preprint]{neurips_2026}

\usepackage[utf8]{inputenc} 
\usepackage[T1]{fontenc}    
\usepackage{hyperref}       
\usepackage{url}            
\usepackage{booktabs}       
\usepackage{amsmath}        
\usepackage{amsfonts}       
\usepackage{nicefrac}       
\usepackage{microtype}      
\usepackage{xcolor}         
\usepackage{graphicx}       
\usepackage{subcaption}     
\usepackage{enumitem}       
\usepackage{adjustbox}      
\usepackage[table]{colortbl} 
\usepackage{multirow}       
\usepackage[most]{tcolorbox}
\usepackage{wrapfig}

\definecolor{darkred}{rgb}{0.7, 0.0, 0.0}
\definecolor{pumagreen}{rgb}{0.0, 0.6, 0.3}
\definecolor{methodgray}{gray}{0.85}

\title{Stop When Reasoning Converges:\\ Semantic-Preserving Early Exit for Reasoning Models}

\author{%
  \textbf{Dehai Min}$^{1,*,\dagger}$,\quad
  \textbf{Giovanni Vaccarino}$^{1,4,*}$,\quad
  \textbf{Huiyi Chen}$^{1}$, \\[3pt]
  \textbf{Yongliang Wu}$^{3}$,\quad
  \textbf{Gal Yona}$^{2}$,\quad
  \textbf{Lu Cheng}$^{1}$ \\[6pt]
  $^{1}$University of Illinois Chicago\quad
  $^{2}$Google Research \\
  $^{3}$University of Illinois Urbana-Champaign\quad
  $^{4}$Politecnico di Milano \\[2pt]
}

\begin{document}

\maketitle

\begingroup
\renewcommand{\thefootnote}{\fnsymbol{footnote}}
\footnotetext[1]{Equal contribution.}
\footnotetext[2]{Corresponding author: \texttt{dmin10@uic.edu}}
\endgroup

\begin{abstract}

Large Reasoning Models (LRMs) achieve strong performance by generating long chains of thought (CoT), but often overthink, continuing to reason after a solution has already stabilized and thereby wasting tokens and increasing latency. 
Existing inference-time early-exit methods rely primarily on answer-level signals, such as confidence or trial-answer consistency, to decide when to stop. However, these signals mainly reflect answer readiness rather than reasoning convergence: they may trigger before the model has finished exploring or self-correcting, causing premature exits that can degrade final-answer accuracy and leave the retained reasoning chain semantically incomplete.
We identify reasoning-level semantic redundancy as a complementary signal for semantic-preserving early exit: when successive steps no longer add novel progress and instead revisit established conclusions, the reasoning trajectory has likely converged. Building on this insight, we propose PUMA, a plug-and-play framework that combines a lightweight Redundancy Detector with answer-level verification. The detector flags semantically redundant candidate exits, while verification confirms whether stopping is safe, allowing PUMA to remove redundant continuation while preserving both answer accuracy and a coherent reasoning prefix.
Across five LRMs and five challenging reasoning benchmarks, PUMA achieves 26.2\% average token reduction while preserving accuracy and retained CoT quality. 
Additional experiments on code generation, zero-shot vision-language reasoning, and learned stopping-policy internalization further demonstrate that reasoning-level redundancy is a robust, transferable, and learnable signal for efficient reasoning.
Our code is available at \url{https://github.com/giovanni-vaccarino/PUMA}.

\end{abstract}

\begin{figure*}[t!]
    \centering
    \vspace{-0.25cm}
    \includegraphics[width=0.95\textwidth]{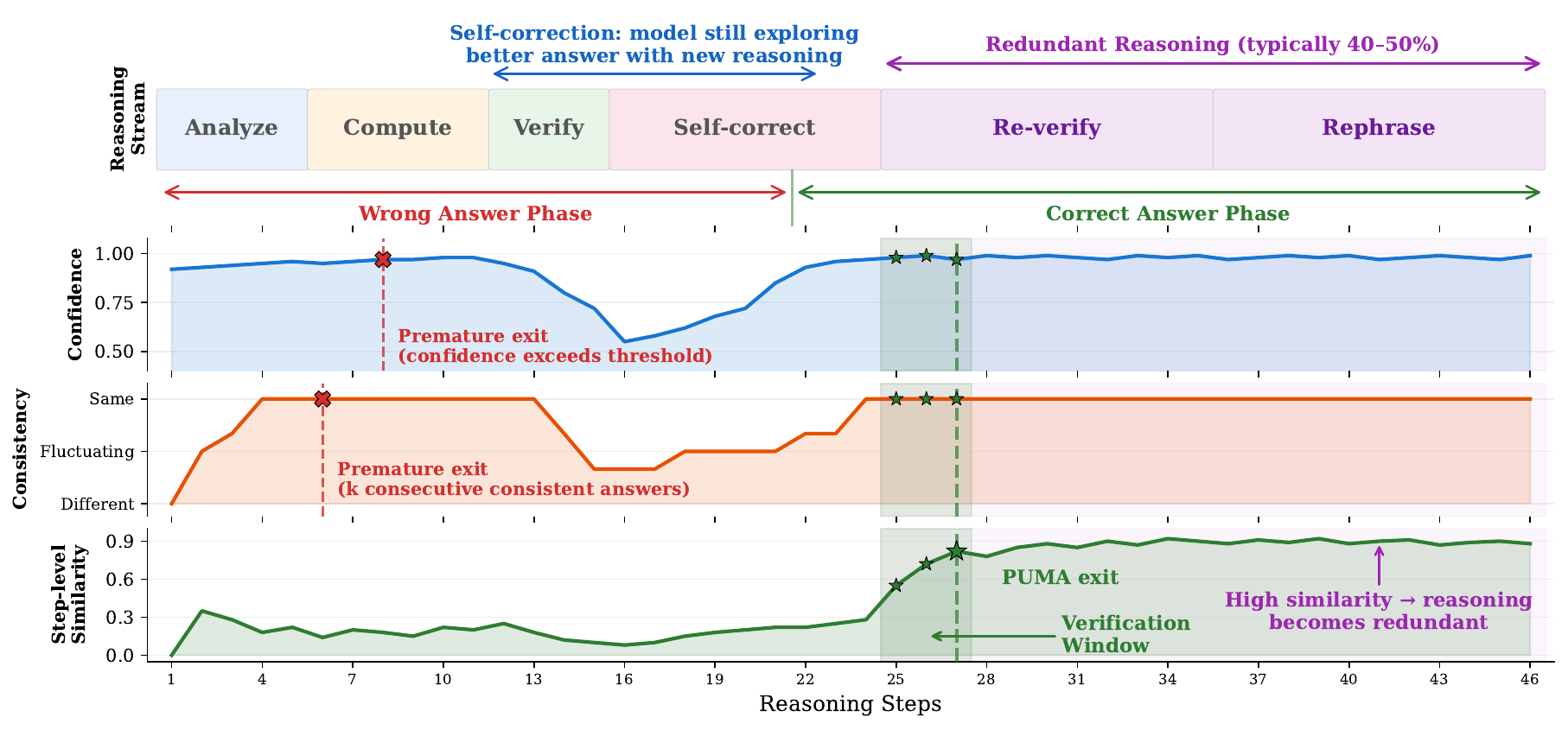}
    \caption{Answer readiness does not imply reasoning convergence.
    Confidence and trial-answer consistency can trigger premature exits while the model is still exploring or self-correcting. In contrast, step-level semantic similarity remains low during exploration and rises near reasoning convergence, enabling a more reliable exit signal aligned with reasoning convergence.}
    \label{fig:signal_comparison}
    \vspace{-0.55cm}
\end{figure*}

\section{Introduction}

Recent Large Reasoning Models (LRMs) such as DeepSeek-R1~\cite{guo2025deepseek} and OpenAI o1~\cite{jaech2024openai} achieve strong performance by generating long chains of thought (CoT)~\cite{wei2022chain} and scaling test-time computation~\cite{muennighoff-etal-2025-s1,yang2026towards,yang2025qwen3,snell2025scaling}. Beyond improving final-answer accuracy, these reasoning chains are often surfaced to users as explanations for final answers~\cite{korbak2025chain,gao2026how,li2025tldrlongreweightingefficient} or agent actions~\cite{10.5555/3692070.3694642,yao2023react,hsu2025groupthinkmultipleconcurrent, li2026phgpo, zou2025llm}, serving as an important basis for interpretability and user trust.
However, long CoTs also introduce substantial inefficiency: models often continue reasoning after a solution has stabilized, repeatedly re-verifying or rephrasing established conclusions~\cite{chen2025do,sun2025stop}. To quantify this redundancy, we analyze five representative LRMs and find that 41--52\% of reasoning tokens are generated after the model has already reached its final answer (see Figure~\ref{fig:overthinking_analysis} in Appendix~\ref{app:overthinking}). This creates a clear opportunity for more efficient reasoning, but the dual role of CoT makes naive truncation inadequate: an effective method should reduce unnecessary tokens while preserving final-answer accuracy and the coherent, semantically complete retained reasoning chain.

A growing body of work has sought to improve reasoning efficiency in LRMs. Training-based methods provide direct control over reasoning length~\cite{zhang-etal-2025-lightthinker,arora2025training,luo2025opruner,kang2025c3ot,ma-etal-2025-cot,team2025kimi}, but typically require per-model retraining. Prompt-based compression methods are lightweight and easy to apply~\cite{xu2025chain,lin2026plan,nayab2024concise,han-etal-2025-token}, but may hurt accuracy when brevity instructions suppress necessary intermediate reasoning. In contrast, inference-time early-exit methods~\cite{rustagi2025confidencecoverage,huang2025efficient,mao2025early,wang2025entropy,yang2026dynamic,fu2025reasoning} are attractive because they can reduce unnecessary reasoning at deployment time without modifying model weights. However, many existing early-exit methods primarily rely on answer-level signals, such as confidence estimates~\cite{yang2026dynamic,huang2026efficient,rustagi2025confidencecoverage} or trial-answer consistency~\cite{liu2025answer,fu2025reasoning,mao2025early}. While these signals are useful for judging whether the current answer appears stable, they do not directly measure whether the reasoning process has converged. As Figure~\ref{fig:signal_comparison} illustrates, confidence and trial-answer consistency can satisfy stopping criteria while the model is still exploring or self-correcting, leading to premature exits that may degrade final-answer accuracy or truncate important intermediate reasoning. This gap motivates a complementary stopping signal that tracks whether the reasoning trajectory is still making semantically novel progress.

We draw inspiration from semantic entropy~\cite{kuhn2023semantic,farquhar2024detecting}, which estimates uncertainty in LLM outputs by measuring whether multiple generated responses are semantically diverse or collapse to the same meaning, rather than comparing surface forms. We transfer this idea from \emph{across-output} diversity to \emph{within-trajectory} progress: if recent reasoning steps become semantically similar to prior steps and no longer introduce new logical or semantic content, the reasoning process is likely shifting from exploration to convergence. 
Under this view, once recent steps become semantically redundant, continued generation is likely to repeat established reasoning rather than add meaningful progress. We therefore use reasoning-level semantic redundancy as a complementary candidate-exit signal.

Building on this signal, we propose \textbf{PUMA}, a \emph{\textbf{P}rogress-aware \textbf{U}nified \textbf{M}onitoring framework for \textbf{A}daptive early exit} in reasoning models. PUMA reduces redundant reasoning tokens while preserving both final-answer accuracy and the semantic quality of the retained CoT. It pairs a lightweight Redundancy Detector with answer-level verification: the detector monitors the reasoning trajectory and flags candidate exit points when the current step appears semantically redundant with recent context, while verification checks whether the trial answer is stable and sufficiently confident before stopping. This design decouples \emph{where} to consider stopping from \emph{whether} the candidate exit is reliable, allowing PUMA to avoid relying on answer-level readiness alone. To instantiate the detector, we fine-tune Qwen3-Embedding-0.6B~\cite{qwen3embedding} with a contrastive objective~\cite{oord2018representation}, training it to distinguish steps that introduce new logical or semantic progress from those that merely restate, re-derive, or loop over prior content. This design also makes PUMA naturally semantic-preserving: because it favors stopping after meaningful exploration, the retained chain is more likely to form a coherent problem-solving narrative rather than cut mid-exploration.

Across five LRMs from diverse model families, including DeepSeek-R1-Distill~\cite{guo2025deepseek}, Llama-Nemotron~\cite{bercovich2025llamanemotron}, and Qwen3~\cite{yang2025qwen3}, and five challenging reasoning benchmarks spanning MATH-500~\cite{hendrycks2021measuring}, AIME24/25~\cite{aime24,aime25}, OlympiadBench~\cite{he-etal-2024-olympiadbench}, and GPQA-Diamond~\cite{rein2024gpqa}, PUMA achieves 26.2\% average token reduction while preserving final-answer accuracy, with token savings translating into practical wall-clock speedups. Importantly, PUMA also preserves retained CoT quality, indicating that its efficiency gains do not come at the cost of a coherent, semantically complete reasoning chain. Additional experiments on code generation, zero-shot vision-language reasoning, and models trained to internalize PUMA-selected exit positions further show that reasoning-level semantic redundancy is a robust, transferable, and learnable signal for efficient reasoning.

\section{Related Work}
\label{sec:related_work}

\textbf{Overthinking and efficient reasoning in LRMs.}
Overthinking refers to the phenomenon where extended CoT reasoning, despite improving performance, often generates more tokens than necessary~\cite{chen2025do,sui2025stop,su2025between,ghosal2026does,wu2025when}. Wei et al.~\cite{wei2026evolutionthoughttrackingllm} characterize this behavior as a transition from active reasoning to a converged phase in which later tokens are largely redundant. Existing efficient reasoning methods intervene at different stages. \textbf{Training-based methods} reshape reasoning through length-penalized RL~\cite{aggarwal2025l, hou2026thinkprune, dai2026sgrpo}, compressed-chain distillation~\cite{zhang-etal-2025-lightthinker, li2026making}, or latent-space reasoning~\cite{hao2025training, he2025semcot}, but require per-model retraining~\cite{ye2025limo,yu2026dapo,brantley2026accelerating}. \textbf{Prompt-based methods} encourage concise reasoning through length budgets or complexity-aware allocation~\cite{xu2025chain, 10852493, lin2026plan}, but such constraints can be ignored on difficult problems or suppress necessary intermediate reasoning~\cite{yu2025premise}. PUMA instead focuses on inference-time early exit, reducing redundant continuation without modifying model weights or prompts.

\textbf{Inference-time early exit.}
Inference-time early-exit methods differ mainly in the signal used to decide when to stop. \emph{Answer-level signals} monitor trial-answer confidence~\cite{yang2026dynamic, huang2026efficient} or agreement across consecutive probes~\cite{liu2025answer, fu2025reasoning, mao2025early}. These signals estimate answer readiness but are blind to whether the reasoning trajectory is still making semantic progress~\cite{tian2023just, chhikara2025mind, 10.1145/3618260.3649777}. \emph{Token-level signals} track decoding artifacts such as the rank of the \texttt{</think>} token, exit-associated neurons, or reflection-trigger words~\cite{wei2026evolutionthoughttrackingllm, liu2026neat, wang-etal-2025-wait}, while \emph{representation-level signals} train hidden-state probes to predict answer correctness~\cite{akgul2025lynx, zhang2025reasoning}. These approaches are often tied to model-specific delimiters, vocabularies, hidden states, or calibration procedures. 

\section{Methodology}

\begin{figure*}[t]
    \centering
    \vspace{-0.25cm}
    \includegraphics[width=0.95\textwidth]{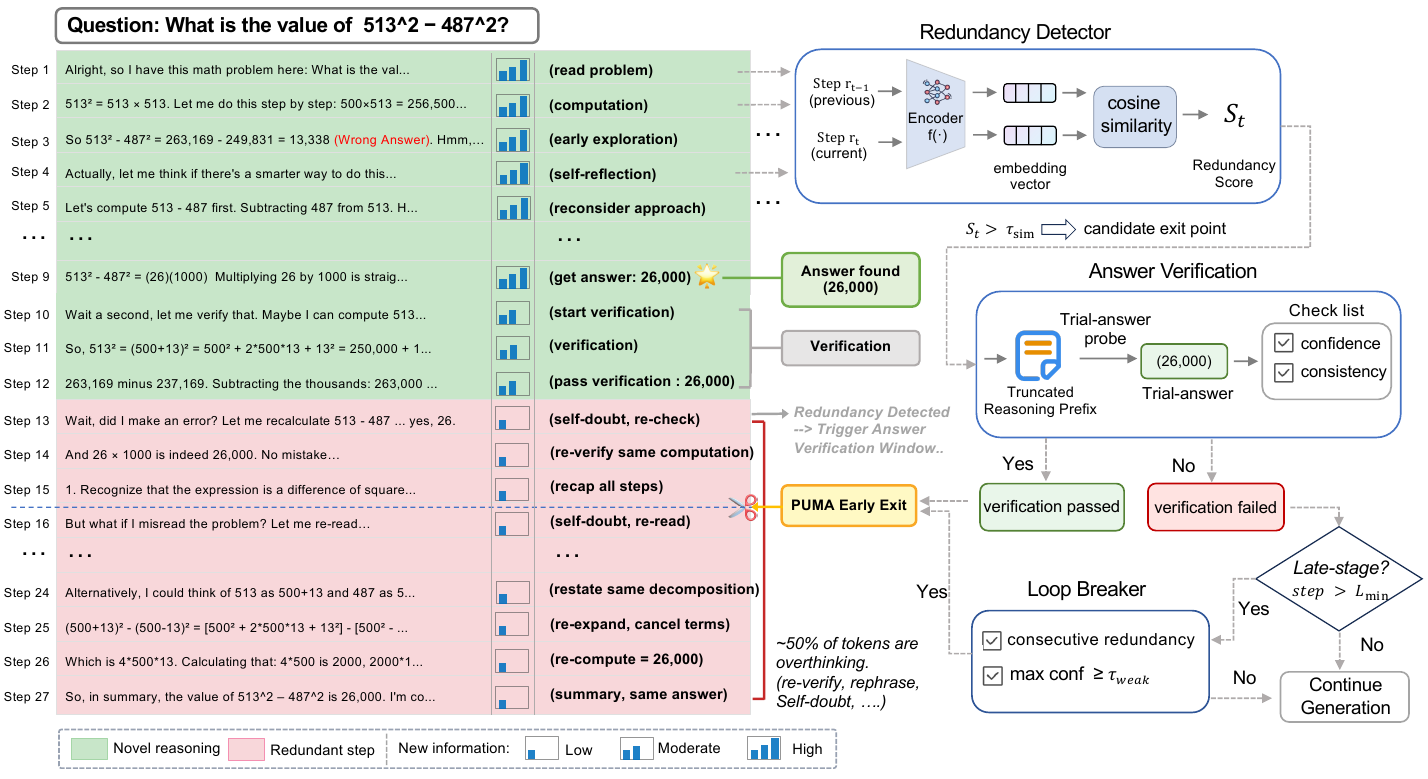}
\caption{Overview of PUMA. The Redundancy Detector compares recent step embeddings and flags candidate exits. Answer Verification then probes a trial answer from the truncated prefix, checking confidence and consistency before stopping. If verification fails, generation continues; if late-stage consecutive redundancy persists, the Loop Breaker provides a fallback exit. The left trace illustrates PUMA removing redundant continuation while preserving the reasoning prefix.}
    \label{fig:puma_overview}
    \vspace{-0.45cm}
\end{figure*}

\textbf{Preliminaries}
Given an input question, a reasoning model generates a reasoning chain $R=(r_1,r_2,\ldots,r_T)$ followed by a final answer $A$, where each $r_t$ denotes a segmented reasoning step. Step segmentation is performed at natural paragraph or step boundaries, with details in Appendix~\ref{appendix:step_segmentation}. 
PUMA operates online over this step sequence, using reasoning-level redundancy to identify candidate exit points where the trajectory appears to have begun converging, i.e., where recent steps mostly revisit established conclusions rather than add meaningful logical or semantic progress. Because reasoning convergence alone does not imply answer readiness, PUMA exits after answer-level verification checks that the trial answer is stable and sufficiently confident.
The goal is to reduce unnecessary tokens while preserving final-answer correctness and a coherent reasoning prefix.

\textbf{Overview of PUMA}
Figure~\ref{fig:puma_overview} illustrates PUMA. PUMA is a plug-and-play early-exit framework that operates during decoding without modifying model weights or the base prompt. It consists of a primary \emph{Verified Early Exit} path and a \emph{Loop Breaker} fallback. Verified Early Exit follows a two-stage design: the Redundancy Detector monitors the reasoning trajectory and flags candidate exit points when the current step appears semantically redundant with recent context; Answer Verification is invoked only at these detector-flagged candidates to check whether the trial answer is stable and sufficiently confident. This separates \emph{where} to consider stopping from \emph{whether} the candidate exit is reliable. The Loop Breaker is activated only when no verified exit occurs and the model produces consecutive redundant steps in the later stages of reasoning.

\textbf{Redundancy Detector}
PUMA implements the Redundancy Detector as a lightweight embedding model trained for reasoning-step redundancy detection. 
We initialize it from Qwen3-Embedding-0.6B~\cite{qwen3embedding} and fine-tune it with LoRA~\cite{hu2022lora} using an InfoNCE contrastive objective~\cite{oord2018representation,pmlr-v139-radford21a} to distinguish steps that introduce new logical or semantic progress from those that restate, re-derive, or loop over prior content. This task-specific training is important because generic semantic similarity does not directly capture reasoning-step novelty: two steps may be topically similar while still advancing the solution, or lexically different while repeating the same verification. Details on supervision construction, labeling prompts, and training hyperparameters are provided in Appendix~\ref{appendix:rd_training}.

Given the trained detector $f(\cdot)$, PUMA scores the redundancy of step $r_t$ by comparing it with the previous k reasoning steps:
\begin{equation}
s_t^{(k)} =
\max_{\max(1,\,t-k)\leq j<t}
\cos\bigl(f(r_j), f(r_t)\bigr).
\end{equation}
A higher score indicates that the current step is semantically close to recent context and is less likely to add novel reasoning progress. PUMA flags $r_t$ as a candidate exit when $s_t^{(k)} > \tau_{\mathrm{sim}}$. By default, we use k=1, comparing each step only with its immediate predecessor, which provides a conservative local redundancy signal. 
Sensitivity to larger lookback windows is analyzed in Appendix~\ref{appendix:rd_window}; comparisons between our embedding-based detector and Natural Language Inference (NLI)-based~\cite{maccartney2009natural} redundancy signals are provided in Appendix~\ref{appendix:rd_signal}.
Conceptually, this detector serves as a lightweight online proxy for the local semantic collapse that semantic entropy would capture through explicit clustering; we discuss this perspective in Appendix~\ref{appendix:se_perspective}.

\textbf{Answer Verification.}
A detector flag is not itself a stopping decision. Once a candidate point $t$ is flagged, PUMA appends a task-specific answer-inducing suffix to the current reasoning prefix $r_{\leq t}$ and probes the model to produce a trial answer $A_t$. The associated confidence $C_t$ is computed as the geometric mean of token probabilities~\cite{yang2026dynamic} over the generated answer tokens:
\begin{equation}
C_t = \exp\!\left(\frac{1}{n}\sum_{i=1}^{n}
\log p(a_i^t \mid a_{<i}^t, \mathbf{c}_t)\right),
\end{equation}
where $\mathbf{c}_t$ denotes the prefix plus the answer-inducing suffix.
Given the first detector-flagged candidate $t_1$, PUMA continues generation until it observes $L-1$ additional detector-flagged candidates, forming a verification window $t_1<t_2<\cdots<t_L$. At each candidate $t_\ell$, PUMA extracts a trial answer $A_{t_\ell}$ and confidence $C_{t_\ell}$. PUMA exits at the end of the window only if
\begin{equation}
\mathrm{Exit}(t_1,\ldots,t_L)
=
[C_{t_1}> \lambda]
\wedge
\left[\bigwedge_{\ell=2}^{L} A_{t_\ell}=A_{t_1}\right]
\wedge
\left[\bigwedge_{\ell=2}^{L} C_{t_\ell}\geq C_{t_1}-\epsilon\right],
\end{equation}
where \(\lambda\) is the confidence threshold and \(\epsilon\) is the stability tolerance.
These checks require the candidate answer to be confident, consistent across redundancy-triggered probes, and not materially declining in confidence. If any condition fails, PUMA resumes generation until the next flagged candidate.

\textbf{Loop Breaker.}
Verified Early Exit is the primary stopping mechanism, but some trajectories produce consecutive redundant steps without satisfying all verification conditions. PUMA therefore includes a late-stage Loop Breaker fallback. After the reasoning chain exceeds a minimum step threshold, if the Redundancy Detector identifies $m$ consecutive redundant steps, PUMA checks whether the highest-confidence trial answer observed so far exceeds a weak minimum-confidence gate. If the gate is satisfied, PUMA terminates generation; otherwise, generation continues. Verified exits take precedence, so the Loop Breaker is used only when no verified exit occurs.
\section{Experimental Setup}
\label{sec:experimental_setup}

\textbf{Benchmarks and models.}\quad
We evaluate PUMA on five challenging reasoning benchmarks covering competition mathematics, olympiad-level STEM reasoning, and graduate-level scientific reasoning: MATH-500~\cite{hendrycks2021measuring}, AIME24/25~\cite{aime24,aime25}, OlympiadBench~\cite{he-etal-2024-olympiadbench}, and GPQA-Diamond~\cite{rein2024gpqa}. Our main experiments use five LRMs from diverse model families and scales: DeepSeek-R1-Distill-Qwen-7B/14B/32B~\cite{guo2025deepseek}, Llama-3.1-Nemotron-Nano-8B~\cite{bercovich2025llamanemotron}, and Qwen3-30B-A3B-Thinking~\cite{yang2025qwen3}. This suite covers 7B--32B-scale models, dense and mixture-of-experts architectures, and both mathematical and scientific reasoning tasks. Dataset statistics are provided in Appendix~\ref{appendix:dataset_stats}.

\textbf{Baselines.}\quad
We focus our main comparisons on deployment-compatible efficiency methods that do not modify model weights or require training model-specific hidden-state probes, and report \textbf{Full-CoT} as the unmodified generation reference.
\emph{Prompt-based:} No-Think~\cite{ma2025reasoning} prompts the model to skip reasoning entirely;
Concise CoT (CCoT;~\cite{nayab2024concise}) imposes a global word budget;
Chain of Draft (CoD;~\cite{xu2025chain}) enforces per-step word limits;
Plan-and-Budget~\cite{lin2026plan} decomposes problems and allocates reasoning budgets
by complexity.
\emph{Inference-time early exit:}
Answer Convergence (Ans.\ Conv.;~\cite{liu2025answer}) stops when consecutive trial
answers agree;
Dynasor~\cite{fu2025reasoning} combines answer consistency with certainty probing;
DEER~\cite{yang2026dynamic} halts when a trial answer exceeds a confidence threshold.
All baselines are reproduced using their official code repositories where available, or following the procedures described in the original papers.

\textbf{Metrics.}\quad
For the main experiments, we report accuracy (\textit{Acc}), average generated tokens
(\textit{Tok}), and token reduction \(\textit{TR}= (1-\textit{Tok}_{\text{method}}/
\textit{Tok}_{\text{Full-CoT}})\times 100\). Higher \textit{TR} indicates greater
efficiency. Overall results follow the benchmark-level reporting convention for accuracy,
while token-reduction statistics are weighted by benchmark size to reflect aggregate token
savings.
We report accuracy and token reduction jointly, since token savings are meaningful only when answer
quality is preserved. For latency experiments, we additionally report wall-clock speedup relative to Full-CoT.
Unless otherwise stated, token counts include all generated tokens incurred by a method, including main reasoning tokens, trial-answer tokens, and final-answer tokens, so TR reflects total generation cost rather than only the retained output length.

\textbf{Implementation Details.}\quad
All experiments are conducted on a single node with 4$\times$NVIDIA GH200 Grace Hopper superchips. We use vLLM for language-model inference and serve the Redundancy Detector on the same node. Unless otherwise specified, we use the models' recommended reasoning settings (\texttt{temperature}=0.6, \texttt{$top_p$}=0.95) and report results averaged over three random seeds (0, 42, 123). PUMA uses conservative stopping hyperparameters to favor high-precision exits: for the Redundancy Detector, local window size $k=1$ and similarity threshold $\tau_{\mathrm{sim}}=0.35$, selected on a held-out detector calibration split (Appendix~\ref{appendix:rd_training}); for Answer Verification, confidence threshold $\lambda=0.98$, stability tolerance $\epsilon=0.03$, and verification window length $L=2$; and for the Loop Breaker, a late-stage activation threshold of 50 reasoning steps and a weak minimum-confidence gate of 0.8. Sensitivity to these choices is analyzed in Appendix~\ref{appendix:hyperparameters}.

\section{Experimental Results}

\subsection{Main Results: Accuracy, Efficiency, and Reasoning Quality}
\label{sec:main_results}
\label{sec:reasoning_quality}

\begin{table*}[t!]
\centering
\caption{Performance comparison of PUMA against baselines on five benchmarks across
three representative reasoning models. Acc (\%, $\uparrow$): accuracy. TR (\%, $\uparrow$):
token reduction relative to Full CoT; negative TR values (shown in \textcolor{darkred}{red})
indicate higher token usage than Full CoT. Full results for all five models, including
per-benchmark token counts, are reported in Table~\ref{tab:full_results}.}
\label{tab:main_results}
\begin{adjustbox}{width=0.9\textwidth}
\tiny
\renewcommand{\arraystretch}{0.9}
\setlength{\tabcolsep}{3pt}
\begin{tabular}{l cc cc cc cc cc | cc}
\toprule
& \multicolumn{2}{c}{\textbf{MATH-500}}
& \multicolumn{2}{c}{\textbf{AIME24}}
& \multicolumn{2}{c}{\textbf{AIME25}}
& \multicolumn{2}{c}{\textbf{GPQA-D}}
& \multicolumn{2}{c}{\textbf{OlymBench}}
& \multicolumn{2}{c}{\textbf{Overall}} \\
\cmidrule(lr){2-3} \cmidrule(lr){4-5} \cmidrule(lr){6-7} \cmidrule(lr){8-9} \cmidrule(lr){10-11} \cmidrule(lr){12-13}
Method
& Acc$\uparrow$ & TR$\uparrow$
& Acc$\uparrow$ & TR$\uparrow$
& Acc$\uparrow$ & TR$\uparrow$
& Acc$\uparrow$ & TR$\uparrow$
& Acc$\uparrow$ & TR$\uparrow$
& Acc$\uparrow$ & TR$\uparrow$ \\
\midrule
\multicolumn{13}{l}{\textbf{DeepSeek-R1-Distill-Qwen-7B}} \\
\rowcolor{methodgray!47}
Full CoT        & 90.0 & 0.0 & 50.0 & 0.0 & 43.3 & 0.0 & 49.0 & 0.0 & 57.6 & 0.0 & 58.0 & 0.0 \\
No-Think        & 79.0 & 80.0 & 20.0 & 67.7 & 23.3 & 57.9 & 28.3 & 87.3 & 42.8 & 79.1 & 38.7 & 79.9 \\
CCoT            & 85.0 & 60.9 & 36.7 & 59.6 & 33.3 & 60.2 & 48.0 & 55.6 & 49.2 & 57.7 & 50.4 & 58.6 \\
CoD             & 80.4 & 59.9 & 40.0 & 56.9 & 33.3 & 58.7 & 50.5 & 50.8 & 43.0 & 55.8 & 49.4 & 56.6 \\
Plan\&Budget    & 84.6 & 48.5 & 43.3 & 52.9 & 23.3 & 53.9 & 38.4 & 36.2 & 49.3 & 50.3 & 47.8 & 47.9 \\
Ans.\ Conv.     & 61.8 & 76.5 & 26.7 & 83.3 & 20.0 & 82.8 & 34.3 & 92.4 & 32.6 & 82.8 & 35.1 & 81.9 \\
Dynasor         & 84.2 & 30.6 & 23.3 & 21.7 & 33.3 & 9.1 & 39.4 & 66.6 & 49.0 & 34.1 & 45.9 & 36.6 \\
DEER            & 90.6 & 39.6 & 40.0 & 14.1 & 50.0 & 13.2 & 51.5 & \textcolor{darkred}{-17.4} & 57.5 & 20.2 & 57.9 & 21.5 \\
\rowcolor{pumagreen!25}
PUMA (ours)     & 89.6 & 38.6 & 60.0 & 30.3 & 46.7 & 29.8 & 49.0 & 4.0 & 55.6 & 43.2 & 60.2 & 35.6 \\
\midrule
\multicolumn{13}{l}{\textbf{Llama-3.1-Nemotron-Nano-8B}} \\
\rowcolor{methodgray!47}
Full CoT        & 93.6 & 0.0 & 66.7 & 0.0 & 50.0 & 0.0 & 48.0 & 0.0 & 63.1 & 0.0 & 64.3 & 0.0 \\
No-Think        & 62.8 & \textcolor{darkred}{-125.5} & 26.7 & \textcolor{darkred}{-32.3} & 16.7 & \textcolor{darkred}{-43.4} & 25.8 & \textcolor{darkred}{-2.2} & 31.9 & \textcolor{darkred}{-74.0} & 32.8 & \textcolor{darkred}{-80.5} \\
CCoT            & 86.6 & 34.5 & 36.7 & 36.0 & 20.0 & 36.1 & 43.9 & 31.1 & 56.9 & 34.6 & 48.8 & 34.1 \\
CoD             & 79.0 & 30.1 & 50.0 & 40.9 & 33.3 & 39.4 & 45.0 & 31.9 & 51.3 & 32.0 & 51.7 & 31.7 \\
Plan\&Budget    & 89.0 & 8.3 & 43.3 & 32.9 & 33.3 & 31.6 & 40.9 & 26.7 & 53.9 & 22.7 & 52.1 & 18.6 \\
Ans.\ Conv.     & 55.8 & 76.3 & 6.7 & 91.9 & 3.3 & 90.4 & 20.2 & 95.3 & 23.6 & 85.1 & 21.9 & 83.7 \\
Dynasor         & 91.4 & \textcolor{darkred}{-9.0} & 50.0 & \textcolor{darkred}{-1.0} & 43.3 & \textcolor{darkred}{-14.4} & 45.0 & 1.5 & 61.9 & \textcolor{darkred}{-3.0} & 58.3 & \textcolor{darkred}{-4.7} \\
DEER            & 91.8 & 22.7 & 53.3 & \textcolor{darkred}{-22.5} & 50.0 & \textcolor{darkred}{-35.9} & 49.5 & \textcolor{darkred}{-313.2} & 61.2 & 12.4 & 61.2 & \textcolor{darkred}{-30.7} \\
\rowcolor{pumagreen!25}
PUMA (ours)     & 92.6 & 15.7 & 70.0 & 12.3 & 50.0 & 18.1 & 48.5 & 9.3 & 62.2 & 27.0 & 64.7 & 20.1 \\
\midrule
\multicolumn{13}{l}{\textbf{Qwen3-30B-A3B-Thinking}} \\
\rowcolor{methodgray!47}
Full CoT        & 94.4 & 0.0 & 83.3 & 0.0 & 83.3 & 0.0 & 72.7 & 0.0 & 75.0 & 0.0 & 81.7 & 0.0 \\
No-Think        & 91.8 & 51.7 & 60.0 & 56.5 & 50.0 & 50.0 & 73.7 & \textcolor{darkred}{-11.5} & 68.4 & 56.6 & 68.8 & 45.3 \\
CCoT            & 89.2 & 54.6 & 36.7 & 55.9 & 20.0 & 60.3 & 72.7 & 45.5 & 53.0 & 58.8 & 54.3 & 55.5 \\
CoD             & 82.8 & 29.9 & 23.3 & 52.5 & 10.0 & 57.8 & 65.2 & 28.8 & 45.3 & 50.3 & 45.3 & 40.4 \\
Plan\&Budget    & 91.8 & 51.4 & 43.3 & 56.6 & 36.7 & 60.5 & 64.7 & 37.1 & 61.3 & 60.2 & 59.6 & 53.9 \\
Ans.\ Conv.     & 57.0 & 87.8 & 0.0 & 94.4 & 0.0 & 90.6 & 35.9 & 95.5 & 23.1 & 91.7 & 23.2 & 90.9 \\
Dynasor         & 94.8 & \textcolor{darkred}{-4.3} & 86.7 & 12.1 & 76.7 & 2.8 & 67.7 & 32.2 & 72.4 & \textcolor{darkred}{-0.6} & 79.7 & 3.0 \\
DEER            & 94.8 & 34.9 & 83.3 & 10.5 & 80.0 & 7.3 & 74.8 & \textcolor{darkred}{-8.2} & 73.5 & 23.4 & 81.3 & 22.4 \\
\rowcolor{pumagreen!25}
PUMA (ours)     & 94.2 & 28.8 & 90.0 & 21.0 & 80.0 & 14.7 & 75.8 & 17.8 & 72.3 & 31.7 & 82.5 & 28.2 \\
\bottomrule
\end{tabular}
\end{adjustbox}
\vspace{-0.4cm}
\end{table*}

Table~\ref{tab:main_results} summarizes the accuracy--efficiency tradeoff on three representative LRMs, with full results across all five LRMs reported in Appendix~\ref{appendix:full_results}. Across five LRMs and five benchmarks, PUMA achieves \textbf{26.2}\% average token reduction while preserving accuracy. This indicates that PUMA removes redundant continuation without sacrificing final-answer quality. The slight average accuracy gain over Full CoT is consistent with the overthinking phenomenon: after reaching a correct answer, LRMs may continue re-verifying or revising and occasionally drift to an incorrect conclusion.

Compared with prompt-based compression, PUMA is more accuracy-preserving. Although CCoT, CoD, and Plan-and-Budget often reduce token usage, their brevity constraints can suppress necessary intermediate reasoning, especially on stronger reasoning models. For example, on Qwen3-30B-A3B-Thinking, their accuracies drop to 54.3\%, 45.3\%, and 59.6\%, respectively, compared with 81.7\% for Full CoT. Compared with answer-level early-exit baselines, PUMA provides a more reliable accuracy--efficiency tradeoff: Answer Convergence stops aggressively but suffers severe accuracy collapse, while Dynasor and DEER show inconsistent efficiency due to repeated trial-answer probing. In contrast, PUMA invokes answer verification only after reasoning-level redundancy is detected.

\begin{table}[b]
\vspace{-0.5cm}
\centering
\caption{Reasoning-chain quality evaluated by an LLM-as-Judge (GPT-5.4-thinking). Individual chains are rated on a 10--100 rubric in 10-point increments, and the higher is better.}
\label{tab:reasoning_quality}
\vspace{0.2cm}
\resizebox{0.82\textwidth}{!}{
\renewcommand{\arraystretch}{1.05}
\setlength{\tabcolsep}{6pt}
\begin{tabular}{l >{\columncolor{methodgray!30}}c >{\columncolor{pumagreen!25}}c c c c c c}
\toprule
\textbf{Metric}
& \textbf{Full-CoT}
& \textbf{PUMA}
& \textbf{CCoT}
& \textbf{Plan\&Budget}
& \textbf{CoD}
& \textbf{Dynasor}
& \textbf{DEER} \\
\midrule
Completeness$\uparrow$
& \textbf{67.6}
& \underline{64.8}
& 64.1
& 62.5
& 62.0
& 57.2
& 62.4 \\
Coherence$\uparrow$
& 41.5
& \textbf{57.5}
& \underline{51.6}
& 48.6
& 49.0
& 46.9
& 37.9 \\
Conciseness$\uparrow$
& 15.7
& \textbf{40.2}
& \underline{30.5}
& 27.2
& 26.1
& 25.0
& 19.1 \\
Justification$\uparrow$
& 51.8
& \textbf{54.8}
& \underline{54.5}
& 51.6
& 50.3
& 46.5
& 46.7 \\
\midrule
Avg.$\uparrow$
& 44.1
& \textbf{54.3}
& \underline{50.2}
& 47.5
& 46.8
& 43.9
& 41.5 \\
\bottomrule
\end{tabular}
}
\vspace{-0.3cm}
\end{table}

Beyond final-answer accuracy, early exit should preserve the retained reasoning chain as a useful explanation. Following the LLM-as-Judge protocol~\cite{li-etal-2025-generation, li2024mateval, zhang2024dee}, Table~\ref{tab:reasoning_quality} evaluates retained CoT quality along completeness, coherence, conciseness, and justification. PUMA achieves the highest average retained-chain quality among compared methods, ranking first in coherence, conciseness, and justification while remaining close to Full CoT in completeness. This supports the semantic-preserving nature of PUMA: it shortens reasoning by removing redundant continuation rather than aggressively cutting the chain mid-development. Details of the evaluation setup and judge prompt are provided in Appendix~\ref{appendix:quality_eval}.

\begin{figure*}[t]
\centering
\includegraphics[width=0.98\textwidth]{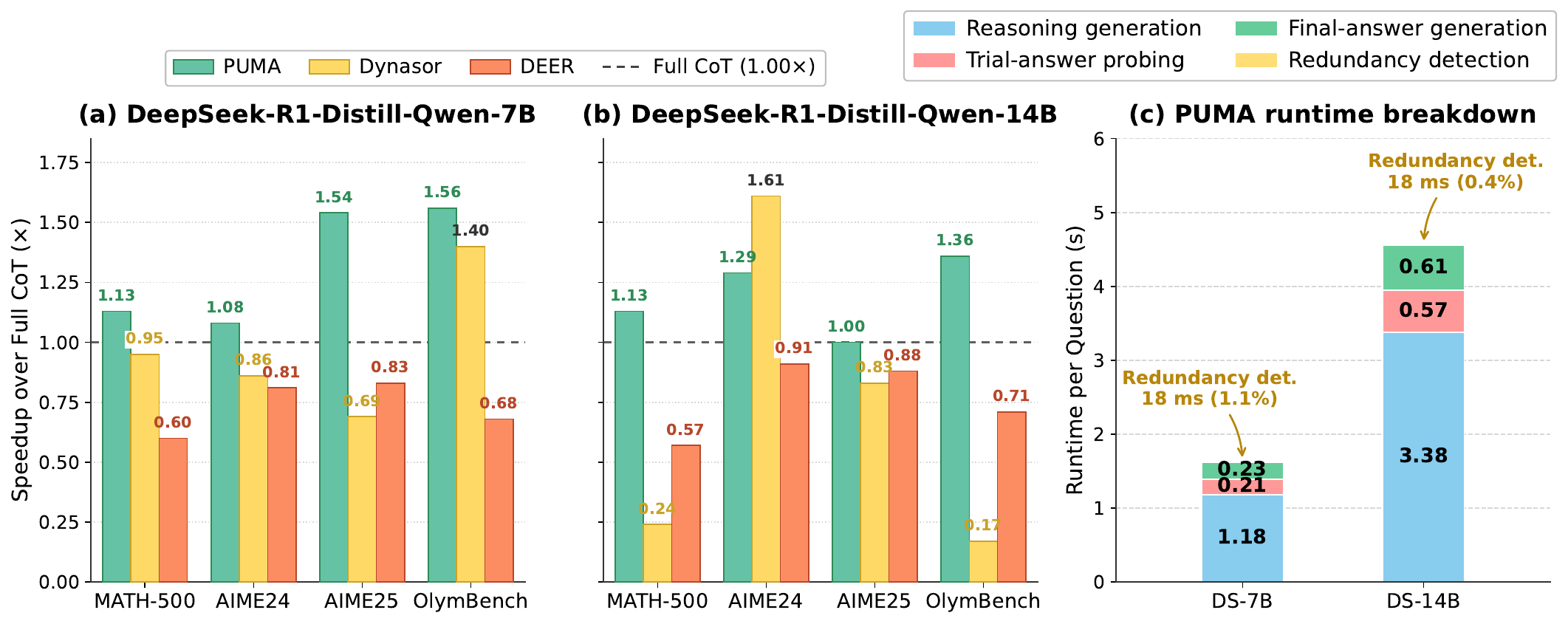}
\caption{Wall-clock latency on DS-7B and DS-14B.
(a,b) Per-benchmark speedup relative to Full-CoT.
(c) PUMA runtime breakdown: the Redundancy Detector adds only 0.4--1.1\% overhead.
Detailed per-benchmark results, including additional models, are reported in Appendix~\ref{appendix:latency_detailed}.}
\label{fig:latency}
\vspace{-0.4cm}
\end{figure*}

\subsection{From Token Savings to Latency Gains}
\label{sec:latency}
Token reduction does not automatically translate to wall-clock speedup, because early-exit methods incur overhead from trial-answer probing. Figure~\ref{fig:latency} shows that PUMA turns token savings into practical speedups, achieving 1.40× speedup on DS-7B and 1.28× on DS-14B on average. By contrast, DEER is slower than Full CoT on both models, while Dynasor is faster on DS-7B but substantially slower on DS-14B, highlighting the overhead of frequent answer-level probing. The runtime breakdown in Figure~\ref{fig:latency}(c) further shows that PUMA's overhead is small: the Redundancy Detector contributes only 0.4--1.1\% of total wall-clock time per question, and the Answer Verification overhead from trial-answer probing is only 0.2--0.57 seconds per question. PUMA keeps overhead low by using a lightweight detector and probing answers only at detector-flagged candidates.

\subsection{Generalization Beyond Text-only QA}
\label{sec:generalization}

Vision-language reasoning introduces unique challenges beyond text-only settings, as models must handle cross-modal interactions~\cite{Li_2024_CVPR, chen2025mvi, chen2025enhancing}.
We further test whether reasoning-level redundancy transfers beyond text-only mathematical and scientific QA. Table~\ref{tab:generalization} evaluates PUMA on code generation with LiveCodeBench~\cite{jain2025livecodebench} and vision-language reasoning with MathVista~\cite{lu2024mathvista} and MathVision~\cite{wang2024measuring}. On LiveCodeBench, PUMA reduces tokens by 18--19\% with at most 1.5 points of pass@1 change. On vision-language reasoning, PUMA is applied without retraining or hyperparameter tuning, yet still reduces tokens by 23.8--33.6\% with at most 1.5 points of accuracy change. These results suggest that reasoning-level redundancy is a transferable stopping signal across code, text, and multimodal reasoning. Additional prompt-budget sweeps for CCoT and CoD are provided in Appendix~\ref{appendix:baseline_tuning}.

\begin{table*}[h]
\centering
\caption{Generalization to code and vision-language reasoning. Code generation is evaluated on LiveCodeBench (880 problems, pass@1); for PUMA, only the redundancy threshold is adjusted to $\tau_{\mathrm{sim}}=0.50$. Vision-language reasoning is evaluated on MathVista and MathVision (200 problems each), where PUMA is applied zero-shot without retraining or hyperparameter tuning.}

\label{tab:generalization}
\begin{adjustbox}{width=0.98\textwidth}
\renewcommand{\arraystretch}{1.05}
\setlength{\tabcolsep}{4pt}
\footnotesize
\begin{tabular}{l cc cc cc cc cc cc}
\toprule
& \multicolumn{4}{c}{\textbf{Code (LiveCodeBench)}}
& \multicolumn{4}{c}{\textbf{VL: Qwen3-VL-8B-Thinking}}
& \multicolumn{4}{c}{\textbf{VL: Kimi-VL-16B-A3B-Thinking}} \\
\cmidrule(lr){2-5} \cmidrule(lr){6-9} \cmidrule(lr){10-13}
& \multicolumn{2}{c}{\textbf{DS-7B}}
& \multicolumn{2}{c}{\textbf{Nemotron-8B}}
& \multicolumn{2}{c}{\textbf{MathVista}}
& \multicolumn{2}{c}{\textbf{MathVision}}
& \multicolumn{2}{c}{\textbf{MathVista}}
& \multicolumn{2}{c}{\textbf{MathVision}} \\
\cmidrule(lr){2-3} \cmidrule(lr){4-5} \cmidrule(lr){6-7} \cmidrule(lr){8-9}
\cmidrule(lr){10-11} \cmidrule(lr){12-13}
\textbf{Method}
& Acc ($\Delta$) & TR$\uparrow$
& Acc ($\Delta$) & TR$\uparrow$
& Acc ($\Delta$) & TR$\uparrow$
& Acc ($\Delta$) & TR$\uparrow$
& Acc ($\Delta$) & TR$\uparrow$
& Acc ($\Delta$) & TR$\uparrow$ \\
\midrule
\rowcolor{methodgray!30}
Full CoT
& 51.8 (\textemdash) & 0.0
& 60.2 (\textemdash) & 0.0
& 76.0 (\textemdash) & 0.0
& 40.0 (\textemdash) & 0.0
& 67.5 (\textemdash) & 0.0
& 24.0 (\textemdash) & 0.0 \\
\rowcolor{pumagreen!25}
\textbf{PUMA (ours)}
& \textbf{50.3} (\textcolor{darkred}{-1.5}) & \textbf{19.0}
& \textbf{59.7} (\textcolor{darkred}{-0.5}) & \textbf{18.0}
& \textbf{75.0} (\textcolor{darkred}{-1.0}) & \textbf{23.8}
& \textbf{38.5} (\textcolor{darkred}{-1.5}) & \textbf{31.2}
& \textbf{68.0} (+0.5) & \textbf{28.7}
& \textbf{24.0} (+0.0) & \textbf{33.6} \\
\midrule
DEER
& 49.6 (\textcolor{darkred}{-2.2}) & 1.6
& 52.6 (\textcolor{darkred}{-7.6}) & 19.3
& 71.0 (\textcolor{darkred}{-5.0}) & 16.1
& 36.5 (\textcolor{darkred}{-3.5}) & 54.3
& 53.0 (\textcolor{darkred}{-14.5}) & 11.5
& 22.5 (\textcolor{darkred}{-1.5}) & \textcolor{darkred}{-1.3} \\
CCoT
& 45.8 (\textcolor{darkred}{-6.0}) & 42.7
& 49.3 (\textcolor{darkred}{-10.9}) & 41.6
& 72.5 (\textcolor{darkred}{-3.5}) & 11.2
& 36.0 (\textcolor{darkred}{-4.0}) & 26.4
& 53.0 (\textcolor{darkred}{-14.5}) & 71.2
& 21.5 (\textcolor{darkred}{-2.5}) & 25.0 \\
CoD
& 43.2 (\textcolor{darkred}{-8.6}) & 40.5
& 48.9 (\textcolor{darkred}{-11.3}) & 41.8
& 32.5 (\textcolor{darkred}{-43.5}) & 60.9
& 24.5 (\textcolor{darkred}{-15.5}) & 58.6
& 37.5 (\textcolor{darkred}{-30.0}) & 38.3
& 19.5 (\textcolor{darkred}{-4.5}) & 42.2 \\
Plan\&Budget
& 39.2 (\textcolor{darkred}{-12.6}) & 31.8
& 50.6 (\textcolor{darkred}{-9.6}) & 28.4
& 62.5 (\textcolor{darkred}{-13.5}) & 1.0
& 30.0 (\textcolor{darkred}{-10.0}) & 37.5
& 57.0 (\textcolor{darkred}{-10.5}) & 24.3
& 22.5 (\textcolor{darkred}{-1.5}) & 29.1 \\
Dynasor
& 43.0 (\textcolor{darkred}{-8.8}) & \textcolor{darkred}{-207.6}
& 45.5 (\textcolor{darkred}{-14.7}) & \textcolor{darkred}{-7.8}
& 64.0 (\textcolor{darkred}{-12.0}) & 61.8
& 36.0 (\textcolor{darkred}{-4.0}) & 48.4
& 40.5 (\textcolor{darkred}{-27.0}) & 48.3
& 19.5 (\textcolor{darkred}{-4.5}) & 92.6 \\
No-Think
& 29.1 (\textcolor{darkred}{-22.7}) & 88.1
& 47.6 (\textcolor{darkred}{-12.6}) & 13.1
& 56.0 (\textcolor{darkred}{-20.0}) & 63.9
& 29.5 (\textcolor{darkred}{-10.5}) & 73.1
& 61.5 (\textcolor{darkred}{-6.0}) & 70.1
& 18.5 (\textcolor{darkred}{-5.5}) & 68.3 \\
\bottomrule
\end{tabular}
\end{adjustbox}
\vspace{-0.3cm}
\end{table*}

\section{Analysis and Discussion}
\label{sec:analysis}

\subsection{Component and Exit Behavior Analysis}
\label{sec:component_exit_analysis}

Table~\ref{tab:ablation} ablates PUMA's candidate-triggering and verification components on DS-7B, with full two-model results and probe statistics provided in Appendix~\ref{appendix:ablation}. The full configuration achieves the best accuracy--efficiency balance, reaching 60.2\% accuracy with 35.6\% token reduction. Removing the Redundancy Detector gate sets the redundancy threshold to zero, causing Answer Verification to be invoked at every step rather than only at redundancy-triggered candidate points. This aggressive variant increases token reduction to 46.0\%, but drops accuracy by 4.1 points and requires 3.3\(\times\) more trial-answer probes per question. This shows that reasoning-level gating is important for both reliability and probe efficiency. Removing the Loop Breaker reduces token reduction from 35.6\% to 22.6\%, confirming that it mainly serves as an efficiency-oriented fallback for sustained redundancy. The Answer-Verification parts are also important for accuracy preservation: removing Answer Consistency lowers accuracy by 2.9 points, while removing the Confidence Gate causes a larger 6.5-point drop despite higher token reduction. Overall, the ablation shows that PUMA's components are complementary: the Redundancy Detector gate controls when verification is invoked, Answer Verification filters unreliable candidate exits, and the Loop Breaker provides additional savings when reasoning enters sustained redundancy.

\begin{figure*}[t]
\centering
\includegraphics[width=0.95\textwidth]{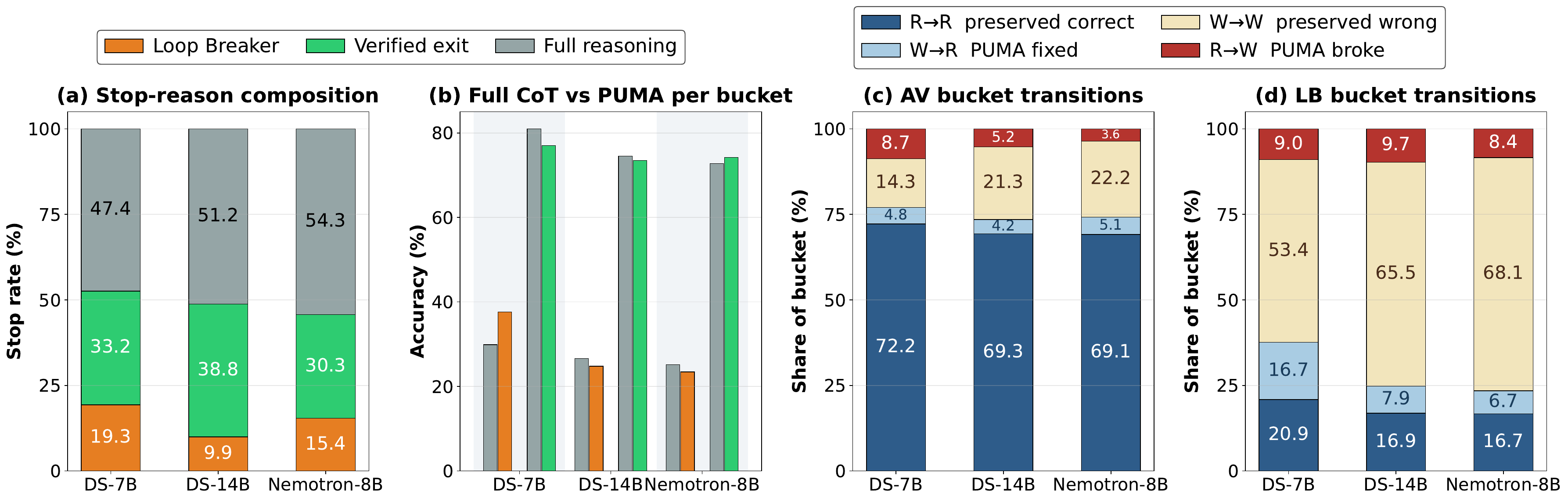}
\caption{Exit behavior of PUMA across three representative LRMs.
\textbf{(a)} Stop-reason composition: Loop Breaker, verified exit, or full reasoning.
\textbf{(b)} Accuracy change of each exit bucket relative to Full CoT.
\textbf{(c,d)} Correctness transitions for verified exits and Loop Breaker exits, where \(R/W\) denote correct/wrong outcomes for Full CoT \(\to\) PUMA.}
\label{fig:exit_behavior}
\vspace{-0.35cm}
\end{figure*}

\begin{wraptable}{r}{0.48\textwidth}
\vspace{-0.4cm}
\centering
\caption{Component ablation on DS-7B, averaged over five benchmarks. Probe\(\times\) denotes trial-answer probes per question relative to PUMA.}
\label{tab:ablation}
\vspace{0.05cm}
\footnotesize
\renewcommand{\arraystretch}{1.05}
\setlength{\tabcolsep}{3pt}
\resizebox{\linewidth}{!}{
\begin{tabular}{l c c c}
\toprule
\textbf{Configuration}
& \textbf{Acc}$\uparrow$
& \textbf{TR}$\uparrow$
& \textbf{Probe}$\times\downarrow$ \\
\midrule
\rowcolor{pumagreen!25}
\textbf{PUMA (full)} & \textbf{60.2} & \textbf{35.6} & \textbf{1.0}\(\times\) \\
\midrule
w/o Redundancy Detector Gate & 56.1 (\textcolor{darkred}{-4.1}) & 46.0 & 3.3\(\times\) \\
w/o Loop Breaker & 59.2 (\textcolor{darkred}{-1.0}) & \textcolor{darkred}{22.6} & 1.3\(\times\) \\
w/o Answer Consistency & 57.3 (\textcolor{darkred}{-2.9}) & 37.8 & 1.2\(\times\) \\
w/o Confidence Gate & 53.7 (\textcolor{darkred}{-6.5}) & 55.3 & 0.6\(\times\) \\
\bottomrule
\end{tabular}
}
\vspace{-0.4cm}
\end{wraptable}

Figure~\ref{fig:exit_behavior} further analyzes when PUMA stops and how different exit modes affect correctness. Figure~\ref{fig:exit_behavior}(a) shows that PUMA does not rely on a single exit mode: verified exits, Loop Breaker exits, and full reasoning occur in different proportions across models, reflecting model-specific redundancy patterns. Figures~\ref{fig:exit_behavior}(c) and~\ref{fig:exit_behavior}(d) further break down correctness transitions for verified exits and Loop Breaker exits, respectively, where \(W\!\to\!R\), for example, denotes a trajectory that is wrong under Full CoT but correct after PUMA stopping. For verified exits, the \(R\!\to\!R\) transition dominates (69--72\%), indicating that Answer Verification usually preserves already-correct trajectories. The \(W\!\to\!R\) cases show that PUMA can also fix some Full-CoT errors by stopping before post-convergence drift, while harmful \(R\!\to\!W\) transitions remain limited. Loop Breaker exits show a different pattern: \(W\!\to\!W\) transitions dominate (53--68\%), meaning that the fallback mostly cuts late redundant continuation from trajectories that Full CoT would also answer incorrectly. Together, these results explain how PUMA gains efficiency with limited accuracy risk, and why it can occasionally outperform Full CoT by avoiding late-stage drift.

\subsection{From Inference-Time Signal to Learned Stopping Policy}
\label{sec:internalization}

\begin{table}[t]

\vspace{-0.2cm}
\centering
\caption{Internalizing PUMA's stopping signal with LoRA on DS-R1-Distill-Qwen-7B. PUMA-selected exit positions supervise SFT, DPO, or GRPO; all trained variants use pure vLLM inference without PUMA modules.}
\label{tab:internalize}
\renewcommand{\arraystretch}{1.0}
\setlength{\tabcolsep}{6pt}
\resizebox{0.67\linewidth}{!}{
\begin{tabular}{l cc cc cc | cc}
\toprule
& \multicolumn{2}{c}{\textbf{MATH-500}}
& \multicolumn{2}{c}{\textbf{AIME24}}
& \multicolumn{2}{c}{\textbf{GPQA-D}}
& \multicolumn{2}{c}{\textbf{Avg.}} \\
\cmidrule(lr){2-3} \cmidrule(lr){4-5} \cmidrule(lr){6-7} \cmidrule(lr){8-9}
\textbf{Method}
& Acc$\uparrow$ & TR$\uparrow$
& Acc$\uparrow$ & TR$\uparrow$
& Acc$\uparrow$ & TR$\uparrow$
& Acc$\uparrow$ & TR$\uparrow$ \\
\midrule
\rowcolor{methodgray!50}
Full CoT (base)        & 90.0 & 0.0  & 50.0 & 0.0  & 49.0 & 0.0  & 63.0 & 0.0 \\
PUMA (inference) & 89.6 & 38.6 & 60.0 & 30.3 & 49.0 & 4.0  & 66.2 & 24.3 \\
\midrule
\multicolumn{9}{l}{\textit{Imitation learning (SFT)}} \\
Standard-SFT           & 91.4 & \textcolor{darkred}{-4.5}  & 43.3 & \textcolor{darkred}{-31.0} & 49.5 & \textcolor{darkred}{-61.9} & 61.4 & \textcolor{darkred}{-32.5} \\
FixedExit-SFT          & 85.8 & 46.3 & 36.7 & 17.2 & 39.9 & 21.3 & 54.1 & 28.3 \\
\rowcolor{pumagreen!15}
\textbf{PUMA-SFT} & 91.4 & 21.9 & 53.3 & 32.7 & 56.1 & 3.3 & \textbf{66.9} & 19.3 \\
\midrule
\multicolumn{9}{l}{\textit{Preference learning (DPO)}} \\
\rowcolor{pumagreen!20}
\textbf{PUMA-DPO} & 90.6 & 39.3 & 56.7 & 59.4 & 45.0 & 47.7 & 64.1 & \textbf{48.8} \\
\midrule
\multicolumn{9}{l}{\textit{Reinforcement learning (GRPO)}} \\
Standard-GRPO          & 91.4 & 1.2  & 50.0 & \textcolor{darkred}{-1.5} & 56.1 & \textcolor{darkred}{-9.8} & 65.8 & \textcolor{darkred}{-3.4} \\
FixedExit-RL           & 91.4 & 41.5 & 46.7 & 38.0 & 51.5 & 32.8 & 63.2 & 37.4 \\
\rowcolor{pumagreen!25}
\textbf{PUMA-RL}  & 90.2 & 42.4 & 56.7 & 35.7 & 54.0 & 26.6 & \textbf{67.0} & 34.9 \\
\bottomrule
\end{tabular}
}
\vspace{-0.5cm}
\end{table}

Finally, we ask whether PUMA-selected exit positions can be internalized into the model itself. We fine-tune DS-R1-Distill-Qwen-7B on 12K math problems using SFT, DPO~\cite{NEURIPS2023_a85b405e}, and GRPO~\cite{shao2024deepseekmath}, with PUMA-selected exits providing stopping supervision. All trained variants are evaluated with pure vLLM inference and use no PUMA modules at deployment. Each paradigm uses PUMA's exit positions differently. 
PUMA-SFT uses PUMA-truncated reasoning traces and their final answers as supervised targets, teaching the model to produce concise reasoning paths that stop near PUMA-selected exit points. PUMA-DPO treats PUMA-truncated chains as preferred over full-length chains when both produce correct answers, teaching the model that shorter-is-better given equal correctness. PUMA-RL uses GRPO on rollouts launched from PUMA's RD-flagged exit positions, with a reward that combines correctness, a length bonus, and a within-group rank bonus that favors the shortest correct trajectory. Detailed data construction and training settings are provided in Appendix~\ref{appendix:internalization_details}.
Table~\ref{tab:internalize} shows that PUMA-selected exits provide useful supervision for concise reasoning. PUMA-RL achieves the strongest overall result, exceeding training-free PUMA in both average accuracy (67.0 vs.\ 66.2) and token reduction (34.9\% vs.\ 24.3\%). PUMA-DPO provides a more compression-oriented alternative, reaching the highest average token reduction (48.8\%) while still improving over the Full CoT baseline in average accuracy (64.1 vs.\ 63.0).
The FixedExit baselines use the same training recipes but replace PUMA-selected positions with fixed-interval candidates. Their weaker performance indicates that PUMA's exit positions are semantically informative rather than merely shorter prefixes: PUMA-SFT improves over FixedExit-SFT by 12.8 accuracy points on average, and PUMA-RL improves over FixedExit-RL by 3.8 points while maintaining comparable token reduction. 
In contrast, Standard-SFT and Standard-GRPO do not learn concise stopping behavior, using more tokens than Full CoT on average (i.e., average token reduction \(\leq 0\)).
These results suggest that reasoning-level redundancy is not only useful for inference-time early exit, but can also be learned as a stopping policy.

\section{Conclusion}

We presented PUMA, a plug-and-play early-exit framework that uses reasoning-level semantic redundancy to identify when a reasoning trajectory appears to have converged, and pairs this signal with answer-level verification for reliable early stopping. Across five LRMs and five challenging reasoning benchmarks, PUMA reduces tokens by 26.2\% on average while preserving accuracy, delivering practical wall-clock speedups, and maintaining higher-quality retained reasoning chains than competing methods. PUMA further transfers to code generation and zero-shot vision-language reasoning, and its selected exit positions can supervise SFT, DPO, and GRPO, enabling models to internalize concise stopping behavior without PUMA modules at inference time. These results suggest that reasoning-level redundancy is a robust, transferable, and learnable signal for efficient reasoning.

\textbf{Limitations.}
PUMA relies on step-level reasoning traces and may be less effective when model outputs are very short, poorly structured, or difficult to segment. Although the Redundancy Detector transfers beyond text-only QA, it is primarily trained on text reasoning traces, so broader domains may require additional calibration. Our internalization experiments are also limited to one base model and math-focused training data, leaving larger-scale and cross-domain learned stopping policies for future work.


\bibliographystyle{unsrtnat}
\bibliography{references}

@article{guo2025deepseek,
  title={DeepSeek-R1 incentivizes reasoning in LLMs through reinforcement learning},
  author={Guo, Daya and Yang, Dejian and Zhang, Haowei and Song, Junxiao and Wang, Peiyi and Zhu, Qihao and Xu, Runxin and Zhang, Ruoyu and Ma, Shirong and Bi, Xiao and others},
  journal={Nature},
  volume={645},
  number={8081},
  pages={633--638},
  year={2025},
  publisher={Nature Publishing Group UK London}
}

@article{jaech2024openai,
  title={Openai o1 system card},
  author={Jaech, Aaron and Kalai, Adam and Lerer, Adam and Richardson, Adam and El-Kishky, Ahmed and Low, Aiden and Helyar, Alec and Madry, Aleksander and Beutel, Alex and Carney, Alex and others},
  journal={arXiv preprint arXiv:2412.16720},
  year={2024}
}

@article{yang2025qwen3,
  title={Qwen3 technical report},
  author={Yang, An and Li, Anfeng and Yang, Baosong and Zhang, Beichen and Hui, Binyuan and Zheng, Bo and Yu, Bowen and Gao, Chang and Huang, Chengen and Lv, Chenxu and others},
  journal={arXiv preprint arXiv:2505.09388},
  year={2025}
}

@article{korbak2025chain,
  title={Chain of thought monitorability: A new and fragile opportunity for ai safety},
  author={Korbak, Tomek and Balesni, Mikita and Barnes, Elizabeth and Bengio, Yoshua and Benton, Joe and Bloom, Joseph and Chen, Mark and Cooney, Alan and Dafoe, Allan and Dragan, Anca and others},
  journal={arXiv preprint arXiv:2507.11473},
  year={2025}
}

@inproceedings{10.5555/3692070.3694642,
author = {Zhou, Andy and Yan, Kai and Shlapentokh-Rothman, Michal and Wang, Haohan and Wang, Yu-Xiong},
title = {Language agent tree search unifies reasoning, acting, and planning in language models},
year = {2024},
publisher = {JMLR.org},
booktitle = {Proceedings of the 41st International Conference on Machine Learning},
articleno = {2572},
numpages = {23},
location = {Vienna, Austria},
series = {ICML'24}
}

@inproceedings{10.1145/3618260.3649777,
author = {Kalai, Adam Tauman and Vempala, Santosh S.},
title = {Calibrated Language Models Must Hallucinate},
year = {2024},
isbn = {9798400703836},
publisher = {Association for Computing Machinery},
address = {New York, NY, USA},
url = {https://doi.org/10.1145/3618260.3649777},
doi = {10.1145/3618260.3649777},
booktitle = {Proceedings of the 56th Annual ACM Symposium on Theory of Computing},
pages = {160–171},
numpages = {12},
location = {Vancouver, BC, Canada},
series = {STOC 2024}
}

@inproceedings{
yao2023react,
title={ReAct: Synergizing Reasoning and Acting in Language Models},
author={Shunyu Yao and Jeffrey Zhao and Dian Yu and Nan Du and Izhak Shafran and Karthik R Narasimhan and Yuan Cao},
booktitle={The Eleventh International Conference on Learning Representations },
year={2023},
url={https://openreview.net/forum?id=WE_vluYUL-X}
}

@inproceedings{snell2025scaling,
  title={Scaling LLM test-time compute optimally can be more effective than scaling parameters for reasoning},
  author={Snell, Charlie Victor and Lee, Jaehoon and Xu, Kelvin and Kumar, Aviral},
  booktitle={The Thirteenth International Conference on Learning Representations},
  year={2025}
}

@inproceedings{
chen2025do,
title={Do {NOT} Think That Much for 2+3=? On the Overthinking of Long Reasoning Models},
author={Xingyu Chen and Jiahao Xu and Tian Liang and Zhiwei He and Jianhui Pang and Dian Yu and Linfeng Song and Qiuzhi Liu and Mengfei Zhou and Zhuosheng Zhang and Rui Wang and Zhaopeng Tu and Haitao Mi and Dong Yu},
booktitle={Forty-second International Conference on Machine Learning},
year={2025},
url={https://openreview.net/forum?id=MSbU3L7V00}
}

@inproceedings{muennighoff-etal-2025-s1,
    title = "s1: Simple test-time scaling",
    author = "Muennighoff, Niklas  and
      Yang, Zitong  and
      Shi, Weijia  and
      Li, Xiang Lisa  and
      Fei-Fei, Li  and
      Hajishirzi, Hannaneh  and
      Zettlemoyer, Luke  and
      Liang, Percy  and
      Candes, Emmanuel  and
      Hashimoto, Tatsunori",
    editor = "Christodoulopoulos, Christos  and
      Chakraborty, Tanmoy  and
      Rose, Carolyn  and
      Peng, Violet",
    booktitle = "Proceedings of the 2025 Conference on Empirical Methods in Natural Language Processing",
    month = nov,
    year = "2025",
    address = "Suzhou, China",
    publisher = "Association for Computational Linguistics",
    url = "https://aclanthology.org/2025.emnlp-main.1025/",
    doi = "10.18653/v1/2025.emnlp-main.1025",
    pages = "20275--20321",
    ISBN = "979-8-89176-332-6"
}

@inproceedings{
yang2026towards,
title={Towards Thinking-Optimal Scaling of Test-Time Compute for {LLM} Reasoning},
author={Wenkai Yang and Shuming Ma and Yankai Lin and Furu Wei},
booktitle={The Thirty-ninth Annual Conference on Neural Information Processing Systems},
year={2026},
url={https://openreview.net/forum?id=6ICFqmixlS}
}

@misc{hsu2025groupthinkmultipleconcurrent,
      title={Group Think: Multiple Concurrent Reasoning Agents Collaborating at Token Level Granularity}, 
      author={Chan-Jan Hsu and Davide Buffelli and Jamie McGowan and Feng-Ting Liao and Yi-Chang Chen and Sattar Vakili and Da-shan Shiu},
      year={2025},
      eprint={2505.11107},
      archivePrefix={arXiv},
      primaryClass={cs.AI},
      url={https://arxiv.org/abs/2505.11107}, 
}

@inproceedings{
gao2026how,
title={How Far Are We from Optimal Reasoning Efficiency?},
author={Jiaxuan Gao and Shu Yan and Qixin Tan and lu Yang and Shusheng Xu and Wei Fu and Zhiyu Mei and Kaifeng Lyu and Yi Wu},
booktitle={The Thirty-ninth Annual Conference on Neural Information Processing Systems},
year={2026},
url={https://openreview.net/forum?id=NhAi1w3s8Z}
}

@misc{li2025tldrlongreweightingefficient,
      title={TL;DR: Too Long, Do Re-weighting for Efficient LLM Reasoning Compression}, 
      author={Zhong-Zhi Li and Xiao Liang and Zihao Tang and Lei Ji and Peijie Wang and Haotian Xu and Xing W and Haizhen Huang and Weiwei Deng and Yeyun Gong and Zhijiang Guo and Xiao Liu and Fei Yin and Cheng-Lin Liu},
      year={2025},
      eprint={2506.02678},
      archivePrefix={arXiv},
      primaryClass={cs.CL},
      url={https://arxiv.org/abs/2506.02678}, 
}

@inproceedings{
lin2026plan,
title={Plan and Budget: Effective and Efficient Test-Time Scaling on Reasoning Large Language Models},
author={Junhong Lin and Xinyue Zeng and Jie Zhu and Song Wang and Julian Shun and Jun Wu and Dawei Zhou},
booktitle={The Fourteenth International Conference on Learning Representations},
year={2026},
url={https://openreview.net/forum?id=ctspw4CqbS}
}

@article{
sui2025stop,
title={Stop Overthinking: A Survey on Efficient Reasoning for Large Language Models},
author={Yang Sui and Yu-Neng Chuang and Guanchu Wang and Jiamu Zhang and Tianyi Zhang and Jiayi Yuan and Hongyi Liu and Andrew Wen and Shaochen Zhong and Na Zou and Hanjie Chen and Xia Hu},
journal={Transactions on Machine Learning Research},
issn={2835-8856},
year={2025},
url={https://openreview.net/forum?id=HvoG8SxggZ},
note={}
}

@article{sun2025stop,
  title={Stop when enough: Adaptive early-stopping for chain-of-thought reasoning},
  author={Sun, Renliang and Cheng, Wei and Li, Dawei and Chen, Haifeng and Wang, Wei},
  journal={arXiv preprint arXiv:2510.10103},
  year={2025}
}

@article{xu2025chain,
  title={Chain of draft: Thinking faster by writing less},
  author={Xu, Silei and Xie, Wenhao and Zhao, Lingxiao and He, Pengcheng},
  journal={arXiv preprint arXiv:2502.18600},
  year={2025}
}

@inproceedings{
arora2025training,
title={Training Language Models to Reason Efficiently},
author={Daman Arora and Andrea Zanette},
booktitle={The Thirty-ninth Annual Conference on Neural Information Processing Systems},
year={2025},
url={https://openreview.net/forum?id=AiZxn84Wdo}
}

@article{team2025kimi,
  title={Kimi k1. 5: Scaling reinforcement learning with llms},
  author={Team, Kimi and Du, Angang and Gao, Bofei and Xing, Bowei and Jiang, Changjiu and Chen, Cheng and Li, Cheng and Xiao, Chenjun and Du, Chenzhuang and Liao, Chonghua and others},
  journal={arXiv preprint arXiv:2501.12599},
  year={2025}
}

@article{nayab2024concise,
  title={Concise thoughts: Impact of output length on llm reasoning and cost},
  author={Nayab, Sania and Rossolini, Giulio and Simoni, Marco and Saracino, Andrea and Buttazzo, Giorgio and Manes, Nicolamaria and Giacomelli, Fabrizio},
  journal={arXiv preprint arXiv:2407.19825},
  year={2024}
}

@inproceedings{han-etal-2025-token,
    title = "Token-Budget-Aware {LLM} Reasoning",
    author = "Han, Tingxu  and
      Wang, Zhenting  and
      Fang, Chunrong  and
      Zhao, Shiyu  and
      Ma, Shiqing  and
      Chen, Zhenyu",
    editor = "Che, Wanxiang  and
      Nabende, Joyce  and
      Shutova, Ekaterina  and
      Pilehvar, Mohammad Taher",
    booktitle = "Findings of the Association for Computational Linguistics: ACL 2025",
    month = jul,
    year = "2025",
    address = "Vienna, Austria",
    publisher = "Association for Computational Linguistics",
    url = "https://aclanthology.org/2025.findings-acl.1274/",
    doi = "10.18653/v1/2025.findings-acl.1274",
    pages = "24842--24855",
    ISBN = "979-8-89176-256-5"
}

@inproceedings{ma-etal-2025-cot,
    title = "{C}o{T}-Valve: Length-Compressible Chain-of-Thought Tuning",
    author = "Ma, Xinyin  and
      Wan, Guangnian  and
      Yu, Runpeng  and
      Fang, Gongfan  and
      Wang, Xinchao",
    editor = "Che, Wanxiang  and
      Nabende, Joyce  and
      Shutova, Ekaterina  and
      Pilehvar, Mohammad Taher",
    booktitle = "Proceedings of the 63rd Annual Meeting of the Association for Computational Linguistics (Volume 1: Long Papers)",
    month = jul,
    year = "2025",
    address = "Vienna, Austria",
    publisher = "Association for Computational Linguistics",
    url = "https://aclanthology.org/2025.acl-long.300/",
    doi = "10.18653/v1/2025.acl-long.300",
    pages = "6025--6035",
    ISBN = "979-8-89176-251-0"
}

@inproceedings{kang2025c3ot,
  title={C3ot: Generating shorter chain-of-thought without compromising effectiveness},
  author={Kang, Yu and Sun, Xianghui and Chen, Liangyu and Zou, Wei},
  booktitle={Proceedings of the AAAI Conference on Artificial Intelligence},
  volume={39},
  number={23},
  pages={24312--24320},
  year={2025}
}

@inproceedings{
luo2025opruner,
title={O1-Pruner: Length-Harmonizing Fine-Tuning for O1-Like Reasoning Pruning},
author={Haotian Luo and Li Shen and Haiying He and Yibo Wang and Shiwei Liu and Wei Li and Naiqiang Tan and Xiaochun Cao and Dacheng Tao},
booktitle={2nd AI for Math Workshop @ ICML 2025},
year={2025},
url={https://openreview.net/forum?id=ioYybCRcyW}
}

@inproceedings{zhang-etal-2025-lightthinker,
    title = "{L}ight{T}hinker: Thinking Step-by-Step Compression",
    author = "Zhang, Jintian  and
      Zhu, Yuqi  and
      Sun, Mengshu  and
      Luo, Yujie  and
      Qiao, Shuofei  and
      Du, Lun  and
      Zheng, Da  and
      Chen, Huajun  and
      Zhang, Ningyu",
    editor = "Christodoulopoulos, Christos  and
      Chakraborty, Tanmoy  and
      Rose, Carolyn  and
      Peng, Violet",
    booktitle = "Proceedings of the 2025 Conference on Empirical Methods in Natural Language Processing",
    month = nov,
    year = "2025",
    address = "Suzhou, China",
    publisher = "Association for Computational Linguistics",
    url = "https://aclanthology.org/2025.emnlp-main.673/",
    doi = "10.18653/v1/2025.emnlp-main.673",
    pages = "13307--13328",
    ISBN = "979-8-89176-332-6"
}

@inproceedings{
aggarwal2025l,
title={L1: Controlling How Long A Reasoning Model Thinks With Reinforcement Learning},
author={Pranjal Aggarwal and Sean Welleck},
booktitle={Second Conference on Language Modeling},
year={2025},
url={https://openreview.net/forum?id=4jdIxXBNve}
}

@INPROCEEDINGS{10852493,
  author={Renze, Matthew and Guven, Erhan},
  booktitle={2024 2nd International Conference on Foundation and Large Language Models (FLLM)}, 
  title={The Benefits of a Concise Chain of Thought on Problem-Solving in Large Language Models}, 
  year={2024},
  volume={},
  number={},
  pages={476-483},
  doi={10.1109/FLLM63129.2024.10852493}}

@inproceedings{
fu2025reasoning,
title={Reasoning Without Self-Doubt: More Efficient Chain-of-Thought Through Certainty Probing},
author={Yichao Fu and Junda Chen and Yonghao Zhuang and Zheyu Fu and Ion Stoica and Hao Zhang},
booktitle={ICLR 2025 Workshop on Foundation Models in the Wild},
year={2025},
url={https://openreview.net/forum?id=wpK4IMJfdX}
}

@inproceedings{
he2025semcot,
title={SemCoT: Accelerating Chain-of-Thought Reasoning through Semantically-Aligned Implicit Tokens},
author={Yinhan He and Wendy Zheng and Yaochen Zhu and Zaiyi Zheng and Lin Su and Sriram Vasudevan and Qi Guo and Liangjie Hong and Jundong Li},
booktitle={The Thirty-ninth Annual Conference on Neural Information Processing Systems},
year={2025},
url={https://openreview.net/forum?id=1ZuzFUMtx6}
}

@inproceedings{
hao2025training,
title={Training Large Language Models to Reason in a Continuous Latent Space},
author={Shibo Hao and Sainbayar Sukhbaatar and DiJia Su and Xian Li and Zhiting Hu and Jason E Weston and Yuandong Tian},
booktitle={Second Conference on Language Modeling},
year={2025},
url={https://openreview.net/forum?id=Itxz7S4Ip3}
}

@inproceedings{
rustagi2025confidencecoverage,
title={Confidence-Coverage Gating for Early Exit},
author={Aaroosh Rustagi and Hsien Xin Peng and Khushal Murthy and Attrey Koul and Ryan Lagasse and Kevin Zhu},
booktitle={NeurIPS 2025 Workshop on Efficient Reasoning},
year={2025},
url={https://openreview.net/forum?id=Ay7sRmWswq}
}

@article{wang2025entropy,
  title={Entropy After  $\{$/Think$\}$ for reasoning model early exiting},
  author={Wang, Xi and McInerney, James and Wang, Lequn and Kallus, Nathan},
  journal={arXiv preprint arXiv:2509.26522},
  year={2025}
}

@inproceedings{
tian2023just,
title={Just Ask for Calibration: Strategies for Eliciting Calibrated Confidence Scores from Language Models Fine-Tuned with Human Feedback},
author={Katherine Tian and Eric Mitchell and Allan Zhou and Archit Sharma and Rafael Rafailov and Huaxiu Yao and Chelsea Finn and Christopher D Manning},
booktitle={The 2023 Conference on Empirical Methods in Natural Language Processing},
year={2023},
url={https://openreview.net/forum?id=g3faCfrwm7}
}

@article{mao2025early,
  title={Early stopping chain-of-thoughts in large language models},
  author={Mao, Minjia and Yin, Bowen and Zhu, Yu and Fang, Xiao},
  journal={arXiv preprint arXiv:2509.14004},
  year={2025}
}

@inproceedings{
huang2025efficient,
title={Efficient Test-Time Scaling via Self-Calibration},
author={Chengsong Huang and Langlin Huang and Jixuan Leng and Jiacheng Liu and Jiaxin Huang},
booktitle={NeurIPS 2025 Workshop on Efficient Reasoning},
year={2025},
url={https://openreview.net/forum?id=RvMjxGpVOa}
}

@inproceedings{wei2022chain,
 author = {Wei, Jason and Wang, Xuezhi and Schuurmans, Dale and Bosma, Maarten and ichter, brian and Xia, Fei and Chi, Ed and Le, Quoc V and Zhou, Denny},
 booktitle = {Advances in Neural Information Processing Systems},
 editor = {S. Koyejo and S. Mohamed and A. Agarwal and D. Belgrave and K. Cho and A. Oh},
 pages = {24824--24837},
 publisher = {Curran Associates, Inc.},
 title = {Chain-of-Thought Prompting Elicits Reasoning in Large Language Models},
 url = {https://proceedings.neurips.cc/paper_files/paper/2022/file/9d5609613524ecf4f15af0f7b31abca4-Paper-Conference.pdf},
 volume = {35},
 year = {2022}
}

@inproceedings{
yang2026dynamic,
title={Dynamic Early Exit in Reasoning Models},
author={Chenxu Yang and Qingyi Si and Yongjie Duan and Zheliang Zhu and Chenyu Zhu and Qiaowei Li and Minghui Chen and Zheng Lin and Weiping Wang},
booktitle={The Fourteenth International Conference on Learning Representations},
year={2026},
url={https://openreview.net/forum?id=NpU7ZXafRi}
}

@misc{aime25,
      title={American Invitational Mathematics Examination (AIME) 2025}, 
      author={Zhang, Yifan and Math-AI, Team},
      year={2025},
}

@misc{aime24,
      title={American Invitational Mathematics Examination (AIME) 2024}, 
      author={Zhang, Yifan and Math-AI, Team},
      year={2024},
}

@inproceedings{
hendrycks2021measuring,
title={Measuring Mathematical Problem Solving With the {MATH} Dataset},
author={Dan Hendrycks and Collin Burns and Saurav Kadavath and Akul Arora and Steven Basart and Eric Tang and Dawn Song and Jacob Steinhardt},
booktitle={Thirty-fifth Conference on Neural Information Processing Systems Datasets and Benchmarks Track (Round 2)},
year={2021},
url={https://openreview.net/forum?id=7Bywt2mQsCe}
}

@inproceedings{he-etal-2024-olympiadbench,
    title = "{O}lympiad{B}ench: A Challenging Benchmark for Promoting {AGI} with Olympiad-Level Bilingual Multimodal Scientific Problems",
    author = "He, Chaoqun  and
      Luo, Renjie  and
      Bai, Yuzhuo  and
      Hu, Shengding  and
      Thai, Zhen  and
      Shen, Junhao  and
      Hu, Jinyi  and
      Han, Xu  and
      Huang, Yujie  and
      Zhang, Yuxiang  and
      Liu, Jie  and
      Qi, Lei  and
      Liu, Zhiyuan  and
      Sun, Maosong",
    editor = "Ku, Lun-Wei  and
      Martins, Andre  and
      Srikumar, Vivek",
    booktitle = "Proceedings of the 62nd Annual Meeting of the Association for Computational Linguistics (Volume 1: Long Papers)",
    month = aug,
    year = "2024",
    address = "Bangkok, Thailand",
    publisher = "Association for Computational Linguistics",
    url = "https://aclanthology.org/2024.acl-long.211/",
    doi = "10.18653/v1/2024.acl-long.211",
    pages = "3828--3850"
}

@inproceedings{
rein2024gpqa,
title={{GPQA}: A Graduate-Level Google-Proof Q\&A Benchmark},
author={David Rein and Betty Li Hou and Asa Cooper Stickland and Jackson Petty and Richard Yuanzhe Pang and Julien Dirani and Julian Michael and Samuel R. Bowman},
booktitle={First Conference on Language Modeling},
year={2024},
url={https://openreview.net/forum?id=Ti67584b98}
}

@inproceedings{liu2025answer,
    title = "Answer Convergence as a Signal for Early Stopping in Reasoning",
    author = "Liu, Xin  and
      Wang, Lu",
    editor = "Christodoulopoulos, Christos  and
      Chakraborty, Tanmoy  and
      Rose, Carolyn  and
      Peng, Violet",
    booktitle = "Proceedings of the 2025 Conference on Empirical Methods in Natural Language Processing",
    month = nov,
    year = "2025",
    address = "Suzhou, China",
    publisher = "Association for Computational Linguistics",
    url = "https://aclanthology.org/2025.emnlp-main.904/",
    doi = "10.18653/v1/2025.emnlp-main.904",
    pages = "17896--17907",
    ISBN = "979-8-89176-332-6"
}

@inproceedings{
kuhn2023semantic,
title={Semantic Uncertainty: Linguistic Invariances for Uncertainty Estimation in Natural Language Generation},
author={Lorenz Kuhn and Yarin Gal and Sebastian Farquhar},
booktitle={The Eleventh International Conference on Learning Representations },
year={2023},
url={https://openreview.net/forum?id=VD-AYtP0dve}
}

@article{farquhar2024detecting,
  title={Detecting hallucinations in large language models using semantic entropy},
  author={Farquhar, Sebastian and Kossen, Jannik and Kuhn, Lorenz and Gal, Yarin},
  journal={Nature},
  volume={630},
  number={8017},
  pages={625--630},
  year={2024},
  publisher={Nature Publishing Group UK London}
}

@article{qwen3embedding,
  title={Qwen3 Embedding: Advancing Text Embedding and Reranking Through Foundation Models},
  author={Zhang, Yanzhao and Li, Mingxin and Long, Dingkun and Zhang, Xin and Lin, Huan and Yang, Baosong and Xie, Pengjun and Yang, An and Liu, Dayiheng and Lin, Junyang and Huang, Fei and Zhou, Jingren},
  journal={arXiv preprint arXiv:2506.05176},
  year={2025}
}

@article{oord2018representation,
  title={Representation learning with contrastive predictive coding},
  author={Oord, Aaron van den and Li, Yazhe and Vinyals, Oriol},
  journal={arXiv preprint arXiv:1807.03748},
  year={2018}
}

@inproceedings{li-etal-2025-generation,
    title = "From Generation to Judgment: Opportunities and Challenges of {LLM}-as-a-judge",
    author = "Li, Dawei  and
      Jiang, Bohan  and
      Huang, Liangjie  and
      Beigi, Alimohammad  and
      Zhao, Chengshuai  and
      Tan, Zhen  and
      Bhattacharjee, Amrita  and
      Jiang, Yuxuan  and
      Chen, Canyu  and
      Wu, Tianhao  and
      Shu, Kai  and
      Cheng, Lu  and
      Liu, Huan",
    editor = "Christodoulopoulos, Christos  and
      Chakraborty, Tanmoy  and
      Rose, Carolyn  and
      Peng, Violet",
    booktitle = "Proceedings of the 2025 Conference on Empirical Methods in Natural Language Processing",
    month = nov,
    year = "2025",
    address = "Suzhou, China",
    publisher = "Association for Computational Linguistics",
    url = "https://aclanthology.org/2025.emnlp-main.138/",
    doi = "10.18653/v1/2025.emnlp-main.138",
    pages = "2757--2791",
    ISBN = "979-8-89176-332-6"
}

@InProceedings{pmlr-v139-radford21a,
  title = 	 {Learning Transferable Visual Models From Natural Language Supervision},
  author =       {Radford, Alec and Kim, Jong Wook and Hallacy, Chris and Ramesh, Aditya and Goh, Gabriel and Agarwal, Sandhini and Sastry, Girish and Askell, Amanda and Mishkin, Pamela and Clark, Jack and Krueger, Gretchen and Sutskever, Ilya},
  booktitle = 	 {Proceedings of the 38th International Conference on Machine Learning},
  pages = 	 {8748--8763},
  year = 	 {2021},
  editor = 	 {Meila, Marina and Zhang, Tong},
  volume = 	 {139},
  series = 	 {Proceedings of Machine Learning Research},
  month = 	 {18--24 Jul},
  publisher =    {PMLR},
  url = 	 {https://proceedings.mlr.press/v139/radford21a.html}
}

@misc{bercovich2025llamanemotron,
      title={Llama-Nemotron: Efficient Reasoning Models}, 
      author={Akhiad Bercovich and Itay Levy and Izik Golan and Mohammad Dabbah and Ran El-Yaniv and Omri Puny and Ido Galil and Zach Moshe and Tomer Ronen and Najeeb Nabwani and Ido Shahaf and Oren Tropp and Ehud Karpas and Ran Zilberstein and Jiaqi Zeng and Soumye Singhal and Alexander Bukharin and Yian Zhang and Tugrul Konuk and Gerald Shen and Ameya Sunil Mahabaleshwarkar and Bilal Kartal and Yoshi Suhara and Olivier Delalleau and Zijia Chen and Zhilin Wang and David Mosallanezhad and Adi Renduchintala and Haifeng Qian and Dima Rekesh and Fei Jia and Somshubra Majumdar and Vahid Noroozi and Wasi Uddin Ahmad and Sean Narenthiran and Aleksander Ficek and Mehrzad Samadi and Jocelyn Huang and Siddhartha Jain and Igor Gitman and Ivan Moshkov and Wei Du and Shubham Toshniwal and George Armstrong and Branislav Kisacanin and Matvei Novikov and Daria Gitman and Evelina Bakhturina and Jane Polak Scowcroft and John Kamalu and Dan Su and Kezhi Kong and Markus Kliegl and Rabeeh Karimi and Ying Lin and Sanjeev Satheesh and Jupinder Parmar and Pritam Gundecha and Brandon Norick and Joseph Jennings and Shrimai Prabhumoye and Syeda Nahida Akter and Mostofa Patwary and Abhinav Khattar and Deepak Narayanan and Roger Waleffe and Jimmy Zhang and Bor-Yiing Su and Guyue Huang and Terry Kong and Parth Chadha and Sahil Jain and Christine Harvey and Elad Segal and Jining Huang and Sergey Kashirsky and Robert McQueen and Izzy Putterman and George Lam and Arun Venkatesan and Sherry Wu and Vinh Nguyen and Manoj Kilaru and Andrew Wang and Anna Warno and Abhilash Somasamudramath and Sandip Bhaskar and Maka Dong and Nave Assaf and Shahar Mor and Omer Ullman Argov and Scot Junkin and Oleksandr Romanenko and Pedro Larroy and Monika Katariya and Marco Rovinelli and Viji Balas and Nicholas Edelman and Anahita Bhiwandiwalla and Muthu Subramaniam and Smita Ithape and Karthik Ramamoorthy and Yuting Wu and Suguna Varshini Velury and Omri Almog and Joyjit Daw and Denys Fridman and Erick Galinkin and Michael Evans and Katherine Luna and Leon Derczynski and Nikki Pope and Eileen Long and Seth Schneider and Guillermo Siman and Tomasz Grzegorzek and Pablo Ribalta and Monika Katariya and Joey Conway and Trisha Saar and Ann Guan and Krzysztof Pawelec and Shyamala Prayaga and Oleksii Kuchaiev and Boris Ginsburg and Oluwatobi Olabiyi and Kari Briski and Jonathan Cohen and Bryan Catanzaro and Jonah Alben and Yonatan Geifman and Eric Chung and Chris Alexiuk},
      year={2025},
      eprint={2505.00949},
      archivePrefix={arXiv},
      primaryClass={cs.CL},
      url={https://arxiv.org/abs/2505.00949}, 
}

@article{ma2025reasoning,
  title={Reasoning models can be effective without thinking},
  author={Ma, Wenjie and He, Jingxuan and Snell, Charlie and Griggs, Tyler and Min, Sewon and Zaharia, Matei},
  journal={arXiv preprint arXiv:2504.09858},
  year={2025}
}

@article{su2025between,
  title={Between underthinking and overthinking: An empirical study of reasoning length and correctness in llms},
  author={Su, Jinyan and Healey, Jennifer and Nakov, Preslav and Cardie, Claire},
  journal={arXiv preprint arXiv:2505.00127},
  year={2025}
}

@inproceedings{
ghosal2026does,
title={Does Thinking More Always Help? Mirage of Test-Time Scaling in Reasoning Models},
author={Soumya Suvra Ghosal and Souradip Chakraborty and Avinash Reddy and Yifu Lu and Mengdi Wang and Dinesh Manocha and Furong Huang and Mohammad Ghavamzadeh and Amrit Singh Bedi},
booktitle={The Thirty-ninth Annual Conference on Neural Information Processing Systems},
year={2026},
url={https://openreview.net/forum?id=tKPqbamNb9}
}

@inproceedings{
wu2025when,
title={When More is Less: Understanding Chain-of-Thought Length in {LLM}s},
author={Yuyang Wu and Yifei Wang and Tianqi Du and Stefanie Jegelka and Yisen Wang},
booktitle={Workshop on Reasoning and Planning for Large Language Models},
year={2025},
url={https://openreview.net/forum?id=W8dxn7hBkO}
}

@misc{wei2026evolutionthoughttrackingllm,
      title={The Evolution of Thought: Tracking LLM Overthinking via Reasoning Dynamics Analysis}, 
      author={Zihao Wei and Liang Pang and Jiahao Liu and Wenjie Shi and Jingcheng Deng and Shicheng Xu and Zenghao Duan and Fei Sun and Huawei Shen and Xueqi Cheng},
      year={2026},
      eprint={2508.17627},
      archivePrefix={arXiv},
      primaryClass={cs.CL},
      url={https://arxiv.org/abs/2508.17627}, 
}

@article{
hou2026thinkprune,
title={ThinkPrune: Pruning Long Chain-of-Thought of {LLM}s via Reinforcement Learning},
author={Bairu Hou and Yang Zhang and Jiabao Ji and Yujian Liu and Kaizhi Qian and Jacob Andreas and Shiyu Chang},
journal={Transactions on Machine Learning Research},
issn={2835-8856},
year={2026},
url={https://openreview.net/forum?id=V51gPu1uQD},
note={}
}

@inproceedings{
dai2026sgrpo,
title={S-{GRPO}: Early Exit via Reinforcement Learning in Reasoning Models},
author={Mz Dai and Chenxu Yang and Qingyi Si},
booktitle={The Thirty-ninth Annual Conference on Neural Information Processing Systems},
year={2026},
url={https://openreview.net/forum?id=wNMK5o0Vfg}
}

@inproceedings{
li2026making,
title={Making Slow Thinking Faster: Compressing {LLM} Chain-of-Thought via Step Entropy},
author={Zeju Li and Jianyuan Zhong and Ziyang Zheng and Xiangyu Wen and Zhijian Xu and Yingying Cheng and Fan Zhang and Qiang Xu},
booktitle={The Fourteenth International Conference on Learning Representations},
year={2026},
url={https://openreview.net/forum?id=cGLqQfS5wH}
}

@inproceedings{huang2026efficient,
  title={Efficient reasoning for large reasoning language models via certainty-guided reflection suppression},
  author={Huang, Jiameng and Lin, Baijiong and Feng, Guhao and Chen, Jierun and He, Di and Hou, Lu},
  booktitle={Proceedings of the AAAI Conference on Artificial Intelligence},
  volume={40},
  number={37},
  pages={31176--31184},
  year={2026}
}

@article{
chhikara2025mind,
title={Mind the Confidence Gap: Overconfidence, Calibration, and Distractor Effects in Large Language Models},
author={Prateek Chhikara},
journal={Transactions on Machine Learning Research},
issn={2835-8856},
year={2025},
url={https://openreview.net/forum?id=lyaHnHDdZl},
note={}
}

@article{liu2026neat,
  title={NEAT: Neuron-Based Early Exit for Large Reasoning Models},
  author={Liu, Kang and Liu, Yongkang and Yang, Xiaocui and Wang, Peidong and Zhang, Wen and Feng, Shi and Zhang, Yifei and Wang, Daling},
  journal={arXiv preprint arXiv:2602.02010},
  year={2026}
}

@inproceedings{wang-etal-2025-wait,
    title = "Wait, We Don{'}t Need to ``Wait''! Removing Thinking Tokens Improves Reasoning Efficiency",
    author = "Wang, Chenlong  and
      Feng, Yuanning  and
      Chen, Dongping  and
      Chu, Zhaoyang  and
      Krishna, Ranjay  and
      Zhou, Tianyi",
    editor = "Christodoulopoulos, Christos  and
      Chakraborty, Tanmoy  and
      Rose, Carolyn  and
      Peng, Violet",
    booktitle = "Findings of the Association for Computational Linguistics: EMNLP 2025",
    month = nov,
    year = "2025",
    address = "Suzhou, China",
    publisher = "Association for Computational Linguistics",
    url = "https://aclanthology.org/2025.findings-emnlp.394/",
    doi = "10.18653/v1/2025.findings-emnlp.394",
    pages = "7459--7482",
    ISBN = "979-8-89176-335-7"
}

@article{akgul2025lynx,
  title={LYNX: Learning Dynamic Exits for Confidence-Controlled Reasoning},
  author={Akg{\"u}l, {\"O}mer Faruk and Kalayc{\i}, Yusuf Hakan and Kannan, Rajgopal and Neiswanger, Willie and Prasanna, Viktor},
  journal={arXiv preprint arXiv:2512.05325},
  year={2025}
}

@inproceedings{
zhang2025reasoning,
title={Reasoning Models Know When They{\textquoteright}re Right: Probing Hidden States for Self-Verification},
author={Anqi Zhang and Yulin Chen and Jane Pan and Chen Zhao and Aurojit Panda and Jinyang Li and He He},
booktitle={Second Conference on Language Modeling},
year={2025},
url={https://openreview.net/forum?id=O6I0Av7683}
}

@inproceedings{qiao-etal-2025-concise,
    title = "{C}on{CISE}: Confidence-guided Compression in Step-by-step Efficient Reasoning",
    author = "Qiao, Ziqing  and
      Deng, Yongheng  and
      Zeng, Jiali  and
      Wang, Dong  and
      Wei, Lai  and
      Wang, Guanbo  and
      Meng, Fandong  and
      Zhou, Jie  and
      Ren, Ju  and
      Zhang, Yaoxue",
    editor = "Christodoulopoulos, Christos  and
      Chakraborty, Tanmoy  and
      Rose, Carolyn  and
      Peng, Violet",
    booktitle = "Proceedings of the 2025 Conference on Empirical Methods in Natural Language Processing",
    month = nov,
    year = "2025",
    address = "Suzhou, China",
    publisher = "Association for Computational Linguistics",
    url = "https://aclanthology.org/2025.emnlp-main.405/",
    doi = "10.18653/v1/2025.emnlp-main.405",
    pages = "8010--8029",
    ISBN = "979-8-89176-332-6"
}

@misc{qwq32b,
    title = {QwQ-32B: Embracing the Power of Reinforcement Learning},
    url = {https://qwenlm.github.io/blog/qwq-32b/},
    author = {Qwen Team},
    month = {March},
    year = {2025}
}

@misc{openai2025gptoss120bgptoss20bmodel,
      title={gpt-oss-120b \& gpt-oss-20b Model Card}, 
      author={OpenAI},
      year={2025},
      eprint={2508.10925},
      archivePrefix={arXiv},
      primaryClass={cs.CL},
      url={https://arxiv.org/abs/2508.10925}, 
}

@misc{5team2025glm45agenticreasoningcoding,
      title={GLM-4.5: Agentic, Reasoning, and Coding (ARC) Foundation Models}, 
      author={GLM Team and Aohan Zeng and Xin Lv and Qinkai Zheng and Zhenyu Hou and Bin Chen and Chengxing Xie and Cunxiang Wang and Da Yin and Hao Zeng and Jiajie Zhang and Kedong Wang and Lucen Zhong and Mingdao Liu and Rui Lu and Shulin Cao and Xiaohan Zhang and Xuancheng Huang and Yao Wei and Yean Cheng and Yifan An and Yilin Niu and Yuanhao Wen and Yushi Bai and Zhengxiao Du and Zihan Wang and Zilin Zhu and Bohan Zhang and Bosi Wen and Bowen Wu and Bowen Xu and Can Huang and Casey Zhao and Changpeng Cai and Chao Yu and Chen Li and Chendi Ge and Chenghua Huang and Chenhui Zhang and Chenxi Xu and Chenzheng Zhu and Chuang Li and Congfeng Yin and Daoyan Lin and Dayong Yang and Dazhi Jiang and Ding Ai and Erle Zhu and Fei Wang and Gengzheng Pan and Guo Wang and Hailong Sun and Haitao Li and Haiyang Li and Haiyi Hu and Hanyu Zhang and Hao Peng and Hao Tai and Haoke Zhang and Haoran Wang and Haoyu Yang and He Liu and He Zhao and Hongwei Liu and Hongxi Yan and Huan Liu and Huilong Chen and Ji Li and Jiajing Zhao and Jiamin Ren and Jian Jiao and Jiani Zhao and Jianyang Yan and Jiaqi Wang and Jiayi Gui and Jiayue Zhao and Jie Liu and Jijie Li and Jing Li and Jing Lu and Jingsen Wang and Jingwei Yuan and Jingxuan Li and Jingzhao Du and Jinhua Du and Jinxin Liu and Junkai Zhi and Junli Gao and Ke Wang and Lekang Yang and Liang Xu and Lin Fan and Lindong Wu and Lintao Ding and Lu Wang and Man Zhang and Minghao Li and Minghuan Xu and Mingming Zhao and Mingshu Zhai and Pengfan Du and Qian Dong and Shangde Lei and Shangqing Tu and Shangtong Yang and Shaoyou Lu and Shijie Li and Shuang Li and Shuang-Li and Shuxun Yang and Sibo Yi and Tianshu Yu and Wei Tian and Weihan Wang and Wenbo Yu and Weng Lam Tam and Wenjie Liang and Wentao Liu and Xiao Wang and Xiaohan Jia and Xiaotao Gu and Xiaoying Ling and Xin Wang and Xing Fan and Xingru Pan and Xinyuan Zhang and Xinze Zhang and Xiuqing Fu and Xunkai Zhang and Yabo Xu and Yandong Wu and Yida Lu and Yidong Wang and Yilin Zhou and Yiming Pan and Ying Zhang and Yingli Wang and Yingru Li and Yinpei Su and Yipeng Geng and Yitong Zhu and Yongkun Yang and Yuhang Li and Yuhao Wu and Yujiang Li and Yunan Liu and Yunqing Wang and Yuntao Li and Yuxuan Zhang and Zezhen Liu and Zhen Yang and Zhengda Zhou and Zhongpei Qiao and Zhuoer Feng and Zhuorui Liu and Zichen Zhang and Zihan Wang and Zijun Yao and Zikang Wang and Ziqiang Liu and Ziwei Chai and Zixuan Li and Zuodong Zhao and Wenguang Chen and Jidong Zhai and Bin Xu and Minlie Huang and Hongning Wang and Juanzi Li and Yuxiao Dong and Jie Tang},
      year={2025},
      eprint={2508.06471},
      archivePrefix={arXiv},
      primaryClass={cs.CL},
      url={https://arxiv.org/abs/2508.06471}, 
}

@misc{kimiteam2025kimik2openagentic,
      title={Kimi K2: Open Agentic Intelligence}, 
      author={Kimi Team and Yifan Bai and Yiping Bao and Guanduo Chen and Jiahao Chen and Ningxin Chen and Ruijue Chen and Yanru Chen and Yuankun Chen and Yutian Chen and Zhuofu Chen and Jialei Cui and Hao Ding and Mengnan Dong and Angang Du and Chenzhuang Du and Dikang Du and Yulun Du and Yu Fan and Yichen Feng and Kelin Fu and Bofei Gao and Hongcheng Gao and Peizhong Gao and Tong Gao and Xinran Gu and Longyu Guan and Haiqing Guo and Jianhang Guo and Hao Hu and Xiaoru Hao and Tianhong He and Weiran He and Wenyang He and Chao Hong and Yangyang Hu and Zhenxing Hu and Weixiao Huang and Zhiqi Huang and Zihao Huang and Tao Jiang and Zhejun Jiang and Xinyi Jin and Yongsheng Kang and Guokun Lai and Cheng Li and Fang Li and Haoyang Li and Ming Li and Wentao Li and Yanhao Li and Yiwei Li and Zhaowei Li and Zheming Li and Hongzhan Lin and Xiaohan Lin and Zongyu Lin and Chengyin Liu and Chenyu Liu and Hongzhang Liu and Jingyuan Liu and Junqi Liu and Liang Liu and Shaowei Liu and T. Y. Liu and Tianwei Liu and Weizhou Liu and Yangyang Liu and Yibo Liu and Yiping Liu and Yue Liu and Zhengying Liu and Enzhe Lu and Lijun Lu and Shengling Ma and Xinyu Ma and Yingwei Ma and Shaoguang Mao and Jie Mei and Xin Men and Yibo Miao and Siyuan Pan and Yebo Peng and Ruoyu Qin and Bowen Qu and Zeyu Shang and Lidong Shi and Shengyuan Shi and Feifan Song and Jianlin Su and Zhengyuan Su and Xinjie Sun and Flood Sung and Heyi Tang and Jiawen Tao and Qifeng Teng and Chensi Wang and Dinglu Wang and Feng Wang and Haiming Wang and Jianzhou Wang and Jiaxing Wang and Jinhong Wang and Shengjie Wang and Shuyi Wang and Yao Wang and Yejie Wang and Yiqin Wang and Yuxin Wang and Yuzhi Wang and Zhaoji Wang and Zhengtao Wang and Zhexu Wang and Chu Wei and Qianqian Wei and Wenhao Wu and Xingzhe Wu and Yuxin Wu and Chenjun Xiao and Xiaotong Xie and Weimin Xiong and Boyu Xu and Jing Xu and Jinjing Xu and L. H. Xu and Lin Xu and Suting Xu and Weixin Xu and Xinran Xu and Yangchuan Xu and Ziyao Xu and Junjie Yan and Yuzi Yan and Xiaofei Yang and Ying Yang and Zhen Yang and Zhilin Yang and Zonghan Yang and Haotian Yao and Xingcheng Yao and Wenjie Ye and Zhuorui Ye and Bohong Yin and Longhui Yu and Enming Yuan and Hongbang Yuan and Mengjie Yuan and Haobing Zhan and Dehao Zhang and Hao Zhang and Wanlu Zhang and Xiaobin Zhang and Yangkun Zhang and Yizhi Zhang and Yongting Zhang and Yu Zhang and Yutao Zhang and Yutong Zhang and Zheng Zhang and Haotian Zhao and Yikai Zhao and Huabin Zheng and Shaojie Zheng and Jianren Zhou and Xinyu Zhou and Zaida Zhou and Zhen Zhu and Weiyu Zhuang and Xinxing Zu},
      year={2025},
      eprint={2507.20534},
      archivePrefix={arXiv},
      primaryClass={cs.LG},
      url={https://arxiv.org/abs/2507.20534}, 
}

@misc{cobbe2021trainingverifierssolvemath,
      title={Training Verifiers to Solve Math Word Problems}, 
      author={Karl Cobbe and Vineet Kosaraju and Mohammad Bavarian and Mark Chen and Heewoo Jun and Lukasz Kaiser and Matthias Plappert and Jerry Tworek and Jacob Hilton and Reiichiro Nakano and Christopher Hesse and John Schulman},
      year={2021},
      eprint={2110.14168},
      archivePrefix={arXiv},
      primaryClass={cs.LG},
      url={https://arxiv.org/abs/2110.14168}, 
}

@inproceedings{
guha2026openthoughts,
title={OpenThoughts: Data Recipes for Reasoning Models},
author={Etash Kumar Guha and Ryan Marten and Sedrick Keh and Negin Raoof and Georgios Smyrnis and Hritik Bansal and Marianna Nezhurina and Jean Mercat and Trung Vu and Zayne Rea Sprague and Ashima Suvarna and Benjamin Feuer and Leon Liangyu Chen and Zaid Khan and Eric Frankel and Sachin Grover and Caroline Choi and Niklas Muennighoff and Shiye Su and Wanjia Zhao and John Yang and Shreyas Pimpalgaonkar and Kartik sharma and Charlie Cheng-Jie Ji and Yichuan Deng and Sarah M Pratt and Vivek Ramanujan and Jon Saad-Falcon and Stutee Acharya and Jeffrey Li and Achal Dave and Alon Albalak and Kushal Arora and Blake Wulfe and Chinmay Hegde and Greg Durrett and Sewoong Oh and Mohit Bansal and Saadia Gabriel and Aditya Grover and Kai-Wei Chang and Vaishaal Shankar and Aaron Gokaslan and Mike A Merrill and Tatsunori Hashimoto and Yejin Choi and Jenia Jitsev and Reinhard Heckel and Maheswaran Sathiamoorthy and Alex Dimakis and Ludwig Schmidt},
booktitle={The Fourteenth International Conference on Learning Representations},
year={2026},
url={https://openreview.net/forum?id=7xjoTuaNmN}
}

@misc{zhao2024swiftascalablelightweightinfrastructure,
      title={SWIFT:A Scalable lightWeight Infrastructure for Fine-Tuning},
      author={Yuze Zhao and Jintao Huang and Jinghan Hu and Xingjun Wang and Yunlin Mao and Daoze Zhang and Zeyinzi Jiang and Zhikai Wu and Baole Ai and Ang Wang and Wenmeng Zhou and Yingda Chen},
      year={2024},
      eprint={2408.05517},
      archivePrefix={arXiv},
      primaryClass={cs.CL},
      url={https://arxiv.org/abs/2408.05517},
}

@inproceedings{
jain2025livecodebench,
title={LiveCodeBench: Holistic and Contamination Free Evaluation of Large Language Models for Code},
author={Naman Jain and King Han and Alex Gu and Wen-Ding Li and Fanjia Yan and Tianjun Zhang and Sida Wang and Armando Solar-Lezama and Koushik Sen and Ion Stoica},
booktitle={The Thirteenth International Conference on Learning Representations},
year={2025},
url={https://openreview.net/forum?id=chfJJYC3iL}
}

@inproceedings{
lu2024mathvista,
title={MathVista: Evaluating Mathematical Reasoning of Foundation Models in Visual Contexts},
author={Pan Lu and Hritik Bansal and Tony Xia and Jiacheng Liu and Chunyuan Li and Hannaneh Hajishirzi and Hao Cheng and Kai-Wei Chang and Michel Galley and Jianfeng Gao},
booktitle={The Twelfth International Conference on Learning Representations},
year={2024},
url={https://openreview.net/forum?id=KUNzEQMWU7}
}

@inproceedings{
wang2024measuring,
title={Measuring Multimodal Mathematical Reasoning with MATH-Vision Dataset},
author={Ke Wang and Junting Pan and Weikang Shi and Zimu Lu and Houxing Ren and Aojun Zhou and Mingjie Zhan and Hongsheng Li},
booktitle={The Thirty-eight Conference on Neural Information Processing Systems Datasets and Benchmarks Track},
year={2024},
url={https://openreview.net/forum?id=QWTCcxMpPA}
}

@inproceedings{
hu2022lora,
title={Lo{RA}: Low-Rank Adaptation of Large Language Models},
author={Edward J Hu and yelong shen and Phillip Wallis and Zeyuan Allen-Zhu and Yuanzhi Li and Shean Wang and Lu Wang and Weizhu Chen},
booktitle={International Conference on Learning Representations},
year={2022},
url={https://openreview.net/forum?id=nZeVKeeFYf9}
}

@inproceedings{NEURIPS2023_a85b405e,
 author = {Rafailov, Rafael and Sharma, Archit and Mitchell, Eric and Manning, Christopher D and Ermon, Stefano and Finn, Chelsea},
 booktitle = {Advances in Neural Information Processing Systems},
 editor = {A. Oh and T. Naumann and A. Globerson and K. Saenko and M. Hardt and S. Levine},
 pages = {53728--53741},
 publisher = {Curran Associates, Inc.},
 title = {Direct Preference Optimization: Your Language Model is Secretly a Reward Model},
 url = {https://proceedings.neurips.cc/paper_files/paper/2023/file/a85b405ed65c6477a4fe8302b5e06ce7-Paper-Conference.pdf},
 volume = {36},
 year = {2023}
}

@article{shao2024deepseekmath,
  title={Deepseekmath: Pushing the limits of mathematical reasoning in open language models},
  author={Shao, Zhihong and Wang, Peiyi and Zhu, Qihao and Xu, Runxin and Song, Junxiao and Bi, Xiao and Zhang, Haowei and Zhang, Mingchuan and Li, YK and others},
  journal={arXiv preprint arXiv:2402.03300},
  year={2024}
}

@book{maccartney2009natural,
  title={Natural language inference},
  author={MacCartney, Bill},
  year={2009},
  publisher={Stanford University}
}

@inproceedings{
ye2025limo,
title={{LIMO}: Less is More for Reasoning},
author={Yixin Ye and Zhen Huang and Yang Xiao and Ethan Chern and Shijie Xia and Pengfei Liu},
booktitle={Second Conference on Language Modeling},
year={2025},
url={https://openreview.net/forum?id=T2TZ0RY4Zk}
}

@inproceedings{
yu2026dapo,
title={{DAPO}: An Open-Source {LLM} Reinforcement Learning System at Scale},
author={Qiying Yu and Zheng Zhang and Ruofei Zhu and Yufeng Yuan and Xiaochen Zuo and YuYue and Weinan Dai and Tiantian Fan and Gaohong Liu and Juncai Liu and LingJun Liu and Xin Liu and Haibin Lin and Zhiqi Lin and Bole Ma and Guangming Sheng and Yuxuan Tong and Chi Zhang and Mofan Zhang and Ru Zhang and Wang Zhang and Hang Zhu and Jinhua Zhu and Jiaze Chen and Jiangjie Chen and Chengyi Wang and Hongli Yu and Yuxuan Song and Xiangpeng Wei and Hao Zhou and Jingjing Liu and Wei-Ying Ma and Ya-Qin Zhang and Lin Yan and Yonghui Wu and Mingxuan Wang},
booktitle={The Thirty-ninth Annual Conference on Neural Information Processing Systems},
year={2026},
url={https://openreview.net/forum?id=2a36EMSSTp}
}

@inproceedings{
brantley2026accelerating,
title={Accelerating {RL} for {LLM} Reasoning with Optimal Advantage Regression},
author={Kiant{\'e} Brantley and Mingyu Chen and Zhaolin Gao and Jason D. Lee and Wen Sun and Wenhao Zhan and Xuezhou Zhang},
booktitle={The Thirty-ninth Annual Conference on Neural Information Processing Systems},
year={2026},
url={https://openreview.net/forum?id=T1V8BJO0iG}
}

@inproceedings{
yu2025premise,
title={{PREMISE}: Scalable and Strategic Prompt Optimization for Efficient Mathematical Reasoning in Large Reasoning Models},
author={Ye Yu and Yaoning Yu and Haibo Jin and Haohan Wang},
booktitle={NeurIPS 2025 Workshop on Efficient Reasoning},
year={2025},
url={https://openreview.net/forum?id=8mI3i9LXj3}
}

@InProceedings{Li_2024_CVPR,
    author    = {Li, Li and Peng, Jiawei and Chen, Huiyi and Gao, Chongyang and Yang, Xu},
    title     = {How to Configure Good In-Context Sequence for Visual Question Answering},
    booktitle = {Proceedings of the IEEE/CVF Conference on Computer Vision and Pattern Recognition (CVPR)},
    month     = {June},
    year      = {2024},
    pages     = {26710-26720}
}

@article{chen2025mvi,
  title={Mvi-bench: A comprehensive benchmark for evaluating robustness to misleading visual inputs in lvlms},
  author={Chen, Huiyi and Peng, Jiawei and Min, Dehai and Sun, Changchang and Chen, Kaijie and Yan, Yan and Yang, Xu and Cheng, Lu},
  journal={arXiv preprint arXiv:2511.14159},
  year={2025}
}

@inproceedings{chen2025enhancing,
  title={Enhancing Multimodal In-Context Learning for Image Classification through Coreset Optimization},
  author={Chen, Huiyi and Peng, Jiawei and Tang, Kaihua and Geng, Xin and Yang, Xu},
  booktitle={Proceedings of the 33rd ACM International Conference on Multimedia},
  pages={5130--5139},
  year={2025}
}

@article{li2026phgpo,
  title={PhGPO: Pheromone-Guided Policy Optimization for Long-Horizon Tool Planning},
  author={Li, Yu and Cai, Guangfeng and Yang, Shengtian and Luo, Han and Han, Shuo and He, Xu and Li, Dong and Feng, Lei},
  journal={arXiv preprint arXiv:2602.13691},
  year={2026}
}

@inproceedings{li2024mateval,
  title={MATEval: a multi-agent discussion framework for advancing open-ended text evaluation},
  author={Li, Yu and Zhang, Shenyu and Wu, Rui and Huang, Xiutian and Chen, Yongrui and Xu, Wenhao and Qi, Guilin and Min, Dehai},
  booktitle={International Conference on Database Systems for Advanced Applications},
  pages={415--426},
  year={2024},
  organization={Springer}
}

@inproceedings{zhang2024dee,
  title={DEE: Dual-Stage Explainable Evaluation Method for Text Generation},
  author={Zhang, Shenyu and Li, Yu and Wu, Rui and Huang, Xiutian and Chen, Yongrui and Xu, Wenhao and Qi, Guilin},
  booktitle={International Conference on Database Systems for Advanced Applications},
  pages={390--401},
  year={2024},
  organization={Springer}
}

@article{zou2025llm,
  title={Llm-based human-agent collaboration and interaction systems: A survey},
  author={Zou, Henry Peng and Huang, Wei-Chieh and Wu, Yaozu and Chen, Yankai and Miao, Chunyu and Nguyen, Hoang and Zhou, Yue and Zhang, Weizhi and Fang, Liancheng and He, Langzhou and others},
  journal={arXiv preprint arXiv:2505.00753},
  year={2025}
}

\newpage

\appendix

\section{Overthinking Analysis and Failure Modes of Answer-Level Signals}
\label{app:overthinking}

\subsection{Overthinking Prevalence}

To quantify the extent of overthinking, we perform a counterfactual analysis
on full CoT reasoning traces across five representative LRMs and five
benchmarks. For each question, we identify the \emph{golden step}: the
earliest reasoning step at which the model's intermediate trial answer matches
its final answer. The tokens generated after the golden step are classified as post-answer
redundancy.

\begin{figure}[h]
    \centering
    \begin{subfigure}[t]{0.48\textwidth}
        \centering
        \includegraphics[width=\linewidth]{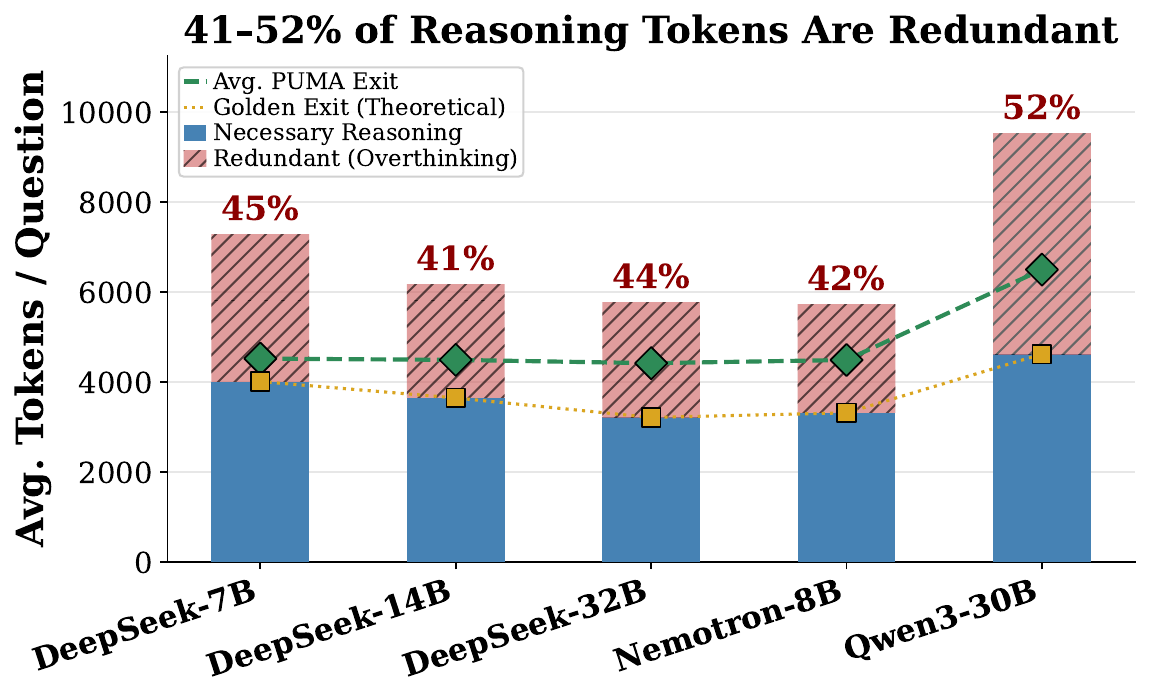}
        \caption{Token breakdown before and after the model first reaches its final answer.}
        \label{fig:overthinking_breakdown}
    \end{subfigure}
    \hfill
    \begin{subfigure}[t]{0.48\textwidth}
        \centering
        \includegraphics[width=\linewidth]{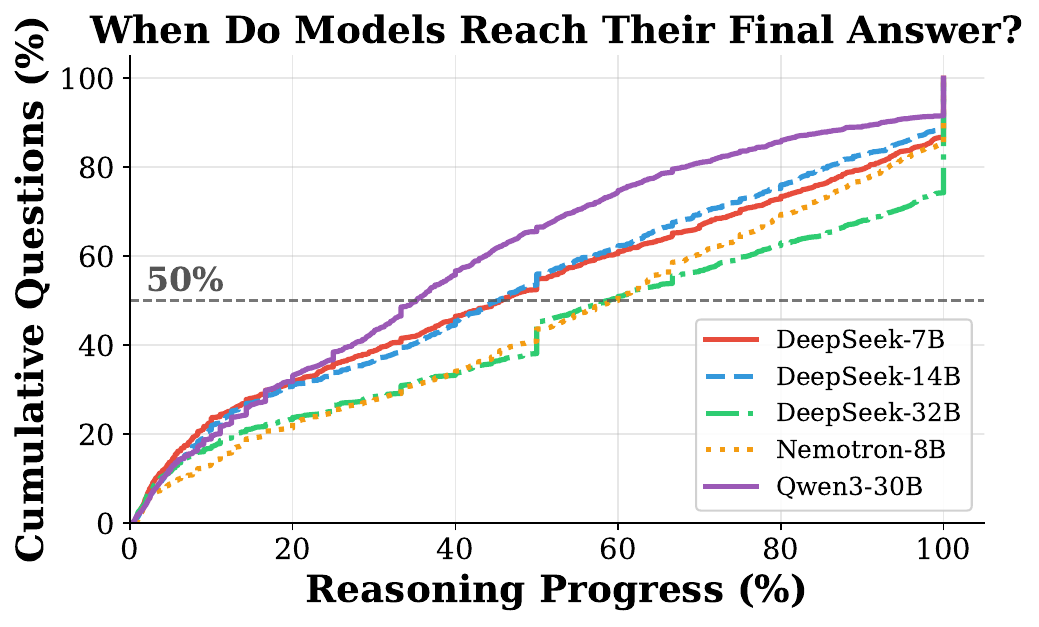}
        \caption{Cumulative fraction of questions whose final answer has already been reached, as a function of reasoning progress.}
        \label{fig:golden_step_cdf}
    \end{subfigure}
    \caption{
    \textbf{Overthinking analysis across five representative LRMs.}
    Left: a substantial fraction of reasoning tokens is generated after the model has already reached its final answer, accounting for 41--52\% of tokens across models.
    Right: many examples reach the final answer well before the end of the reasoning chain, further illustrating the prevalence of post-answer overthinking.
    }
    \label{fig:overthinking_analysis}
\end{figure}

Figure~\ref{fig:overthinking_analysis} summarizes the results. Across all
five models, 41--52\% of reasoning tokens are generated after the model has
already committed to its final answer. The pattern is consistent across
model families and scales: even the strongest model (Qwen3-30B-A3B-Thinking)
spends over half its tokens on post-answer re-verification, rephrasing, and
re-derivation. The cumulative distribution in Figure~\ref{fig:golden_step_cdf} further
shows that models typically commit to their final answer well before
the reasoning chain completes, with most models reaching the 50\%
mark between 40--60\% of reasoning progress.

\subsection{Failure Rates of Answer-Level Stopping Signals}
\label{app:failure_modes}

We analyze the failure rates of the two dominant classes of answer-level
stopping signals by applying their criteria retroactively to full CoT
traces. At each reasoning step, we induce a trial answer and compute its
confidence score as the geometric mean of token-level log-probabilities,
following DEER~\cite{yang2026dynamic}, and check trial-answer consistency
across consecutive steps~\cite{liu2025answer,fu2025reasoning}.

\begin{table*}[h]
\centering
\caption{Failure rates (\%) of answer-level stopping signals across benchmarks and model families. A failure is defined as a first-triggered exit where the trial answer is incorrect.}
\label{tab:signal_failure_rates}
\vspace{0.1cm}
\begin{adjustbox}{width=0.65\textwidth}
\setlength{\tabcolsep}{4pt}
\footnotesize
\begin{tabular}{l ccccc c}
\toprule
& \textbf{AIME24} & \textbf{AIME25} & \textbf{MATH-500}
& \textbf{OlymBench} & \textbf{GPQA-D} & \textbf{Avg.} \\
\midrule
\multicolumn{7}{l}{\textbf{DeepSeek-R1-Distill-Qwen-7B}} \\
\quad Confidence  & 50.0 & 63.3 & 31.9 & 59.3 & 38.4 & 48.6 \\
\quad Consistency & 76.9 & 70.0 & 36.5 & 67.9 & 67.2 & 63.7 \\
\midrule
\multicolumn{7}{l}{\textbf{DeepSeek-R1-Distill-Qwen-14B}} \\
\quad Confidence  & 53.3 & 62.1 & 28.8 & 52.3 & 35.3 & 46.4 \\
\quad Consistency & 80.0 & 79.3 & 37.1 & 70.5 & 61.7 & 65.7 \\
\midrule
\multicolumn{7}{l}{\textbf{DeepSeek-R1-Distill-Qwen-32B}} \\
\quad Confidence  & 34.5 & 56.7 & 19.9 & 46.8 & 26.8 & 36.9 \\
\quad Consistency & 69.0 & 72.4 & 45.5 & 70.1 & 59.6 & 63.3 \\
\midrule
\multicolumn{7}{l}{\textbf{Llama-3.1-Nemotron-Nano-8B}} \\
\quad Confidence  & 44.8 & 69.0 & 32.7 & 57.3 & 54.2 & 51.6 \\
\quad Consistency & 83.3 & 83.3 & 34.9 & 67.9 & 77.3 & 69.3 \\
\midrule
\multicolumn{7}{l}{\textbf{Qwen3-30B-A3B-Thinking}} \\
\quad Confidence  & 29.6 & 51.9 & 26.3 & 48.4 & 36.5 & 38.5 \\
\quad Consistency & 73.3 & 67.9 & 31.1 & 64.0 & 55.3 & 58.3 \\
\midrule
\rowcolor{methodgray!30}
\multicolumn{7}{l}{\textit{All models avg.}} \\
\rowcolor{methodgray!30}
\quad Confidence  & 42.4 & 60.6 & 27.9 & 52.8 & 38.2 & \textbf{44.4} \\
\rowcolor{methodgray!30}
\quad Consistency & 76.5 & 74.6 & 37.0 & 68.1 & 64.2 & \textbf{64.1} \\
\bottomrule
\end{tabular}
\end{adjustbox}
\vspace{-0.3cm}
\end{table*}

We evaluate across five models and five benchmarks. For the confidence
signal, we flag a step as a candidate exit whenever the confidence score
exceeds $\lambda = 0.95$, following DEER~\cite{yang2026dynamic}. For the
consistency signal, we flag a step whenever the model produces $k{=}3$
consecutive identical trial answers, following
Dynasor~\cite{fu2025reasoning}. Table~\ref{tab:signal_failure_rates} reports the
failure rate of each signal, defined as the fraction of first-triggered
exits where the trial answer is incorrect. Averaged over all 25
(model, benchmark) combinations, the failure rate is \textbf{44\%} for
confidence-based exits and \textbf{64\%} for consistency-based exits.
Failure rates are highest on the most challenging benchmarks: on AIME24
and AIME25, consistency-based exits are incorrect in up to 83\% of
triggered steps, while even confidence-based exits fail up to 69\% of the time.

\begin{table}[h]
\centering
\caption{Counterfactual analysis of answer-level stopping signals,
aggregated across all five benchmarks. \textbf{Premature\%} denotes the
fraction of incorrect exits where the corresponding uninterrupted Full-CoT
trajectory would eventually reach the correct answer, meaning that early
stopping would prevent later self-correction and induce accuracy loss.}
\label{tab:counterfactual}
\vspace{0.2cm}
\resizebox{0.6\textwidth}{!}{
\renewcommand{\arraystretch}{1.1}
\setlength{\tabcolsep}{8pt}
\begin{tabular}{l cc}
\toprule
& \multicolumn{2}{c}{\textbf{Premature\%}} \\
\cmidrule(lr){2-3}
\textbf{Model} & \textbf{Confidence} & \textbf{Consistency} \\
\midrule
DeepSeek-R1-Distill-Qwen-7B  & 41.1 & 47.8 \\
DeepSeek-R1-Distill-Qwen-14B & 39.8 & 54.6 \\
DeepSeek-R1-Distill-Qwen-32B & 28.7 & 52.0 \\
Llama-3.1-Nemotron-Nano-8B   & 45.2 & 53.5 \\
Qwen3-30B-A3B-Thinking       & 57.4 & 66.0 \\
\midrule
\rowcolor{methodgray!30}
\textit{Average}              & \textbf{42.4} & \textbf{54.8} \\
\bottomrule
\end{tabular}
}
\vspace{-0.3cm}
\end{table}

\subsection{Counterfactual Analysis: Do Signal Misfires Prevent Self-Correction?}
\label{app:counterfactual}

The failure rates above show how often a signal would stop at an incorrect trial answer, but they do not tell whether early stopping itself changes the final outcome: in some cases, the model may already be on a trajectory that remains wrong even if allowed to continue. To distinguish genuinely premature exits from non-recoverable failures, we compare each failed first-triggered exit against the corresponding full, uninterrupted reasoning chain.

We categorize a failed exit as a \textbf{premature exit} if the trial answer at the triggered step is incorrect but the full reasoning chain eventually reaches the correct final answer. In contrast, we categorize it as a \textbf{non-recoverable failure} if the full reasoning chain also ends with an incorrect answer. Premature exits are especially harmful because the stopping signal would convert an otherwise correct full-CoT trajectory into an incorrect early-exit output.

Table~\ref{tab:counterfactual} reports the breakdown aggregated across all five benchmarks. A substantial fraction of signal failures are premature rather than non-recoverable: across all models, \textbf{42.4}\% of confidence-based failures and \textbf{54.8}\% of consistency-based failures occur on traces where the model would have self-corrected if allowed to continue. This shows that answer-level signals do not merely fire on trajectories that remain wrong; they often fire while the model is still in the process of correcting or revising its answer.

The counterfactual results also explain why trial-answer consistency can be particularly misleading. A model may produce the same wrong trial answer for several consecutive steps while it is still exploring the problem, creating the appearance of answer stability before the reasoning trajectory has actually converged. Stopping at this point prevents later correction. These findings reinforce the need for a reasoning-level signal: safe early exit should not only ask whether the current answer appears stable, but also whether the reasoning process has stopped making semantically novel progress.

\subsection{Threshold Sensitivity of Answer-Level Stopping Signals}
\label{app:answer_signal_threshold_sensitivity}

One possible concern is that the high failure rates above may simply reflect
suboptimal threshold choices. To test this, we sweep each answer-level signal
over a range of operating points and examine the resulting tradeoff between
token reduction and failure rate. For confidence-based stopping, we vary the
confidence threshold $\lambda \in \{0.93,0.94,0.95,0.96,0.97\}$. For
consistency-based stopping, we vary the required number of consecutive identical
trial answers $k \in \{1,2,3,4,5,6,7,8\}$. We run this diagnostic analysis on
DeepSeek-R1-Distill-Qwen-7B over two challenging benchmarks, OlympiadBench and
GPQA Diamond.

\begin{figure}[t]
    \centering
    \includegraphics[width=0.93\textwidth]{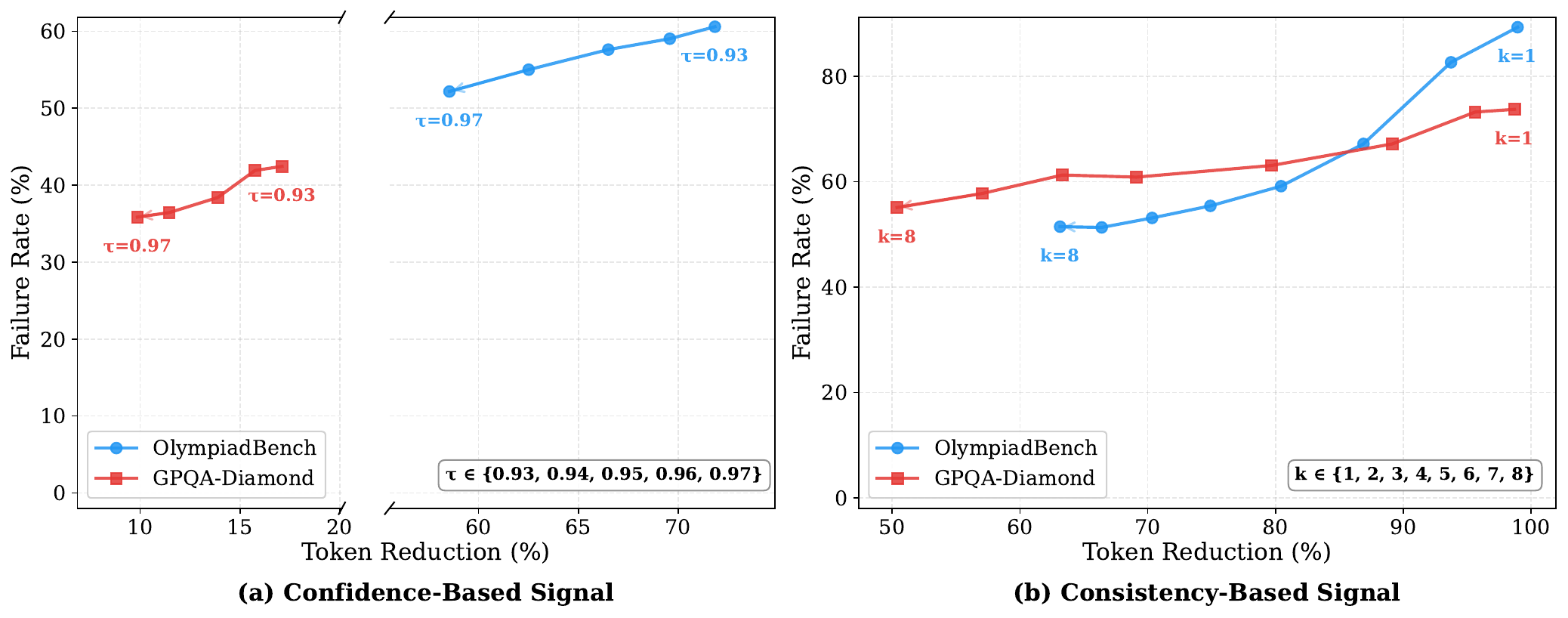}
    \caption{
    Threshold sensitivity of standalone answer-level stopping signals.
    We sweep the confidence threshold $\lambda$ and the consistency window size
    $k$ on OlympiadBench and GPQA Diamond using DeepSeek-R1-Distill-Qwen-7B.
    Each point corresponds to one operating point. Across thresholds, standalone
    answer-level signals exhibit a persistent tradeoff between failure rate and
    token reduction, without a clear setting that is both safe and efficient.
    }
    \label{fig:answer_signal_threshold_sensitivity}
\end{figure}

Figure~\ref{fig:answer_signal_threshold_sensitivity} shows that threshold tuning
does not remove the fundamental tradeoff for standalone answer-level stopping
signals. For confidence-based stopping on OlympiadBench, even conservative
thresholds still yield high failure rates while stopping a large fraction of
reasoning tokens. On GPQA Diamond, lower failure rates are possible only when
token reduction becomes small, making early exit much less useful.
Consistency-based stopping shows a similar pattern: increasing $k$ reduces some
premature triggers, but does not produce a clear operating point that is both
safe and efficient. These results suggest that the limitation is not merely a
poor threshold choice; standalone answer-level stopping signals are insufficient
to determine whether the reasoning trajectory has actually converged.

\section{Implementation Details}
\label{appendix:implementation_details}

\subsection{Reasoning Step Segmentation}
\label{appendix:step_segmentation}

PUMA operates at the level of reasoning steps rather than individual tokens. We therefore
segment each reasoning chain into coherent steps using a deterministic, lightweight
procedure. Reasoning step segmentation is a common preprocessing step in recent work on
efficient reasoning, where approaches often rely on heuristic segmentations such as blank-line
splitting~\cite{li2026making}, keyword-based transition
detection~\cite{zhang2025reasoning}, or explicit block-marker
matching~\cite{qiao-etal-2025-concise}. Our procedure follows the blank-line approach
and adds a lightweight merging stage to control step granularity.
The segmentation is applied identically during Redundancy Detector training and
at inference time, ensuring that the detector sees the same type of step units in both
settings.

Given the raw reasoning text, we first split the chain at blank-line boundaries, which
preserves the model's natural paragraph structure. We then assign each paragraph a coarse
semantic role using simple string-level cues, such as problem setup, calculation,
self-correction, verification, conclusion, or general reasoning. These labels are not used
as model inputs; they only guide how adjacent paragraphs are grouped into larger reasoning
steps.

The final merging stage balances semantic coherence and step length. Enumerated paragraphs
or major semantic transitions are treated as natural step boundaries, while short adjacent
paragraphs with compatible roles are merged when doing so preserves coherence. We use a
target step length range of $[L_{\min}, L_{\max}] = [200,1000]$ characters: very short
segments are often too noisy for redundancy detection, while overly long segments may hide
local repetition. Tiny trailing segments are merged into the preceding step.

The procedure is intentionally string-based and does not invoke any learned segmenter,
adding negligible overhead to generation. This design keeps PUMA compatible with standard
serving pipelines while providing stable step units for both detector training and online
early-exit decisions.

\subsection{Redundancy Detector Training}
\label{appendix:rd_training}

We construct a large-scale contrastive dataset to train the Redundancy Detector.
The goal is to teach the detector whether a new reasoning step contributes
semantically novel progress relative to prior reasoning, or instead restates,
re-derives, or loops over existing content.

\textbf{Source reasoning traces.}
We collect long-CoT reasoning traces from models that are not used as main
evaluation models, including QwQ-32B~\cite{qwq32b}, GPT-OSS-120B~\cite{openai2025gptoss120bgptoss20bmodel}, GLM-4.7-Thinking~\cite{5team2025glm45agenticreasoningcoding}, and
Kimi-K2-Thinking~\cite{kimiteam2025kimik2openagentic}. These models are prompted on AMC23 and GSM8K~\cite{cobbe2021trainingverifierssolvemath}, and we further
include reasoning chains sampled from Open-Thoughts-114K~\cite{guha2026openthoughts}. In total, the source collection covers approximately 5,098 distinct questions. After applying the
reasoning-step segmentation procedure in Appendix~\ref{appendix:step_segmentation},
we obtain roughly 1.97M candidate step pairs.

\textbf{Stage 1: seed annotation.}\quad
We first annotate a seed set of reasoning-step pairs using GPT-5-mini. The
annotation prompt asks the model to provide a short rationale and then assign a
binary label, where \(y=0\) denotes a novel step and \(y=1\) denotes a redundant step.
This stage yields 40,844 annotated examples across the four source models.
Each example can contain multiple redundant positives and multiple novel
negatives; on average, a seed example contains 3.8 positives and 26.6 negatives.
After expanding multi-positive examples into single-positive contrastive rows,
this stage contributes approximately 155K anchor--positive training pairs.

\textbf{Stage 2: redundant-step synthesis.}\quad
To increase coverage of redundant reasoning patterns, we synthesize redundant
counterparts for seed-labeled novel steps. Given a question, a previous reasoning
step, and a novel current step, GPT-4o-mini rewrites the current step into a
redundant version that preserves the topic but removes substantive new progress.
The rewrite prompt covers common redundancy modes such as restatement,
hesitation, paraphrase, and transitional padding. This stage adds approximately
560K synthetic redundant pairs. The total annotation cost across both stages is around \$2,000.

\textbf{Final contrastive dataset.}\quad
We convert all data into the InfoNCE format used for training, where each row
contains an anchor step, one redundant positive, and a set of novel negatives.
After expanding multi-positive annotations, adding synthetic redundant pairs,
deduplicating examples, and removing rows without valid negatives, the final
dataset contains 701,641 contrastive training rows.

\paragraph{Training setup.}
We implement detector training using the MS-Swift framework~\cite{zhao2024swiftascalablelightweightinfrastructure}. Instead of full-parameter fine-tuning, we adapt Qwen3-Embedding-0.6B with LoRA to reduce overfitting risk. We use LoRA rank \(r=32\), \(\alpha=64\), dropout \(0.1\), and apply LoRA to the \texttt{q\_proj}, \texttt{k\_proj}, \texttt{v\_proj}, \texttt{o\_proj}, \texttt{gate\_proj}, \texttt{up\_proj}, and \texttt{down\_proj} modules. The detector is trained with an InfoNCE contrastive objective, batch size 16 with gradient accumulation 4 (effective batch size 64), learning rate \(1\times 10^{-4}\), and 5 epochs.

\textbf{Detector quality and threshold selection.}
We evaluate the trained detector on a held-out set of anchor-positive-negative
triples, where the positive is a redundant step and the negative is a novel step
relative to the anchor. The detector achieves 91.26\% pairwise ranking accuracy,
where a prediction is counted as correct if
\begin{equation}
\mathrm{sim}(\mathrm{anchor}, \mathrm{redundant})
>
\mathrm{sim}(\mathrm{anchor}, \mathrm{novel}).
\end{equation}
The average cosine similarity is 0.333 for redundant pairs and 0.184 for novel
pairs, yielding an average margin of 0.148.
For the default PUMA setting, we set \(\tau_{\mathrm{sim}}=0.35\), selected on a
held-out calibration set as a conservative operating point for candidate-exit
detection. At this threshold, the detector achieves 91.54\% absolute
classification accuracy and 93.58\% true-negative rate. The high true-negative
rate makes the detector conservative: it avoids over-triggering on genuinely
novel reasoning steps, while missed redundant steps mainly reduce potential token
savings rather than directly harming correctness. Moreover, all detector-flagged
candidate exits are still filtered by Answer Verification before PUMA stops.

\paragraph{LLM Annotation Prompts for Redundancy Detector Training}
\label{appendix:rd_prompts}

We use two LLM-based prompts to construct supervision for the Redundancy Detector:
a seed novelty annotation prompt (Table~\ref{tab:rd_seed_annotation_prompt}) and a
redundant-step synthesis prompt (Table~\ref{tab:rd_redundant_synthesis_prompt}).
For clarity, we present both prompts under the label convention used in our final
training data, where \(y=0\) denotes a novel reasoning step and \(y=1\) denotes a
redundant reasoning step.

\begin{table}[h!]
\small
\centering
\begin{tcolorbox}[width=0.95\textwidth]
\textcolor{blue}{\textbf{Role:}}
\\
You are an AI assistant for assessing the quality of logical reasoning.
The user will provide a question, a previous reasoning step, and a current reasoning step.
Your job is to judge whether the current step provides additional information compared
to the previous step. Do not solve the problem yourself; only evaluate the novelty of
the current step.
\\\\
\textcolor{blue}{\textbf{Label Convention:}}
\\
Output \texttt{0} if the current step introduces new logical or semantic information.
\\
Output \texttt{1} if the current step is redundant, i.e., it mainly restates,
rephrases, verifies, or loops over information already present in the previous step.
\\\\
\textcolor{blue}{\textbf{Output Format:}}
\\
First provide a short explanation enclosed in
\texttt{<explanation>...</explanation>}.
Then output the final label enclosed in \texttt{<result>...</result>}.
The result must be exactly one character, either \texttt{0} or \texttt{1}.
Output nothing after the closing \texttt{</result>} tag.
\\\\
\textcolor{blue}{\textbf{Few-shot Examples:}}\par
\textit{[Four worked demonstrations illustrating novel vs. redundant reasoning-step judgments.]}
\\\\
\textcolor{blue}{\textbf{Input:}}
\\
\textbf{Problem:} \textit{<input question>}
\\
\textbf{Previous Step:} \textit{<previous reasoning step>}
\\
\textbf{Current Step:} \textit{<current reasoning step>}
\\\\
\textcolor{blue}{\textbf{Expected Output:}}
\\
\texttt{<explanation>}
\\
\textit{<brief rationale explaining whether the current step is novel or redundant>}
\\
\texttt{</explanation>}
\\
\texttt{<result>0 or 1</result>}
\end{tcolorbox}
\caption{Prompt template for seed novelty annotation. The displayed label convention matches the remapped labels used in the final Redundancy Detector training data.}
\label{tab:rd_seed_annotation_prompt}
\end{table}

\begin{table}[h!]
\small
\centering
\begin{tcolorbox}[width=0.95\textwidth]
\textcolor{blue}{\textbf{Role:}}
\\
You are an expert at analyzing mathematical reasoning steps. Your task is to
generate a redundant version of a given reasoning step.
\\\\
\textcolor{blue}{\textbf{Definition:}}
\\
A redundant step is one that does not advance the solution compared to the previous
step. It may restate or summarize what was already said, express hesitation or
self-checking without new insight, rephrase the same mathematical relationship in
different words, or add transitional phrases without actual progress.
\\\\
\textcolor{blue}{\textbf{Instruction:}}
\\
Given a mathematical problem, a previous reasoning step, and a current step that
provides new information, generate a redundant version of the current step.
The redundant version should:
\\
-- not provide new information beyond what is already in the previous step;
\\
-- sound natural as a reasoning step;
\\
-- be similar in length to the current step.
\\\\
\textcolor{blue}{\textbf{Few-shot Examples:}}
\\
\textit{[Examples demonstrate common redundancy modes, including restatement,
hesitation, paraphrase, and transitional padding.]}
\\\\
\textcolor{blue}{\textbf{Input:}}
\\
\textbf{Problem:} \textit{<input question>}
\\
\textbf{Previous Step:} \textit{<previous reasoning step>}
\\
\textbf{Current Step (Novel):} \textit{<current reasoning step that introduces new information>}
\\\\
\textcolor{blue}{\textbf{Output:}}
\\
\textbf{Redundant Version:} \textit{<rewritten reasoning step that sounds natural but adds no substantive new progress>}
\end{tcolorbox}
\caption{Prompt template for redundant-step synthesis. GPT-4o-mini rewrites seed-labeled novel steps into redundant counterparts to augment positive examples for contrastive training.}
\label{tab:rd_redundant_synthesis_prompt}
\end{table}

\subsection{Choice of Redundancy Signal}
\label{appendix:rd_signal}

Semantic Entropy commonly operationalizes semantic equivalence using Natural Language
Inference (NLI), by asking whether two textual outputs mutually entail each other.
Motivated by this connection, we compare PUMA's embedding-based Redundancy Detector
with two NLI-style alternatives. The first variant, \textbf{ICL-NLI}, uses Qwen3-0.6B
with 4-shot in-context prompting: given two reasoning steps and four labelled examples,
the model emits a binary ``Yes/No'' judgment for whether the current step is redundant
with respect to the previous one. The second variant, \textbf{FT-NLI}, uses the same
Qwen3-0.6B backbone fine-tuned on 10K NLI-style redundancy examples. For both variants,
the binary judgment replaces PUMA's default redundancy score; all other PUMA components
are kept unchanged.

Table~\ref{tab:rd_signal} reports results on DS-7B and Nemotron-8B across five benchmarks.
The embedding-based detector achieves the best accuracy--efficiency trade-off: it is the
only signal that preserves (and even slightly improves) accuracy over Full CoT (+1.3 on
average) while still achieving meaningful token reduction (27.9\%). ICL-NLI is too
conservative---it rarely fires, leaving most reasoning uncompressed (8.2\% TR). FT-NLI
shows the opposite failure mode: it achieves 39.9\% TR but drops accuracy by 3.8 points,
indicating that NLI-style judgments do not align well with reasoning-level convergence
even after task-specific fine-tuning.

Beyond accuracy and token reduction, the embedding detector is substantially cheaper at
inference time, averaging about 21\,ms per question compared with $\sim$93\,ms for both NLI variants. These results support using the trained embedding model as PUMA's default redundancy signal.

We additionally compare against the base (unfine-tuned) Qwen3-Embedding-0.6B to isolate the contribution of contrastive training. Base-Emb achieves higher token reduction but at a substantial accuracy cost, and varying its similarity threshold does not recover the gap, suggesting that task-specific fine-tuning is essential for reliable redundancy detection.

\begin{table}[t]
\centering
\caption{Redundancy signal and detector comparison. Fine-tuned Embedding is PUMA's default detector: a LoRA-adapted Qwen3-Embedding-0.6B trained with an InfoNCE contrastive objective. Base Embedding uses the off-the-shelf Qwen3-Embedding-0.6B without contrastive fine-tuning, with two representative similarity thresholds. ICL-NLI and FT-NLI replace the detector with NLI-style redundancy judgments while keeping all other PUMA components unchanged. Results are averaged over five benchmarks.}
\label{tab:rd_signal}
\vspace{0.2cm}
\resizebox{0.82\textwidth}{!}{
\renewcommand{\arraystretch}{1.1}
\setlength{\tabcolsep}{7pt}
\begin{tabular}{l cc cc cc}
\toprule
& \multicolumn{2}{c}{\textbf{DS-7B}}
& \multicolumn{2}{c}{\textbf{Nemotron-8B}}
& \multicolumn{2}{c}{\textbf{Average}} \\
\cmidrule(lr){2-3} \cmidrule(lr){4-5} \cmidrule(lr){6-7}
\textbf{Signal}
& \textbf{$\Delta$Acc}$\uparrow$ & \textbf{TR}$\uparrow$
& \textbf{$\Delta$Acc}$\uparrow$ & \textbf{TR}$\uparrow$
& \textbf{$\Delta$Acc}$\uparrow$ & \textbf{TR}$\uparrow$ \\
\midrule
\rowcolor{pumagreen!25}
\textbf{Fine-tuned Emb. (PUMA)}
 & \textbf{+2.2} & \textbf{35.6}
 & \textbf{+0.4} & \textbf{20.1}
 & \textbf{+1.3} & \textbf{27.9} \\
\midrule
Base Emb. ($\tau=0.42$)
 & \textcolor{darkred}{-2.9} & 46.1
 & \textcolor{darkred}{-3.3} & 37.3
 & \textcolor{darkred}{-3.1} & 41.7 \\
Base Emb. ($\tau=0.50$)
 & \textcolor{darkred}{-4.1} & 45.5
 & \textcolor{darkred}{-3.3} & 37.5
 & \textcolor{darkred}{-3.7} & 41.5 \\
\midrule
ICL-NLI
 & \textcolor{darkred}{-2.1} & 10.9
 & \textcolor{darkred}{-0.1} & 5.5
 & \textcolor{darkred}{-1.1} & 8.2 \\
FT-NLI
 & \textcolor{darkred}{-4.1} & 44.3
 & \textcolor{darkred}{-3.5} & 35.4
 & \textcolor{darkred}{-3.8} & 39.9 \\
\bottomrule
\end{tabular}
}
\vspace{-0.3cm}
\end{table}

\subsection{Redundancy Detector Lookback Window}
\label{appendix:rd_window}

The Redundancy Detector compares the current reasoning step against the previous
\(k\) reasoning steps and flags a candidate exit when the maximum similarity exceeds
\(\tau_{\mathrm{sim}}\). PUMA uses \(k=1\) by default, comparing each step only
with its immediate predecessor. Table~\ref{tab:window_size} evaluates
\(k \in \{1,2,4,8,\mathrm{all}\}\) on DS-7B and Nemotron-8B, where
\(k=\mathrm{all}\) compares the current step against all preceding steps.

Across both models, \(k=1\) provides the best accuracy-preserving behavior. Increasing
\(k\) yields more aggressive exits and higher token reduction, but it also worsens
\(\Delta\)Acc on average, from \(+1.3\) at \(k=1\) to \(-2.4\) at
\(k=\mathrm{all}\). This pattern suggests that larger windows introduce additional
false-positive triggers: a current step may resemble a much earlier step while still
contributing useful progress relative to the immediately preceding context. As a result, looking too far back can incorrectly trigger exits during productive reasoning. We therefore use
\(k=1\) as the default because it provides the most favorable accuracy--compression
tradeoff in this sweep.

\begin{table}[h]
\centering
\caption{Redundancy Detector lookback window: $k$ is the number of
preceding reasoning steps the detector compares the current step against
(PUMA default: $k{=}1$); $k{=}\text{all}$ compares against every prior
step. Each row reports averages over five benchmarks.}
\label{tab:window_size}
\vspace{0.2cm}
\resizebox{0.68\textwidth}{!}{
\renewcommand{\arraystretch}{1.1}
\setlength{\tabcolsep}{7pt}
\begin{tabular}{c cc cc cc}
\toprule
& \multicolumn{2}{c}{\textbf{DS-7B}}
& \multicolumn{2}{c}{\textbf{Nemotron-8B}}
& \multicolumn{2}{c}{\textbf{Average}} \\
\cmidrule(lr){2-3} \cmidrule(lr){4-5} \cmidrule(lr){6-7}
\textbf{$k$}
& \textbf{$\Delta$Acc}$\uparrow$ & \textbf{TR$\uparrow$}$\uparrow$
& \textbf{$\Delta$Acc}$\uparrow$ & \textbf{TR$\uparrow$}$\uparrow$
& \textbf{$\Delta$Acc}$\uparrow$ & \textbf{TR$\uparrow$}$\uparrow$ \\
\midrule
\rowcolor{pumagreen!25}
\textbf{1} (default)
 & \textbf{+2.2} & 35.6
 & \textbf{+0.4} & 20.1
 & \textbf{+1.3} & 27.9 \\
\midrule
2    & +0.0                       & 37.7
     & \textcolor{darkred}{-0.8}  & 25.5
     & \textcolor{darkred}{-0.4}  & 31.6 \\
4    & \textcolor{darkred}{-1.7}  & 40.0
     & \textcolor{darkred}{-3.3}  & 28.3
     & \textcolor{darkred}{-2.5}  & 34.2 \\
8    & \textcolor{darkred}{-0.5}  & 41.6
     & \textcolor{darkred}{-1.5}  & 31.2
     & \textcolor{darkred}{-1.0}  & 36.4 \\
all  & \textcolor{darkred}{-2.0}  & 46.4
     & \textcolor{darkred}{-3.8}  & 37.5
     & \textcolor{darkred}{-2.9}  & 41.9 \\
\bottomrule
\end{tabular}
}
\vspace{-0.3cm}
\end{table}

\subsection{A Semantic-Entropy Perspective on Reasoning Convergence}
\label{appendix:se_perspective}
We provide a semantic-entropy perspective to explain why reasoning-step redundancy
is a useful signal for convergence. Semantic Entropy~\cite{kuhn2023semantic,farquhar2024detecting}
measures uncertainty in LLM outputs through semantic diversity rather than surface-form
variation. We adapt this intuition from multiple sampled outputs to successive steps
within a single reasoning trajectory: if recent steps continue to introduce semantically
distinct content, the model is still exploring; if they collapse into repeated verification,
restatement, or re-derivation, the trajectory has likely entered a locally converged phase.
To formalize this view, consider a recent reasoning window $W_t^{(w)} = (r_{t-w+1}, \ldots, r_t)$ and let \(c(r)\) denote the semantic cluster associated with reasoning step \(r\). We define
the \emph{Reasoning Semantic Entropy} of this window as
\begin{equation}
\mathrm{RSE}_w(t) = -\sum_{c} p_t(c)\log p_t(c),
\end{equation} 
where \(p_t(c)\) is the fraction of steps in \(W_t^{(w)}\) assigned to cluster \(c\).
High \(\mathrm{RSE}_w(t)\) indicates semantically diverse recent reasoning and continued
exploration, whereas low \(\mathrm{RSE}_w(t)\) indicates that recent steps have collapsed
into a small set of redundant semantic patterns. Thus, low local RSE corresponds to local
reasoning convergence.

PUMA does not explicitly compute \(\mathrm{RSE}_w(t)\) online. Doing so would require
clustering recent reasoning steps, for example using entailment-based semantic equivalence
as in Semantic Entropy, which would add substantial overhead for deployment-time early exit. Instead,
PUMA uses the Redundancy Detector as a lightweight online proxy for local semantic collapse:
high similarity between recent steps suggests that the reasoning trajectory is becoming less
semantically diverse and less likely to be making novel progress. 
This provides an interpretive lens for PUMA: the Redundancy Detector approximates the local semantic collapse that low RSE would capture, without explicitly computing semantic entropy.

\subsection{Trial Answer Induction}
\label{appendix:trial_answer}

At each candidate stopping point, PUMA appends a \emph{confident ending}
suffix to the current reasoning prefix and prompts the model to commit to
a trial answer. The suffix is task-specific:

\begin{itemize}[itemsep=1pt, topsep=2pt, parsep=0pt, leftmargin=1.5em]
\item \textbf{Math} (AIME, MATH-500, OlympiadBench):
\texttt{\textbackslash n**Final Answer**\textbackslash n\textbackslash nThe final answer is \textbackslash boxed\{}
\item \textbf{Multiple choice} (GPQA-Diamond):
\texttt{\textbackslash n**Final Answer**\textbackslash n\textbackslash nThe answer choice is \textbackslash boxed\{}
\item \textbf{Code} (LiveCodeBench):
\texttt{</think>\textbackslash n\textbackslash n\#\#\# Solution Code\textbackslash n```python\textbackslash n}
\end{itemize}

For math and multiple-choice tasks, the suffix intentionally does not close
the \texttt{</think>} tag, keeping the model in thinking mode so that it
commits to an answer without switching to an explanation mode that might
introduce new reasoning. For code, we close the tag explicitly because code
answers require the model to generate a complete Python program in solution
mode.

\textbf{Confidence computation.}\quad
For math tasks, we extract the token span inside \texttt{\textbackslash boxed\{...\}}
using brace matching and compute confidence as the geometric mean of token-level
log-probabilities over the answer tokens (excluding braces), following
DEER~\cite{yang2026dynamic}. For GPQA, the single answer token (A/B/C/D)
competes against the full vocabulary, yielding artificially low absolute
log-probabilities. We therefore apply a temperature-scaled softmax restricted
to the four answer choices to obtain a calibrated confidence score. For code,
we use the log-probabilities of all generated tokens, since code answers lack
a \texttt{\textbackslash boxed\{\}} structure.

\textbf{Answer consistency.}\quad
Trial answers across the verification window are compared via exact string
match for math and multiple-choice tasks. For code, we use fuzzy matching via Python's \texttt{difflib.SequenceMatcher}
(sequence similarity $\geq 0.8$) to accommodate minor whitespace and
formatting differences.

Generation is capped at ${\sim}30$ tokens for math, which is typically
sufficient for a complete boxed answer. For code, we cap at ${\sim}50$
tokens, which rarely covers a full program but suffices to compute a
meaningful confidence estimate from the opening tokens of the solution.
Generating complete code trial answers could require hundreds of tokens
per probe, making the overhead prohibitive for an online early-exit system.

\subsection{Extended Hyperparameter Analysis}
\label{appendix:hyperparameters}

\begin{figure}[t]
\centering
\includegraphics[width=0.98\textwidth]{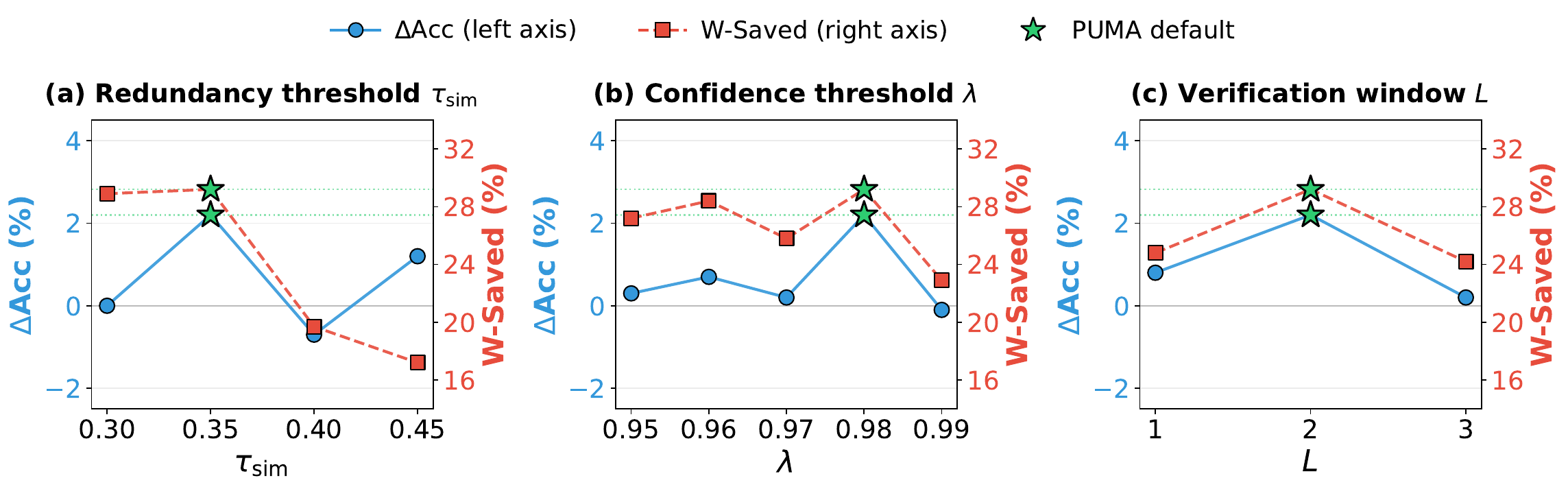}
\caption{Sensitivity to PUMA's three core stopping hyperparameters on
DS-7B, averaged over five benchmarks. Star markers and dashed
horizontal lines indicate the default configuration.}
\label{fig:sensitivity}
\end{figure}

Figure~\ref{fig:sensitivity} sweeps PUMA's three core stopping
hyperparameters on DS-7B across five benchmarks. A notable finding is
that PUMA is robust across a wide range of settings: out of 9
alternative configurations tested, only two produce a negative
$\Delta$Acc ($-0.7$ at $\tau_{\text{sim}}{=}0.40$ and $-0.1$ at
$\lambda{=}0.99$), and both are within 1pp of Full CoT. This
robustness reflects PUMA's layered safety design: even when one
hyperparameter is set suboptimally, the remaining components
(Redundancy Detector, Answer Verification, Loop Breaker) continue to
prevent most unsafe exits.

The redundancy threshold $\tau_{\text{sim}}$ is the most
safety-critical parameter. Lowering it to 0.30 maintains token savings
but eliminates the accuracy margin ($\Delta$Acc drops from +2.2 to
0.0), while raising it to 0.40 over-suppresses the redundancy signal,
nearly halving token savings without recovering accuracy.
The confidence threshold $\lambda$ is stable across [0.95, 0.98], with
only minor accuracy--efficiency variation. The sharp drop at 0.99
($-$6.3pp TR) indicates that requiring near-perfect trial-answer
confidence causes the verifier to veto most candidate exits.
The verification window \(L\) controls how much evidence PUMA requires before
stopping. \(L=1\) exits after the first verified candidate and is therefore more
susceptible to single-step false positives. \(L=3\) is overly conservative: the
additional verified candidate often does not appear before generation continues,
reducing token reduction without improving accuracy. The default \(L=2\) provides
the best balance.

The main stopping hyperparameters ($\tau_{\mathrm{sim}}$, $\lambda$, $L$) are shared across all models. The Loop Breaker consecutive-redundancy threshold $m$ is configured per model based on validation performance on a held-out AMC23 set, as different models exhibit different redundancy patterns in late-stage reasoning. DS-R1-Distill-Qwen-7B and Nemotron-Nano-8B use $m{=}1$; DS-R1-Distill-Qwen-14B uses $m{=}3$; DS-R1-Distill-Qwen-32B uses $m{=}4$. For Qwen3-30B-A3B-Thinking, the Loop Breaker is not activated, as verified exits alone provide sufficient token savings.

\section{Experimental Details and Full Results}
\label{appendix:full_results}

\subsection{Benchmark and Dataset Statistics}
\label{appendix:dataset_stats}

Table~\ref{tab:dataset_stats} summarizes the benchmarks used in our experiments. The main evaluation covers five challenging reasoning benchmarks spanning competition mathematics, olympiad-level STEM, and graduate-level science. For OlympiadBench, we use the standard text-only English open-ended math subset (675 problems). Generalization experiments additionally evaluate on code generation (LiveCodeBench) and vision-language reasoning (MathVista and MathVision). For MathVista, we evaluate on the first 200 problems from the testmini split (1,000 problems total). For MathVision, we evaluate on the first 200 problems from the full test set (3,040 problems total).

\begin{table}[t]
\centering
\caption{Dataset statistics for all evaluation benchmarks.}
\label{tab:dataset_stats}
\vspace{0.1cm}
\small
\renewcommand{\arraystretch}{1.05}
\setlength{\tabcolsep}{6pt}
\begin{tabular}{l c l}
\toprule
\textbf{Benchmark} & \textbf{\# Examples} & \textbf{Domain} \\
\midrule
\multicolumn{3}{l}{\textit{Main reasoning benchmarks}} \\
\midrule
MATH-500~\cite{hendrycks2021measuring} & 500 & Competition mathematics \\
AIME24~\cite{aime24} & 30 & Contest mathematics \\
AIME25~\cite{aime25} & 30 & Contest mathematics \\
OlympiadBench~\cite{he-etal-2024-olympiadbench} & 675 & Olympiad-level math and physics \\
GPQA-Diamond~\cite{rein2024gpqa} & 198 & Graduate-level science QA \\
\midrule
\multicolumn{3}{l}{\textit{Generalization benchmarks}} \\
\midrule
LiveCodeBench~\cite{jain2025livecodebench} & 880 & Code generation \\
MathVista~\cite{lu2024mathvista} & 200 & Vision-language mathematical reasoning \\
MathVision~\cite{wang2024measuring} & 200 & Vision-language mathematical reasoning \\
\bottomrule
\end{tabular}
\vspace{-0.2cm}
\end{table}

\subsection{Existing Assets and Licenses}
\label{appendix:assets}

Table~\ref{tab:assets} summarizes the main existing assets used in this work and their licenses. All assets are publicly available and cited with their original references. We use these assets for research evaluation and method development, and do not redistribute third-party model weights or benchmark datasets as part of this work.

\begin{table}[h]
\centering
\caption{Licenses for main existing assets used in this work.}
\label{tab:assets}
\small
\renewcommand{\arraystretch}{1.05}
\setlength{\tabcolsep}{5pt}
\begin{tabular}{l l l}
\toprule
\textbf{Asset} & \textbf{Type} & \textbf{License} \\
\midrule
DeepSeek-R1-Distill & Model & MIT \\
Llama-3.1-Nemotron-Nano-8B & Model & NVIDIA Open Model License \\
Qwen3-30B-A3B-Thinking & Model & Apache 2.0 \\
Qwen3-Embedding-0.6B & Model & Apache 2.0 \\
\midrule
MATH-500 & Dataset & MIT \\
AIME24/25 & Dataset & Apache 2.0 \\
OlympiadBench & Dataset & Apache 2.0 \\
GPQA-Diamond & Dataset & CC-BY 4.0 \\
LiveCodeBench & Dataset & MIT \\
MathVista & Dataset & CC-BY-SA 4.0 \\
MathVision & Dataset & MIT \\
\midrule
vLLM & Software & Apache 2.0 \\
MS-Swift & Software & Apache 2.0 \\
\bottomrule
\end{tabular}
\end{table}

\subsection{Full Main Results Across Five Models}
\label{appendix:full_main_results}

\begin{figure}[h]
\centering
\includegraphics[width=0.75\textwidth]{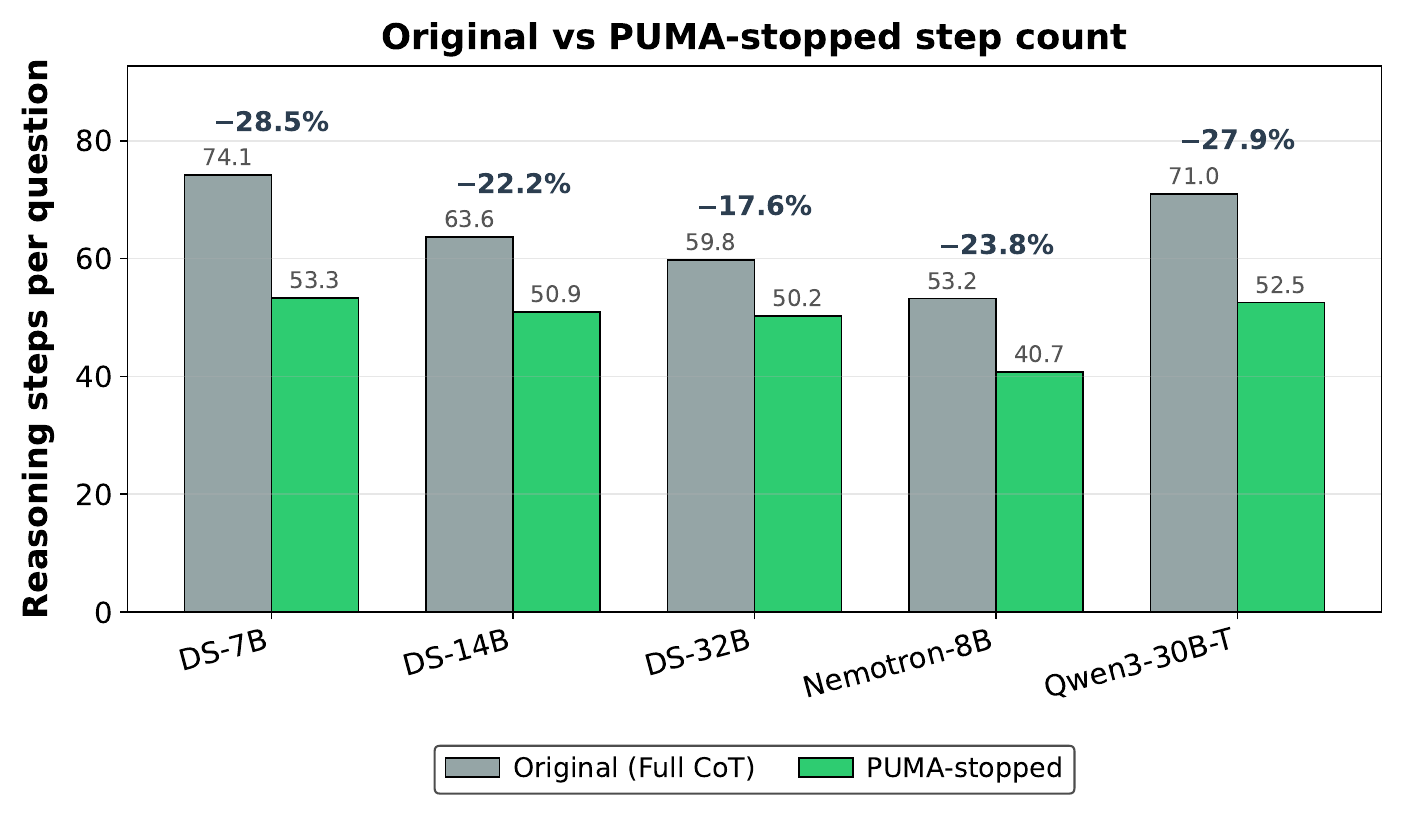}
\caption{Reasoning-step count per question: original Full-CoT (gray)
vs PUMA-stopped (green), averaged over five benchmarks. Annotated
percentages report $(\text{orig} - \text{stopped}) / \text{orig}$.}
\label{fig:step_savings}
\end{figure}

Table~\ref{tab:main_results} in the main text reports results for three
models. Table~\ref{tab:full_results} extends this to all five models,
including DS-R1-Distill-Qwen-14B, and reports per-benchmark token
counts alongside accuracy and token reduction.
Figure~\ref{fig:step_savings} further reports the average number of
reasoning steps before and after PUMA's early-exit intervention. PUMA
saves 17.6--28.5\% of reasoning steps across models, with deeper
compression on longer chains (DS-7B: 28.5\%, Qwen3-30B-T: 27.9\%)
and more conservative savings on shorter chains (DS-32B: 17.6\%).
No-Think is a prompt-only baseline that asks the model to skip explicit reasoning, but it does not impose a decoding-time stopping rule or a hard length budget. Its effectiveness therefore depends strongly on model-specific instruction following and reasoning-format behavior. On Llama-3.1-Nemotron-Nano-8B, the No-Think prompt often fails to suppress long-form generation, leading to outputs that are longer than Full CoT and substantially less accurate. This explains the negative token reduction observed for No-Think on Nemotron and highlights a limitation of prompt-only compression: asking a reasoning model to skip thinking is not a reliable substitute for monitoring and stopping the reasoning process online.

\begin{table*}[h]
\centering
\caption{Full results across all five reasoning models and five benchmarks.
Acc (\%, $\uparrow$): accuracy; Tok ($\downarrow$): average output tokens per question; TR (\%, $\uparrow$): token reduction percentage relative to Full CoT.}
\label{tab:full_results}
\vspace{0.2cm}
\begin{adjustbox}{width=\textwidth}
\tiny
\renewcommand{\arraystretch}{0.9}
\setlength{\tabcolsep}{2pt}
\begin{tabular}{l ccc ccc ccc ccc ccc cc}
\toprule
& \multicolumn{3}{c}{\textbf{MATH-500}}
& \multicolumn{3}{c}{\textbf{AIME24}}
& \multicolumn{3}{c}{\textbf{AIME25}}
& \multicolumn{3}{c}{\textbf{GPQA-D}}
& \multicolumn{3}{c}{\textbf{OlymBench}}
& \multicolumn{2}{c}{\textbf{Overall}} \\
\cmidrule(lr){2-4} \cmidrule(lr){5-7} \cmidrule(lr){8-10} \cmidrule(lr){11-13} \cmidrule(lr){14-16} \cmidrule(lr){17-18}
Method
& Acc$\uparrow$ & Tok$\downarrow$ & TR$\uparrow$
& Acc$\uparrow$ & Tok$\downarrow$ & TR$\uparrow$
& Acc$\uparrow$ & Tok$\downarrow$ & TR$\uparrow$
& Acc$\uparrow$ & Tok$\downarrow$ & TR$\uparrow$
& Acc$\uparrow$ & Tok$\downarrow$ & TR$\uparrow$
& Acc$\uparrow$ & TR$\uparrow$ \\
\midrule
\multicolumn{18}{l}{\textbf{DeepSeek-R1-Distill-Qwen-7B}} \\
\rowcolor{methodgray!30}
Full CoT & 90.0 & 4145 & 0.0 & 50.0 & 14280 & 0.0 & 43.3 & 15040 & 0.0 & 49.0 & 7153 & 0.0 & 57.6 & 8988 & 0.0 & 58.0 & 0.0 \\
No-Think & 79.0 & 830 & 80.0 & 20.0 & 4615 & 67.7 & 23.3 & 6326 & 57.9 & 28.3 & 904 & 87.3 & 42.8 & 1882 & 79.1 & 38.7 & 79.9 \\
CCoT & 85.0 & 1622 & 60.9 & 36.7 & 5775 & 59.6 & 33.3 & 5989 & 60.2 & 48.0 & 3178 & 55.6 & 49.2 & 3804 & 57.7 & 50.4 & 58.6 \\
CoD & 80.4 & 1662 & 59.9 & 40.0 & 6151 & 56.9 & 33.3 & 6214 & 58.7 & 50.5 & 3518 & 50.8 & 43.0 & 3973 & 55.8 & 49.4 & 56.6 \\
Plan\&Budget & 84.6 & 2133 & 48.5 & 43.3 & 6722 & 52.9 & 23.3 & 6938 & 53.9 & 38.4 & 4565 & 36.2 & 49.3 & 4470 & 50.3 & 47.8 & 47.9 \\
Ans.\ Conv. & 61.8 & 974 & 76.5 & 26.7 & 2390 & 83.3 & 20.0 & 2583 & 82.8 & 34.3 & 544 & 92.4 & 32.6 & 1550 & 82.8 & 35.1 & 81.9 \\
Dynasor & 84.2 & 2878 & 30.6 & 23.3 & 11175 & 21.7 & 33.3 & 13677 & 9.1 & 39.4 & 2388 & 66.6 & 49.0 & 5923 & 34.1 & 45.9 & 36.6 \\
DEER & 90.6 & 2503 & 39.6 & 40.0 & 12261 & 14.1 & 50.0 & 13058 & 13.2 & 51.5 & 8398 & \textcolor{darkred}{-17.4} & 57.5 & 7171 & 20.2 & 57.9 & 21.5 \\
\rowcolor{pumagreen!25}
PUMA (ours) & 89.6 & 2546 & 38.6 & 60.0 & 9957 & 30.3 & 46.7 & 10558 & 29.8 & 49.0 & 6866 & 4.0 & 55.6 & 5107 & 43.2 & 60.2 & 35.6 \\
\midrule
\multicolumn{18}{l}{\textbf{DeepSeek-R1-Distill-Qwen-14B}} \\
\rowcolor{methodgray!30}
Full CoT & 91.0 & 3526 & 0.0 & 63.3 & 10209 & 0.0 & 50.0 & 12663 & 0.0 & 52.5 & 6132 & 0.0 & 61.5 & 7678 & 0.0 & 63.7 & 0.0 \\
No-Think & 80.8 & 1010 & 71.3 & 46.7 & 8689 & 14.9 & 40.0 & 9964 & 21.3 & 35.4 & 966 & 84.2 & 45.5 & 3348 & 56.4 & 49.7 & 63.8 \\
CCoT & 89.6 & 1771 & 49.8 & 50.0 & 5873 & 42.5 & 33.3 & 6376 & 49.6 & 57.6 & 3352 & 45.3 & 52.3 & 3970 & 48.3 & 56.6 & 48.3 \\
CoD & 88.8 & 2211 & 37.3 & 53.3 & 6189 & 39.4 & 30.0 & 6581 & 48.0 & 52.5 & 4009 & 34.6 & 54.8 & 4441 & 42.2 & 55.9 & 39.5 \\
Plan\&Budget & 90.2 & 2703 & 23.3 & 43.3 & 6755 & 33.8 & 36.7 & 7059 & 44.3 & 55.0 & 4803 & 21.7 & 53.8 & 4898 & 36.2 & 55.8 & 29.8 \\
Ans.\ Conv. & 59.0 & 858 & 75.7 & 26.7 & 1947 & 80.9 & 13.3 & 2805 & 77.8 & 42.4 & 332 & 94.6 & 27.9 & 1117 & 85.5 & 33.9 & 83.1 \\
Dynasor & 89.2 & 3606 & \textcolor{darkred}{-2.3} & 56.7 & 10561 & \textcolor{darkred}{-3.5} & 53.3 & 13398 & \textcolor{darkred}{-5.8} & 58.6 & 4253 & 30.6 & 60.1 & 6805 & 11.4 & 63.6 & 8.6 \\
DEER & 91.2 & 2608 & 26.0 & 70.0 & 9936 & 2.7 & 50.0 & 12397 & 2.1 & 60.6 & 6918 & \textcolor{darkred}{-12.8} & 62.1 & 6950 & 9.5 & 66.8 & 11.9 \\
\rowcolor{pumagreen!25}
PUMA (ours) & 91.2 & 2776 & 21.3 & 66.7 & 8549 & 16.3 & 50.0 & 11788 & 6.9 & 55.6 & 5270 & 14.1 & 59.6 & 5229 & 31.9 & 64.6 & 24.9 \\
\midrule
\multicolumn{18}{l}{\textbf{DeepSeek-R1-Distill-Qwen-32B}} \\
\rowcolor{methodgray!30}
Full CoT & 89.2 & 2957 & 0.0 & 76.7 & 10011 & 0.0 & 63.3 & 12606 & 0.0 & 62.6 & 6186 & 0.0 & 59.3 & 7241 & 0.0 & 70.2 & 0.0 \\
No-Think & 84.2 & 1264 & 57.2 & 56.7 & 5781 & 42.2 & 36.7 & 7435 & 41.0 & 47.0 & 1196 & 80.7 & 49.3 & 4060 & 43.9 & 54.8 & 53.5 \\
CCoT & 87.6 & 1521 & 48.6 & 60.0 & 5600 & 44.1 & 26.7 & 6162 & 51.1 & 60.6 & 2968 & 52.0 & 53.5 & 3356 & 53.6 & 57.7 & 51.4 \\
CoD & 86.4 & 1558 & 47.3 & 56.7 & 5390 & 46.2 & 36.7 & 6103 & 51.6 & 64.7 & 3277 & 47.0 & 53.6 & 3724 & 48.6 & 59.6 & 47.9 \\
Plan\&Budget & 90.8 & 2462 & 16.7 & 56.7 & 6257 & 37.5 & 40.0 & 6492 & 48.5 & 59.6 & 4335 & 29.9 & 54.1 & 4522 & 37.5 & 60.2 & 29.4 \\
Ans.\ Conv. & 65.6 & 631 & 78.7 & 16.7 & 1244 & 87.6 & 6.7 & 1638 & 87.0 & 37.9 & 310 & 95.0 & 30.2 & 889 & 87.7 & 31.4 & 85.6 \\
Dynasor & 92.0 & 3651 & \textcolor{darkred}{-23.5} & 53.3 & 10120 & \textcolor{darkred}{-1.1} & 50.0 & 12822 & \textcolor{darkred}{-1.7} & 58.1 & 4121 & 33.4 & 64.6 & 7050 & 2.6 & 63.6 & \textcolor{darkred}{-2.4} \\
DEER & 94.2 & 2424 & 18.0 & 66.7 & 10439 & \textcolor{darkred}{-4.3} & 46.7 & 11753 & 6.8 & 67.2 & 6222 & \textcolor{darkred}{-0.6} & 63.1 & 6804 & 6.0 & 67.6 & 9.1 \\
\rowcolor{pumagreen!25}
PUMA (ours) & 88.4 & 2296 & 22.3 & 73.3 & 9046 & 9.6 & 60.0 & 11144 & 11.6 & 66.7 & 5394 & 12.8 & 59.9 & 5344 & 26.2 & 69.7 & 22.3 \\
\midrule
\multicolumn{18}{l}{\textbf{Llama-3.1-Nemotron-Nano-8B}} \\
\rowcolor{methodgray!30}
Full CoT & 93.6 & 3109 & 0.0 & 66.7 & 10463 & 0.0 & 50.0 & 10898 & 0.0 & 48.0 & 7001 & 0.0 & 63.1 & 6857 & 0.0 & 64.3 & 0.0 \\
No-Think & 62.8 & 7012 & \textcolor{darkred}{-125.5} & 26.7 & 13842 & \textcolor{darkred}{-32.3} & 16.7 & 15628 & \textcolor{darkred}{-43.4} & 25.8 & 7154 & \textcolor{darkred}{-2.2} & 31.9 & 11935 & \textcolor{darkred}{-74.0} & 32.8 & \textcolor{darkred}{-80.5} \\
CCoT & 86.6 & 2035 & 34.5 & 36.7 & 6701 & 36.0 & 20.0 & 6966 & 36.1 & 43.9 & 4824 & 31.1 & 56.9 & 4485 & 34.6 & 48.8 & 34.1 \\
CoD & 79.0 & 2173 & 30.1 & 50.0 & 6189 & 40.9 & 33.3 & 6610 & 39.4 & 45.0 & 4766 & 31.9 & 51.3 & 4665 & 32.0 & 51.7 & 31.7 \\
Plan\&Budget & 89.0 & 2852 & 8.3 & 43.3 & 7022 & 32.9 & 33.3 & 7460 & 31.6 & 40.9 & 5133 & 26.7 & 53.9 & 5299 & 22.7 & 52.1 & 18.6 \\
Ans.\ Conv. & 55.8 & 735 & 76.3 & 6.7 & 849 & 91.9 & 3.3 & 1050 & 90.4 & 20.2 & 331 & 95.3 & 23.6 & 1024 & 85.1 & 21.9 & 83.7 \\
Dynasor & 91.4 & 3388 & \textcolor{darkred}{-9.0} & 50.0 & 10572 & \textcolor{darkred}{-1.0} & 43.3 & 12467 & \textcolor{darkred}{-14.4} & 45.0 & 6894 & 1.5 & 61.9 & 7059 & \textcolor{darkred}{-3.0} & 58.3 & \textcolor{darkred}{-4.7} \\
DEER & 91.8 & 2403 & 22.7 & 53.3 & 12814 & \textcolor{darkred}{-22.5} & 50.0 & 14805 & \textcolor{darkred}{-35.9} & 49.5 & 28931 & \textcolor{darkred}{-313.2} & 61.2 & 6004 & 12.4 & 61.2 & \textcolor{darkred}{-30.7} \\
\rowcolor{pumagreen!25}
PUMA (ours) & 92.6 & 2621 & 15.7 & 70.0 & 9180 & 12.3 & 50.0 & 8928 & 18.1 & 48.5 & 6349 & 9.3 & 62.2 & 5007 & 27.0 & 64.7 & 20.1 \\
\midrule
\multicolumn{18}{l}{\textbf{Qwen3-30B-A3B-Thinking}} \\
\rowcolor{methodgray!30}
Full CoT & 94.4 & 5206 & 0.0 & 83.3 & 16354 & 0.0 & 83.3 & 18473 & 0.0 & 72.7 & 7372 & 0.0 & 75.0 & 12668 & 0.0 & 81.7 & 0.0 \\
No-Think & 91.8 & 2516 & 51.7 & 60.0 & 7107 & 56.5 & 50.0 & 9246 & 50.0 & 73.7 & 8221 & \textcolor{darkred}{-11.5} & 68.4 & 5503 & 56.6 & 68.8 & 45.3 \\
CCoT & 89.2 & 2361 & 54.6 & 36.7 & 7217 & 55.9 & 20.0 & 7341 & 60.3 & 72.7 & 4021 & 45.5 & 53.0 & 5220 & 58.8 & 54.3 & 55.5 \\
CoD & 82.8 & 3652 & 29.9 & 23.3 & 7770 & 52.5 & 10.0 & 7788 & 57.8 & 65.2 & 5247 & 28.8 & 45.3 & 6297 & 50.3 & 45.3 & 40.4 \\
Plan\&Budget & 91.8 & 2532 & 51.4 & 43.3 & 7100 & 56.6 & 36.7 & 7306 & 60.5 & 64.7 & 4635 & 37.1 & 61.3 & 5039 & 60.2 & 59.6 & 53.9 \\
Ans.\ Conv. & 57.0 & 635 & 87.8 & 0.0 & 921 & 94.4 & 0.0 & 1738 & 90.6 & 35.9 & 333 & 95.5 & 23.1 & 1050 & 91.7 & 23.2 & 90.9 \\
Dynasor & 94.8 & 5432 & \textcolor{darkred}{-4.3} & 86.7 & 14372 & 12.1 & 76.7 & 17959 & 2.8 & 67.7 & 5001 & 32.2 & 72.4 & 12748 & \textcolor{darkred}{-0.6} & 79.7 & 3.0 \\
DEER & 94.8 & 3416 & 34.9 & 83.3 & 14682 & 10.5 & 80.0 & 17181 & 7.3 & 74.8 & 8062 & \textcolor{darkred}{-8.2} & 73.5 & 9749 & 23.4 & 81.3 & 22.4 \\
\rowcolor{pumagreen!25}
PUMA (ours) & 94.2 & 3707 & 28.8 & 90.0 & 12914 & 21.0 & 80.0 & 15767 & 14.7 & 75.8 & 6058 & 17.8 & 72.3 & 8652 & 31.7 & 82.5 & 28.2 \\
\bottomrule
\end{tabular}
\end{adjustbox}
\vspace{-0.3cm}
\end{table*}

\begin{table}[t]
\centering
\caption{Full component ablation and probe overhead on DS-R1-Distill-Qwen-7B and Qwen3-30B-A3B-Thinking, averaged over AIME24, AIME25, GPQA-Diamond, MATH-500, and OlympiadBench. \textit{Probes/q} is the average number of trial-answer probes per question, and Probe\(\times\) is normalized by full PUMA on the same model. ``w/o RD Gate'' disables redundancy-based candidate filtering by setting the redundancy threshold to zero, causing Answer Verification to be invoked at every step.}
\label{tab:ablation_full}
\vspace{0.2cm}
\resizebox{\textwidth}{!}{
\renewcommand{\arraystretch}{1.05}
\setlength{\tabcolsep}{5pt}
\begin{tabular}{l c c c c c c c c}
\toprule
& \multicolumn{4}{c}{\textbf{DS-7B}}
& \multicolumn{4}{c}{\textbf{Qwen3-30B-T}} \\
\cmidrule(lr){2-5} \cmidrule(lr){6-9}
\textbf{Configuration}
& \textbf{Acc}$\uparrow$
& \textbf{TR}$\uparrow$
& \textbf{Probes/q}$\downarrow$
& \textbf{Probe}$\times\downarrow$
& \textbf{Acc}$\uparrow$
& \textbf{TR}$\uparrow$
& \textbf{Probes/q}$\downarrow$
& \textbf{Probe}$\times\downarrow$ \\
\midrule
\rowcolor{pumagreen!25}
\textbf{PUMA (full)}
& \textbf{60.2} & \textbf{35.6} & \textbf{13.75} & \textbf{1.0}\(\times\)
& \textbf{82.5} & \textbf{28.2} & \textbf{10.46} & \textbf{1.0}\(\times\) \\
\midrule
w/o RD Gate
& 56.1 (\textcolor{darkred}{-4.1}) & 46.0 & 45.69 & 3.3\(\times\)
& 78.6 (\textcolor{darkred}{-3.9}) & 42.9 & 45.11 & 4.3\(\times\) \\
w/o Loop Breaker
& 59.2 (\textcolor{darkred}{-1.0}) & 22.6 & 17.48 & 1.3\(\times\)
& 80.5 (\textcolor{darkred}{-2.0}) & 27.4 & 11.88 & 1.1\(\times\) \\
AV only (w/o RD and LB)
& 58.3 (\textcolor{darkred}{-1.9}) & 37.3 & 55.27 & 4.0\(\times\)
& 77.6 (\textcolor{darkred}{-4.9}) & 42.8 & 45.13 & 4.3\(\times\) \\
\midrule
w/o Answer Consistency
& 57.3 (\textcolor{darkred}{-2.9}) & 37.8 & 15.96 & 1.2\(\times\)
& 79.9 (\textcolor{darkred}{-2.6}) & 34.7 & 9.04 & 0.9\(\times\) \\
w/o Confidence Gate
& 53.7 (\textcolor{darkred}{-6.5}) & 55.3 & 8.64 & 0.6\(\times\)
& 72.6 (\textcolor{darkred}{-9.9}) & 47.9 & 5.11 & 0.5\(\times\) \\
w/o AC and CG
& 44.0 (\textcolor{darkred}{-16.2}) & 66.1 & 4.55 & 0.3\(\times\)
& 61.9 (\textcolor{darkred}{-20.6}) & 55.8 & 2.95 & 0.3\(\times\) \\
\bottomrule
\end{tabular}
}
\end{table}

\subsection{Full Component Ablation}
\label{appendix:ablation}


Table~\ref{tab:ablation} in the main text reports a compact ablation on DS-7B. Table~\ref{tab:ablation_full} extends the analysis to both DS-7B and Qwen3-30B-A3B-Thinking, reports trial-answer probe statistics, and includes two additional variants: \emph{AV only} and \emph{w/o AC and CG}. \textit{Probes/q} is the average number of trial-answer probes per question, and Probe\(\times\) is normalized by full PUMA on the same model.

The full results provide a more detailed view of how each component affects accuracy, token reduction, and probing overhead. First, disabling the RD gate causes Answer Verification to be invoked at every eligible generated reasoning step. This increases token reduction, but also lowers accuracy and increases probe overhead substantially: 3.3\(\times\) more probes on DS-7B and 4.3\(\times\) more probes on Qwen3-30B-T. This shows that answer-level verification without redundancy-based candidate filtering is both less reliable and more expensive. Second, the \emph{AV only} variant disables both the RD gate and Loop Breaker, leaving Answer Verification as the only stopping mechanism. Although it can still reduce tokens, it requires 4.0--4.3\(\times\) more probes than full PUMA and does not match the full model's accuracy--efficiency balance, confirming the value of reasoning-level stopping signals. Third, removing the Loop Breaker has model-dependent effects: it sharply reduces token savings on DS-7B, while having smaller effect on Qwen3-30B-T, suggesting that different LRMs rely on the fallback to different degrees. Finally, disabling the answer-verification gates shows their role in preventing premature exits. Removing either Answer Consistency or the Confidence Gate degrades accuracy, and disabling both gates causes the largest accuracy collapse despite high token reduction. Overall, the full ablation shows that PUMA's components are complementary: the RD gate controls when verification is invoked, Answer Verification filters unreliable candidates, and the Loop Breaker provides additional savings when reasoning enters sustained redundancy.

\subsection{Detailed Latency Analysis}
\label{appendix:latency_detailed}

\begin{table*}[h]
\centering
\caption{Per-benchmark wall-clock latency across three models. For each benchmark we report accuracy (Acc, \%, $\uparrow$), token reduction
(TR, \%, $\uparrow$), and average wall-clock seconds per question ($s$/$q$, $\downarrow$).
Overall Speedup is computed from total wall-clock time relative to Full CoT. Speedup below 1$\times$ (in
\textcolor{darkred}{red}) indicates slower wall-clock performance than Full CoT. For the
$s$/$q$ columns, the \textbf{best} (lowest) and \underline{second-best} values within
each model are highlighted.}
\label{tab:latency_detailed}
\vspace{0.2cm}
\begin{adjustbox}{width=\textwidth}
\tiny
\renewcommand{\arraystretch}{0.95}
\setlength{\tabcolsep}{2.5pt}
\begin{tabular}{l ccc ccc ccc ccc | cccc}
\toprule
& \multicolumn{3}{c}{\textbf{MATH-500}}
& \multicolumn{3}{c}{\textbf{AIME24}}
& \multicolumn{3}{c}{\textbf{AIME25}}
& \multicolumn{3}{c}{\textbf{OlympiadBench}}
& \multicolumn{4}{c}{\textbf{Overall}} \\
\cmidrule(lr){2-4} \cmidrule(lr){5-7} \cmidrule(lr){8-10} \cmidrule(lr){11-13} \cmidrule(lr){14-17}
Method
& Acc$\uparrow$ & TR$\uparrow$ & $s$/$q$$\downarrow$
& Acc$\uparrow$ & TR$\uparrow$ & $s$/$q$$\downarrow$
& Acc$\uparrow$ & TR$\uparrow$ & $s$/$q$$\downarrow$
& Acc$\uparrow$ & TR$\uparrow$ & $s$/$q$$\downarrow$
& Acc$\uparrow$ & TR$\uparrow$ & $s$/$q$$\downarrow$ & Speedup$\uparrow$ \\
\midrule
\multicolumn{17}{l}{\textbf{DeepSeek-R1-Distill-Qwen-7B}} \\
\rowcolor{methodgray!30}
Full CoT & 90.0 & 0.0 & \underline{0.98} & 50.0 & 0.0 & \underline{10.18} & 43.3 & 0.0 & \underline{10.56} & 57.6 & 0.0 & 2.57 & 60.2 & 0.0 & 2.30 & 1.00$\times$ \\
DEER & 90.6 & 39.6 & 1.63 & 40.0 & 14.1 & 12.64 & 50.0 & 13.2 & 12.80 & 57.5 & 20.2 & 3.78 & 59.5 & 27.7 & 3.35 & \textcolor{darkred}{0.69$\times$} \\
Dynasor & 84.2 & 30.6 & 1.03 & 23.3 & 21.7 & 11.86 & 33.3 & 9.1 & 15.26 & 49.0 & 34.1 & \underline{1.84} & 47.5 & 31.8 & \underline{2.08} & 1.11$\times$ \\
\rowcolor{pumagreen!25}
\textbf{PUMA (ours)} & \textbf{89.6} & \textbf{38.6} & \textbf{0.87} & \textbf{60.0} & \textbf{30.3} & \textbf{9.41} & \textbf{46.7} & \textbf{29.8} & \textbf{6.88} & \textbf{55.6} & \textbf{43.2} & \textbf{1.64} & \textbf{63.0} & \textbf{40.7} & \textbf{1.64} & \textbf{1.40$\times$} \\
\midrule
\multicolumn{17}{l}{\textbf{DeepSeek-R1-Distill-Qwen-14B}} \\
\rowcolor{methodgray!30}
Full CoT & 91.0 & 0.0 & \underline{2.09} & 63.3 & 0.0 & 20.20 & 50.0 & 0.0 & \underline{20.95} & 61.5 & 0.0 & \underline{7.32} & 66.5 & 0.0 & \underline{5.84} & 1.00$\times$ \\
DEER & 91.2 & 26.0 & 3.68 & 70.0 & 2.7 & 22.21 & 50.0 & 2.1 & 23.88 & 62.1 & 9.5 & 10.36 & 68.3 & 15.8 & 8.27 & \textcolor{darkred}{0.71$\times$} \\
Dynasor & 89.2 & \textcolor{darkred}{-2.3} & 8.59 & 56.7 & \textcolor{darkred}{-3.5} & \textbf{12.54} & 53.3 & \textcolor{darkred}{-5.8} & 25.09 & 60.1 & 11.4 & 42.96 & 64.8 & 5.1 & 27.87 & \textcolor{darkred}{0.21$\times$} \\
\rowcolor{pumagreen!25}
\textbf{PUMA (ours)} & \textbf{91.2} & \textbf{21.3} & \textbf{1.84} & \textbf{66.7} & \textbf{16.3} & \underline{15.62} & \textbf{50.0} & \textbf{6.9} & \textbf{20.92} & \textbf{59.6} & \textbf{31.9} & \textbf{5.39} & \textbf{66.9} & \textbf{26.6} & \textbf{4.58} & \textbf{1.28$\times$} \\
\midrule
\multicolumn{17}{l}{\textbf{Llama-3.1-Nemotron-Nano-8B}} \\
\rowcolor{methodgray!30}
Full CoT & 93.6 & 0.0 & 1.07 & 66.7 & 0.0 & \textbf{7.68} & 50.0 & 0.0 & \textbf{9.82} & 63.1 & 0.0 & \underline{2.83} & 68.3 & 0.0 & \underline{2.41} & 1.00$\times$ \\
DEER & 91.8 & 22.7 & 1.83 & 53.3 & \textcolor{darkred}{-22.5} & 23.10 & 50.0 & \textcolor{darkred}{-35.9} & 26.61 & 61.2 & 12.4 & 4.99 & 64.1 & 14.5 & 4.67 & \textcolor{darkred}{0.52$\times$} \\
Dynasor & 91.4 & \textcolor{darkred}{-9.0} & \underline{0.99} & 50.0 & \textcolor{darkred}{-1.0} & 9.17 & 43.3 & \textcolor{darkred}{-14.4} & 11.37 & 61.9 & \textcolor{darkred}{-3.0} & 24.32 & 61.7 & \textcolor{darkred}{-5.7} & 14.19 & \textcolor{darkred}{0.17$\times$} \\
\rowcolor{pumagreen!25}
\textbf{PUMA (ours)} & \textbf{92.6} & \textbf{15.7} & \textbf{0.85} & \textbf{70.0} & \textbf{12.3} & \underline{8.71} & \textbf{50.0} & \textbf{18.1} & \underline{11.31} & \textbf{62.2} & \textbf{27.0} & \textbf{2.30} & \textbf{68.7} & \textbf{21.9} & \textbf{2.09} & \textbf{1.15$\times$} \\
\bottomrule
\end{tabular}
\end{adjustbox}
\vspace{-0.3cm}
\end{table*}

Figure~\ref{fig:latency} in the main text reports average speedup across
benchmarks. Table~\ref{tab:latency_detailed} provides per-benchmark
wall-clock latency for three models. PUMA achieves consistent speedup
across all settings (1.15--1.40$\times$), while DEER is uniformly slower
than Full CoT (0.52--0.71$\times$) and Dynasor ranges from 1.11$\times$
to 0.17$\times$ depending on model size. These results show that token reduction alone does not guarantee wall-clock gains. Both DEER and Dynasor can reduce generated tokens yet still run slower than Full CoT,
because frequent trial-answer probing introduces additional forward-pass overhead.
When this overhead outweighs the time saved from shorter reasoning traces, positive
token reduction does not translate into end-to-end speedup.

\section{Budget Tuning Does Not Rescue Prompt-Based Baselines}
\label{appendix:baseline_tuning}

A natural question is whether the accuracy gap between PUMA and prompt-based
baselines can be closed by relaxing their word-budget instructions. We therefore
sweep the word budgets specified in the prompts for CCoT and CoD on DS-7B
LiveCodeBench. We note that LRMs do not always strictly follow these requested
budgets; thus, the reported token reduction reflects the actual generated outputs
rather than the nominal prompt budget. Table~\ref{tab:baseline_tuning} shows that
budget tuning does not close the gap: CCoT remains in the 42.7--45.8\% accuracy
range across requested budgets from 30 to 300 words, and CoD remains in the
43.2--44.9\% range across requested budgets from 5 to 30 words per step, both
below PUMA's 50.3\%.

\begin{table*}[h]
\vspace{0.2cm}
\centering
\caption{Budget-sweep analysis of prompt-based baselines on DS-7B LiveCodeBench.
Relaxing CCoT's global word budget or CoD's per-step word budget does not close the
accuracy gap to PUMA. \(\Delta\) is accuracy change relative to Full CoT. ``def.'' denotes the default budget used in the main experiments.}
\label{tab:baseline_tuning}
\resizebox{0.92\textwidth}{!}{
\renewcommand{\arraystretch}{1.08}
\setlength{\tabcolsep}{6pt}
\begin{tabular}{l c >{\columncolor{pumagreen!25}}c ccccc ccc}
\toprule
& \multicolumn{2}{c}{\textbf{Reference}}
& \multicolumn{5}{c}{\textbf{CCoT: global word budget}}
& \multicolumn{3}{c}{\textbf{CoD: per-step word budget}} \\
\cmidrule(lr){2-3} \cmidrule(lr){4-8} \cmidrule(lr){9-11}
\textbf{Metric}
& \textbf{Full CoT} & \textbf{PUMA}
& \textbf{30} & \textbf{45 (def.)} & \textbf{75} & \textbf{150} & \textbf{300}
& \textbf{5 (def.)} & \textbf{10} & \textbf{30} \\
\midrule
Acc\(\uparrow\)
& 51.82 & \textbf{50.34}
& 45.57 & 45.80 & 42.73 & 44.20 & 44.32
& 43.18 & 43.52 & 44.89 \\
\(\Delta\)\(\uparrow\)
& \textemdash & \textbf{\textcolor{darkred}{-1.48}}
& \textcolor{darkred}{-6.25}
& \textcolor{darkred}{-6.02}
& \textcolor{darkred}{-9.09}
& \textcolor{darkred}{-7.62}
& \textcolor{darkred}{-7.50}
& \textcolor{darkred}{-8.64}
& \textcolor{darkred}{-8.30}
& \textcolor{darkred}{-6.93} \\
TR\(\uparrow\)
& 0.0 & \textbf{19.0}
& 42.9 & 42.7 & 41.8 & 41.4 & 40.0
& 40.5 & 40.6 & 41.3 \\
\bottomrule
\end{tabular}
}
\vspace{-0.3cm}
\end{table*}

\section{Reasoning Quality Evaluation Details}
\label{appendix:quality_eval}

We evaluate the quality of retained reasoning chains using an LLM-as-Judge protocol.
The goal is to measure whether an early-exit method preserves a readable and sufficient
reasoning chain, beyond merely preserving final-answer accuracy.

\paragraph{Judge model and protocol.}
We use GPT-5.4-thinking as the judge via the OpenAI Batch API.  Each instance contains
the original question and the retained reasoning chain produced by the evaluated method. The judge is
instructed to first provide a brief rationale and then assign scores, following standard
LLM-as-Judge practice~\cite{li-etal-2025-generation}. The judge prompt template is summarized in Table~\ref{tab:quality_judge_prompt}.

\paragraph{Scope.}
We evaluate three (model, benchmark) combinations:
DS-R1-Distill-Qwen-7B on GPQA-Diamond (198 questions),
Nemotron-Nano-8B on GPQA-Diamond (198 questions), and
Qwen3-30B-A3B-Thinking on MATH-500 (500 questions).
All questions are evaluated regardless of answer correctness, avoiding
subset-selection bias. The judge is not given gold answers, method names, or answer-correctness labels. It evaluates each retained reasoning chain as written, including whether the chain provides sufficient and coherent support for its own stated final answer.
Since LLM-as-Judge evaluation over long reasoning chains is substantially more
expensive than answer-only evaluation, we use these representative combinations
rather than the full model--benchmark grid. The reported reasoning-quality evaluation
costs approximately \$300 in OpenAI API usage.

\paragraph{Rubric.}
Each reasoning chain is scored on a 10--100 scale, in increments of 10, along four
dimensions:
\begin{itemize}[itemsep=1pt, topsep=2pt, parsep=1pt, leftmargin=1.5em]
    \item \textbf{Completeness}: whether the chain contains a semantically sufficient
    derivation from the problem statement to the final answer.
    \item \textbf{Coherence}: whether the chain forms a smooth and logically connected
    line of reasoning, without abrupt jumps, contradictions, or disruptive interruptions.
    \item \textbf{Conciseness}: whether the chain avoids unnecessary repetition,
    redundant verification, and non-productive loops.
    \item \textbf{Justification}: whether a reader can understand why the final answer
    follows from the main derivation.
\end{itemize}

\paragraph{Few-shot calibration.}
We include six manually written in-context examples covering distinct quality
patterns: complete and concise, complete but redundant, brief but sufficient,
brief and incomplete, derivation with a corrected false start, and unresolved
contradiction. All examples are based on a single math problem to avoid
confounding rubric calibration with problem difficulty.

\paragraph{Evaluator bias control.}
To mitigate evaluator bias, the judge is not shown method names and evaluates all retained reasoning chains using the same rubric. The judge model is not used to train PUMA's Redundancy Detector or to select stopping hyperparameters. Although the detector supervision uses LLM-generated redundancy annotations, it is constructed at the reasoning-step-pair level, whereas the judge evaluates complete retained reasoning chains; the two stages use different inputs, prompts, and objectives. To further validate judge reliability, we randomly sample 100 pairwise comparisons from the evaluation results. Each comparison contains two anonymized retained reasoning chains for the same question, and a human annotator independently selects the chain with higher overall explanation quality. The human judgments agree with GPT-5.4-thinking's score-induced preferences in 85\% of cases, suggesting that the judge-based relative comparisons are broadly aligned with human assessment.
The absolute scores are modest across all methods, with the best average score in the mid-50s. This is because the rubric evaluates the quality of the retained reasoning chain, rather than final-answer correctness alone: a chain must be complete, coherent, concise, and sufficiently justified to receive a high score. A correct final answer is therefore not sufficient if the retained reasoning is incomplete, hard to follow, overly repetitive, or weakly justified. We therefore focus on relative comparisons across methods under the same anonymized judge and rubric.

\begin{table}[t]
\small
\centering
\begin{tcolorbox}[width=0.95\textwidth]
\textcolor{blue}{\textbf{Role:}}\par
You are an expert evaluator of reasoning chains. Your task is to evaluate
how well a reasoning chain communicates a sufficient, productive derivation of the stated answer.

\medskip
\textcolor{blue}{\textbf{Core Principle:}}\par
Reward one clear successful line of reasoning. Do not reward extra exploratory branches,
repeated verification, generic tutoring preambles, or exhaustive enumeration once the
answer is already derivable.

\medskip
\textcolor{blue}{\textbf{Scoring Dimensions:}}\par
Score each chain from 10 to 100, in steps of 10, on:
Completeness, Coherence, Conciseness, and Answer Justification.

\medskip
\textcolor{blue}{\textbf{Important Constraints:}}\par
Evaluate the reasoning chain as written.
If the stated final answer is unsupported, contradicted, or does not follow from the reasoning, reflect this in Completeness and Answer Justification. Score Coherence and Conciseness independently. Do not use method names, gold answers, or external knowledge. Do not fill in missing proof steps with your own solution.

\medskip
\textcolor{blue}{\textbf{Rubric:}}\par
Completeness measures whether the chain contains a sufficient derivation from the
problem to the stated final answer. Coherence measures whether the reasoning flow is logically
connected, without abrupt jumps, contradictions, or unresolved self-corrections.
Conciseness measures whether the chain avoids repetition, redundant verification, and
non-productive loops. Justification measures whether a reader can understand why the
stated final answer follows from the main derivation.

\medskip
\textcolor{blue}{\textbf{Few-shot Examples:}}\par
\textit{[The prompt includes six calibration examples covering concise sufficient
reasoning, complete but redundant reasoning, brief but incomplete reasoning, false
starts, and unresolved contradictions.]}

\medskip
\textcolor{blue}{\textbf{Output Format:}}\par
Return only a JSON object with a brief rationale and four scores:
\par
\texttt{\{"brief\_rationale": "<1--2 sentences>",}
\par
\texttt{"completeness": <10--100>, "coherence": <10--100>,}
\par
\texttt{"conciseness": <10--100>, "justification": <10--100>\}}
\end{tcolorbox}
\caption{Judge prompt template for reasoning-chain quality evaluation. The full prompt includes per-dimension anchors and six calibration examples.}
\label{tab:quality_judge_prompt}
\end{table}

\section{Internalizing PUMA into Model Weights: Implementation Details}
\label{appendix:internalization_details}

This appendix provides full details for the internalization experiments reported in
Section~\ref{sec:internalization}. All variants fine-tune DS-R1-Distill-Qwen-7B with
LoRA (rank 64, alpha 128, \texttt{all-linear}) and are evaluated with pure vLLM
inference, without any RD or AV at test time. Our RL training design is partly inspired by DEER~\cite{yang2026dynamic}.

\textbf{Training data.}
We use the 12K problems from the Lightman split of MATH-benchmark~\cite{hendrycks2021measuring},
which is mathematics-only; consequently, MATH-500 performance can be regarded as
in-distribution behavior, while the resulting models generalize zero-shot to AIME24
(harder math) and GPQA-D (cross-domain science). We verify that no problem in this
train split overlaps with the MATH-500 test split, ensuring zero contamination of our
in-distribution evaluation. For each problem, we generate one Full CoT trajectory
from the base model and run the PUMA inference pipeline to record RD-flagged
candidate exit positions and the verified stopping point.

\textbf{Per-variant data construction.}
PUMA-SFT keeps rows $(Q, R_{\le t^*}, A)$ where PUMA verifies an early exit at $t^*$
and the regenerated answer is correct, filtered to aggressive compressions
($t^*/|R| < 0.6$), yielding $\sim$6.5K examples. FixedExit-SFT replaces $t^*$ with the
earliest fixed-interval position ($K{=}3$) yielding a correct forced-stop answer;
filtering is identical. PUMA-DPO pairs PUMA-truncated chains (chosen) with Full CoT
chains (rejected) when both are correct ($\sim$5.8K pairs). 
PUMA-RL and FixedExit-RL
use a pre-expanded GRPO dataset where each row is a question concatenated with a
reasoning prefix and the closing $\langle/\text{think}\rangle$ token, with the prefix
truncated at a PUMA-flagged or fixed-interval position; the model is trained to
generate only the answer phase that follows ($\sim$15K samples each).
Standard-SFT and Standard-GRPO use no exit-position signal: Standard-SFT trains on
the base model's untruncated Full CoT trajectories, and Standard-GRPO performs
standard GRPO on $(Q, \text{ground truth})$ pairs without any prefix conditioning,
generating the entire $\langle\text{think}\rangle\dots\langle/\text{think}\rangle$
chain from scratch at each rollout.

\textbf{Training.}
SFT runs for 3 epochs with learning rate $2{\times}10^{-4}$, batch size 1 with
gradient accumulation 16, and \texttt{max\_length} 16384. DPO runs for 1 epoch with the same learning rate,
default $\beta$, and \texttt{rpo\_alpha}=0.1 (an SFT anchor on the chosen response
that we found necessary to prevent drift). GRPO runs for \texttt{max\_steps}=1500
with learning rate $1{\times}10^{-6}$, \texttt{num\_generations}=4 per group,
\texttt{max\_completion\_length}=4096, and vLLM colocated generation. 
For Standard-GRPO, the reward is pure answer correctness. For PUMA-RL and
FixedExit-RL, which generate only the answer phase after a provided reasoning
prefix, we use the following reward:
\begin{equation}
R = r_{\text{correct}} \cdot \big(1.0 + 0.5\,(1 - \ell/4096)\big) + r_{\text{rank}},
\label{eq:reward}
\end{equation}
where $r_{\text{correct}} \in \{0, 1\}$ indicates answer correctness, $\ell$ is the
completion length in tokens, and $r_{\text{rank}} \in \{+0.5, +0.25, 0, -0.25\}$ is a
within-group rank bonus favoring the shortest correct trajectory among the four rollouts.
All training uses
bfloat16 with gradient checkpointing on 4$\times$NVIDIA GH200, implemented in ms-swift~\cite{zhao2024swiftascalablelightweightinfrastructure}.



\end{document}